\documentclass[letterpaper]{article} % DO NOT CHANGE THIS
\usepackage{aaai24}  % DO NOT CHANGE THIS
\usepackage{times}  % DO NOT CHANGE THIS
\usepackage{helvet}  % DO NOT CHANGE THIS
\usepackage{courier}  % DO NOT CHANGE THIS
\usepackage[hyphens]{url}  % DO NOT CHANGE THIS
\usepackage{graphicx} % DO NOT CHANGE THIS
\def\UrlFont{\rm}  % DO NOT CHANGE THIS
\usepackage{natbib}  % DO NOT CHANGE THIS AND DO NOT ADD ANY OPTIONS TO IT
\usepackage{caption} % DO NOT CHANGE THIS AND DO NOT ADD ANY OPTIONS TO IT
\usepackage{algorithm}
\usepackage{algorithmic}

\usepackage{amsmath}
\usepackage{amssymb}
\usepackage{amsthm}
\usepackage{multirow}
\usepackage{makecell}
\usepackage{subfig}
\usepackage{bm}
\usepackage{array}
\usepackage{booktabs}
\usepackage{wasysym}

\title{
MLNet: Mutual Learning Network with Neighborhood Invariance\\for Universal Domain Adaptation
}
\author{
    Yanzuo Lu\textsuperscript{\rm 1}, 
    Meng Shen\textsuperscript{\rm 1}, 
    Andy J Ma\textsuperscript{\rm 1,\rm 2,\rm 3 \footnote{Corresponding author.}}, 
    Xiaohua Xie\textsuperscript{\rm 1,\rm 2,\rm 3}, 
    Jian-Huang Lai\textsuperscript{\rm 1,\rm 2,\rm 3,\rm 4}
}
\affiliations{
    \textsuperscript{\rm 1}School of Computer Science and Engineering, Sun Yat-sen University, Guangzhou, China\\
    \textsuperscript{\rm 2}Guangdong Province Key Laboratory of Information Security Technology, China\\
    \textsuperscript{\rm 3}Key Laboratory of Machine Intelligence and Advanced Computing, Ministry of Education, China\\
    \textsuperscript{\rm 4}Pazhou Lab (HuangPu), Guangzhou, China\\
    \{luyz5, shenm25\}@mail2.sysu.edu.cn, \{majh8, xiexiaoh6, stsljh\}@mail.sysu.edu.cn\\
}

\begin{document}

\maketitle

% \begin{abstract}
% AAAI creates proceedings, working notes, and technical reports directly from electronic source furnished by the authors. To ensure that all papers in the publication have a uniform appearance, authors must adhere to the following instructions.
% \end{abstract}

\begin{abstract}

Universal domain adaptation (UniDA) is a practical but challenging problem, in which information about the relation between the source and the target domains is not given for knowledge transfer.
Existing UniDA methods may suffer from the problems of overlooking intra-domain variations in the target domain and difficulty in separating between the similar known and unknown class.
To address these issues, we propose a novel \textbf{Mutual Learning Network (MLNet)} with neighborhood invariance for UniDA.
In our method, confidence-guided invariant feature learning with self-adaptive neighbor selection is designed to reduce the intra-domain variations for more generalizable feature representation.
By using the cross-domain mixup scheme for better unknown-class identification, the proposed method compensates for the misidentified known-class errors by mutual learning between the closed-set and open-set classifiers.
Extensive experiments on three publicly available benchmarks demonstrate that our method achieves the best results compared to the state-of-the-arts in most cases and significantly outperforms the baseline across all the four settings in UniDA.
Code is available at https://github.com/YanzuoLu/MLNet.

\end{abstract}
\section{Introduction}
\label{sec:1}

Deep neural networks have made significant progress in computer vision, natural language processing and many other research areas in recent years~\cite{lecun2015deep}. 
With the rapid development of hardware, existing architectures are expanding with strong data-fitting capacity. 
Nevertheless, supervised learners may not generalize well when testing on novel domains with distribution shift.
Unsupervised domain adaptation (UDA) is proposed to mitigate this by transferring knowledge from the labeled source domain to the unlabeled target domain~\cite{kouw2021review}.
Denote the label sets of the source and target domains as $\mathcal{C}_s$ and $\mathcal{C}_t$, respectively. 
Though existing works on closed-set domain adaptation (CDA)~\cite{chen2022reusing, rangwani2022closer} achieve promising results under the assumption that $\mathcal{C}_s = \mathcal{C}_t$, the validity of this assumption cannot be verified without labels in target domain.

Without the assumption of identical label sets across domains in CDA, several variants of UDA have been developed to solve the practical problem of category shift.
In open-set domain adaptation (ODA)~\cite{saito2018open, jing2021towards} $\mathcal{C}_s$ is a subset of $\mathcal{C}_t$, while  $\mathcal{C}_s$ is a superset of $\mathcal{C}_t$ in partial domain adaptation (PDA)~\cite{lin2022adversarial, cao2018partial}.
If $\mathcal{C}_s\cap\mathcal{C}_t\neq \emptyset$, $\mathcal{C}_s-\mathcal{C}_t\neq\emptyset$ and $\mathcal{C}_t-\mathcal{C}_s\neq\emptyset$, it becomes open partial domain adaptation (OPDA)~\cite{busto2017open, kundu2020universal}. 
More difficultly but practically, the relation between $\mathcal{C}_s$ and $\mathcal{C}_t$ is unknown, giving rise to the problem of universal domain adaptation (UniDA)~\cite{you2019universal,shen2023collaborative}. 
The objective of UniDA is to classify target samples into either one of the known classes in the source domain or an unknown class, without prior knowledge on the relation between label sets across domains.

\begin{figure}[t]
    \centering
    \includegraphics[width=\linewidth]{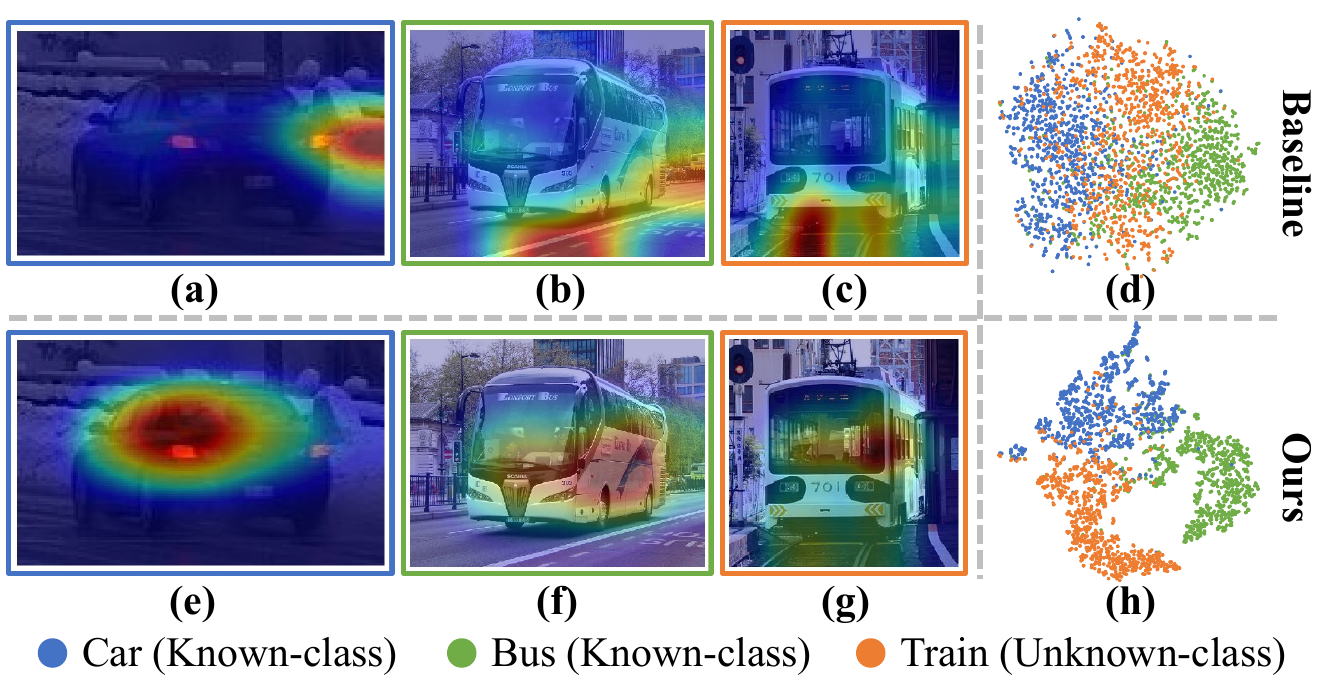}
    \caption{
        CAM~\cite{selvaraju2017grad} and t-SNE~\cite{van2008visualizing} visualizations on the VisDA~\cite{peng2017visda} dataset.
        Our method outperforms the OVANet baseline by learning more discriminative features for both known and unknown classes even though they are similar. 
        More t-SNE examples are provided in supplementary.
    }
    \label{fig:intro}
\end{figure}

In UniDA, it is challenging to identify unknown-class samples in the target domain, since the labeled source domain does not contain the unknown class.
In several works~\cite{you2019universal,fu2020learning}, a fixed threshold is selected to reject a certain proportion of target samples for learning the concept of the unknown.
Due to the difficulty in determining a proper decision threshold, universal classifier is designed for both the known classes and the additional unknown class~\cite{yang2022one,chen2022geometric}, where samples are classified into one of the $|\mathcal{C}_s|+1$ categories.
To address the problem caused by category-shift, \citeauthor{chen2022geometric} construct mixup features within the source domain to simulate the potential unknown-class samples in the target domain.
Although the source mixup features obtained by combining samples of different classes will not belong to any of the known classes, they may fail to represent the target unknown class because of the distribution shift.

Another approach to avoid determining the threshold is the OVANet~\cite{saito2021ovanet} which trains $|\mathcal{C}_s|$ one-vs-all (open-set) classifiers for each known class.
In order to train these classifiers, \citeauthor{saito2021ovanet} leverage the inter-class distances learned in the source domain and adapt them to the target domain through entropy minimization~\cite{grandvalet2004semi}.
Target samples are first classified into one of the known classes (e.g., class $l$) by a $|\mathcal{C}_s|$-way closed-set classifier, and then decided as known or unknown by the scores from the $l$-th open-set sub-classifier.
Despite the great success, the feature space may not be discriminative enough for separating the known and unknown classes. 
As shown in Figure~\ref{fig:intro}(a)(b), the OVANet may wrongly pay attention to the background area even for target samples of the known classes.
Moreover, the unknown class may be mistakenly identified as known by simple entropy minimization, so that the attention to the unknown-class samples are misleading as illustrated in Figure~\ref{fig:intro}(c).
With the difficulty in distinguishing between known and unknown classes, features of similar objects in the target domain are scatted and mixed together (Figure~\ref{fig:intro}(d)).

To address these challenges, we propose a novel Mutual Learning Network (MLNet) with neighborhood invariance for UniDA.
Due to the distribution shift, the intra-domain variations in the target domain differs from the source domain in different degrees.
Without taking the intra-domain variations into account, existing UniDA methods cannot learn feature representations well generalized to the target domain.
In our method, we propose to learn more generalizable features in the target domain by neighborhood invariance.
For this purpose, reliable neighbors of each target sample are retrieved and pairwise similarities between the input and its neighbors are maximized to reduce the intra-domain variations.
The neighborhood invariance means that samples within the neighborhood have a high probability to share the same category.
A self-adaptive neighbor search strategy based on the nearest neighbor is designed to overcome limitations of different sample sizes within or in different datasets.
Moreover, we measure the confidence of each neighbor by the additional information about the neighborhood relationship to guide the invariant feature learning and obtain a more robust embedding space in the target domain.

Beyond learning better feature representation with neighborhood invariance, we propose to further mitigate the category shift by cross-domain manifold mixup (CMM) for unknown-class discovery and mutual learning via consistency constraint.
Inter-class distances in the source domain are observed to be not trustworthy when adapted to target domain.
Thus, a novel cross-domain manifold mixup scheme is designed to explicitly simulate unknown-class samples by leveraging the arbitrary intermediate states of mixups across domains to smooth the transfer.
Though the CMM greatly improves the unknown-class identification, the decision space of the known class will become smaller such that a number of known-class samples are misidentified as unknown.
To trade off this error, we further propose an innovative consistency constraint between the closed-set and open-set classifiers.
The misidentified known-class samples with inconsistent decisions are optimized to correct the wrong predictions, while the decisions of unknown-class samples are inherently consistent so that the ability to identify the unknown can be well preserved.
As visualized in the bottom row of Figure~\ref{fig:intro}, our method can better separate similar objects of both the known and unknown classes by paying attention to the object areas correctly.

Main contributions of this paper can be summarized as:
\begin{itemize}
    \item We develop a novel Mutual Learning Network with the ability to identify unknown-class samples more accurately by cross-domain manifold mixup.
    To the best of our knowledge, we are the first to exploit the relation between closed-set and open-set classifiers for better identifying both known and unknown classes in the literature.
    \item We propose to learn more generalizable features by neighborhood invariance which maximizes pairwise similarities between each target sample and its neighbors to reduce the intra-class variations in the target domain.
    \item Our proposed method achieves the best results in most cases compared to the state-of-the-arts and outperforms the baseline by a large margin across all the settings in UniDA on three publicly available benchmarks.
\end{itemize}
\section{Related Work}

\paragraph{Universal domain adaptation (UniDA).}
UniDA is a practical but challenging problem that assumes no prior knowledge about the relation between the label sets of source and target domains for adaptation.
To address the issue of category-shift, early works~\cite{you2019universal,fu2020learning,saito2020universal,liang2021umad} measure the similarity of target samples with known classes and adopt a manually set threshold (e.g. with a margin) to discriminate between known and unknown classes.
DCC~\cite{li2021domain} and follow-ups~\cite{chang2022unified,chen2022mutual} utilize clustering algorithms to learn virtual prototypes for known classes and make decisions based on the distance from clusters, but are sensitive to the choices of hyperparameters.
GATE~\cite{chen2022geometric} and OneRing~\cite{yang2022one} propose $(|\mathcal{C}_s|+1)$-way universal classifiers to classify all unknown classes into the same one, while OVANet~\cite{saito2021ovanet} trains $|\mathcal{C}_s|$ one-vs-all classifiers for each known classes.
We argue that the simple entropy minimization in the OVANet to adapt inter-class distances from source domain to target domain can lead to a non-discriminative feature space and confusing predictions.
In this paper, we first present the neighborhood invariance learning to better separate different classes in the target domain, and then give details on the novel MLNet with cross-domain mixup to obtain more accurate decision boundaries.

\paragraph{Mixup scheme.}
For unsupervised domain adaptation, several works~\cite{mao2019virtual,yan2020improve,wu2020dual} take online predictions from classifiers as pseudo-labels for target samples and introduce the vanilla mixup~\cite{zhang2018mixup} within the target domain to promote the prediction stability.
To denoise incorrect pseudo-labels, a semi-supervised approach DeCoTa~\cite{yang2021deep} proposes to mix the unlabeled target sample with a labeled one from the source or target domain, such that the obtained mixups at least contain a portion of correct labels.
DACS~\cite{tranheden2021dacs} follows the idea of pseudo-labels but replaces the vanilla mixup with ClassMix~\cite{olsson2021classmix} that is tailored for semantic segmentation to combine pixels of different positions and classes.
Unfortunately, none of the above efforts can be applied to UniDA.
Because of the category-shift issue we need to simulate unknown-class samples in the target domain instead of known classes.
The existing work most relevant to our proposal is the manifold mixup~\cite{verma2019manifold} within source domain utilized in GATE~\cite{chen2022geometric}, which is also designed for the UniDA task.
We would like to highlight that the mixups obtained within source domain may fail to represent the unknown due to distribution shift.
Therefore, a novel cross-domain manifold mixup is proposed in this paper.
We derive that the mixup belongs to a certain known class with low probability, so it is beneficial to unknown-class identification in UniDA.

\section{Method}

In universal domain adaptation (UniDA), we are given a labeled source domain $\mathcal{D}_s = \{(x_i^s, y_i^s)\}_{i=1}^{N_s}$ and an unlabeled target domain $\mathcal{D}_t = \{(x_j^t)\}_{j=1}^{N_t}$.
Let $\mathcal{C}_s$ and $\mathcal{C}_t$ be the label sets of the source and target domains, respectively.
Our goal is to accurately classify target samples into either one of the known classes in $\mathcal{C}_s$ or identify them as the unknown, without prior knowledge of the relation between $\mathcal{C}_s$ and $\mathcal{C}_t$.

\subsection{Preliminaries}

The network architecture of our method is the same as the OVANet~\cite{saito2021ovanet}.
It consists of three modules: 
(1) a feature extractor ${\mathcal{F}}$ that maps RGB image of input $x$ into an embedding feature $z={\mathcal{F}}(x)\in\mathbb{R}^{D}$,  
(2) a closed-set classifier ${\mathcal{C}}$ to 
classify input $x$ into one of $K$ known classes, where $p_c(l|x)$ denotes the $l$-th class probability,
and (3) $K$ open-set classifiers ${\mathcal{O}^l}$ that outputs positive and negative scores to decide whether $x$ is from the $l$-th known class or the unknown.
For simplicity, we only record the positive score as $p_o(l|x)$ for the $l$-th sub-classifier.

\begin{figure}[t]
    \centering
    \includegraphics[width=\linewidth]{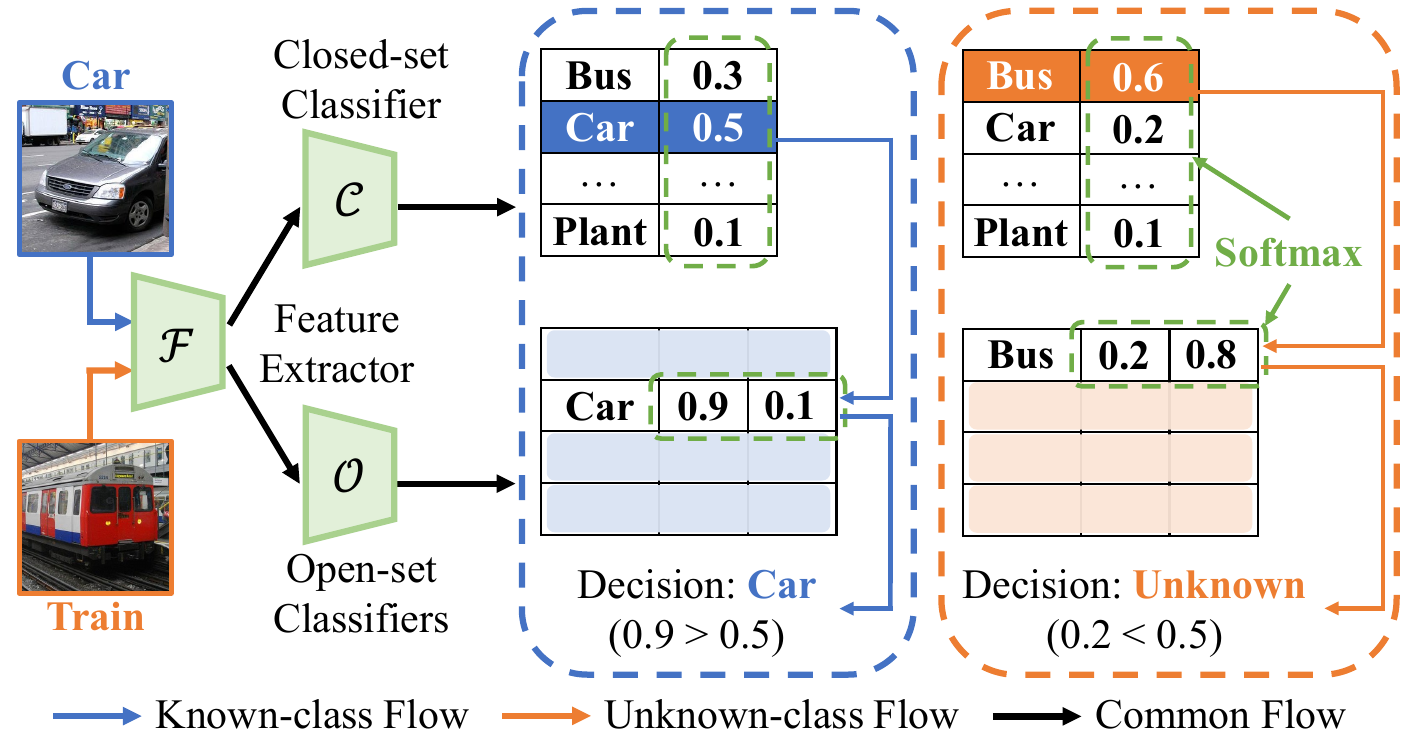}
    \caption{
    During testing, decisions are made by comparing the positive score with the decision threshold (i.e., 0.5).
    Only the open-set classifier scores of the maximum-probability known class are used to decide between known or unknown.
    }
    \label{fig:preliminary}
\end{figure}

During testing, the known class with maximum probability is first picked by the closed-set classifier.
Then, the positive score from corresponding open-set sub-classifier is used to compare with the decision threshold of 0.5 to determine known or unknown, as illustrated in Figure~\ref{fig:preliminary}.
Note that the probabilities from closed-set and open-set classifiers are normalized by Softmax function on different directions.

The training objective of the baseline OVANet can be divided by closed-set and open-set classifiers.
The closed-set classifier is trained on the labeled source domain using cross-entropy loss, denoted as $\mathcal{L}_{cls}(x_i^s, y_i^s)$.
For the open-set classifiers, \citeauthor{saito2021ovanet} first introduce hard-negative classifier sampling (HNCS) to maximise the inter-class margins in the source domain.
Then, target samples are aligned to either known or unknown class through open-set entropy minimization (OEM).
For HNCS, the one-vs-all loss is computed by a positive class and the hardest negative class, i.e.,
\begin{equation}
\label{eq:hncs}
\begin{split}
    \mathcal{L}_{ova}(x_i^s, y_i^s)=-\log p_o(y_i^s|x_i^s) & \\
    -\min_{l\ne y_i^s}\log(1-p_o(l|x_i^s)) &.
\end{split}
\end{equation}
OEM is a variant of entropy minimization~\cite{grandvalet2004semi} that performs on each sub-classifier, i.e.,
\begin{equation}
\label{eq:oem}
\begin{split}
   \mathcal{L}_{ent}(x_j^t)=-\frac{1}{K}\sum_{l=1}^K p_o(l|x_j^t) \log p_o(l|x_j^t) & \\
   + (1- p_o(l|x_j^t)) \log (1- p_o(l|x_j^t)) &.
\end{split}
\end{equation}
The overall training loss of the baseline is summarized as,
\begin{equation}
\label{eq:base}
\begin{split}
     \mathcal{L}_{base}  = & \mathop{\mathbb{E}}_{(x_i^s, y_i^s)\sim \mathcal{D}_s} [\mathcal{L}_{cls}(x_i^s,y_i^s)+\mathcal{L}_{ova}(x_i^s,y_i^s)] \\
    & + \gamma \mathop{\mathbb{E}}_{x_j^t\sim \mathcal{D}_t} \mathcal{L}_{ent}(x_j^t), 
    % & \text{where } \mathcal{L}_{src}(x_i^s,y_i^s) = \mathcal{L}_{cls}(x_i^s,y_i^s)+\mathcal{L}_{ova}(x_i^s,y_i^s),
\end{split}
\end{equation}
where the weight $\gamma$ is set to 0.1 as the same in OVANet.

\begin{figure*}[t]
    \centering
    \includegraphics[width=0.933\linewidth]{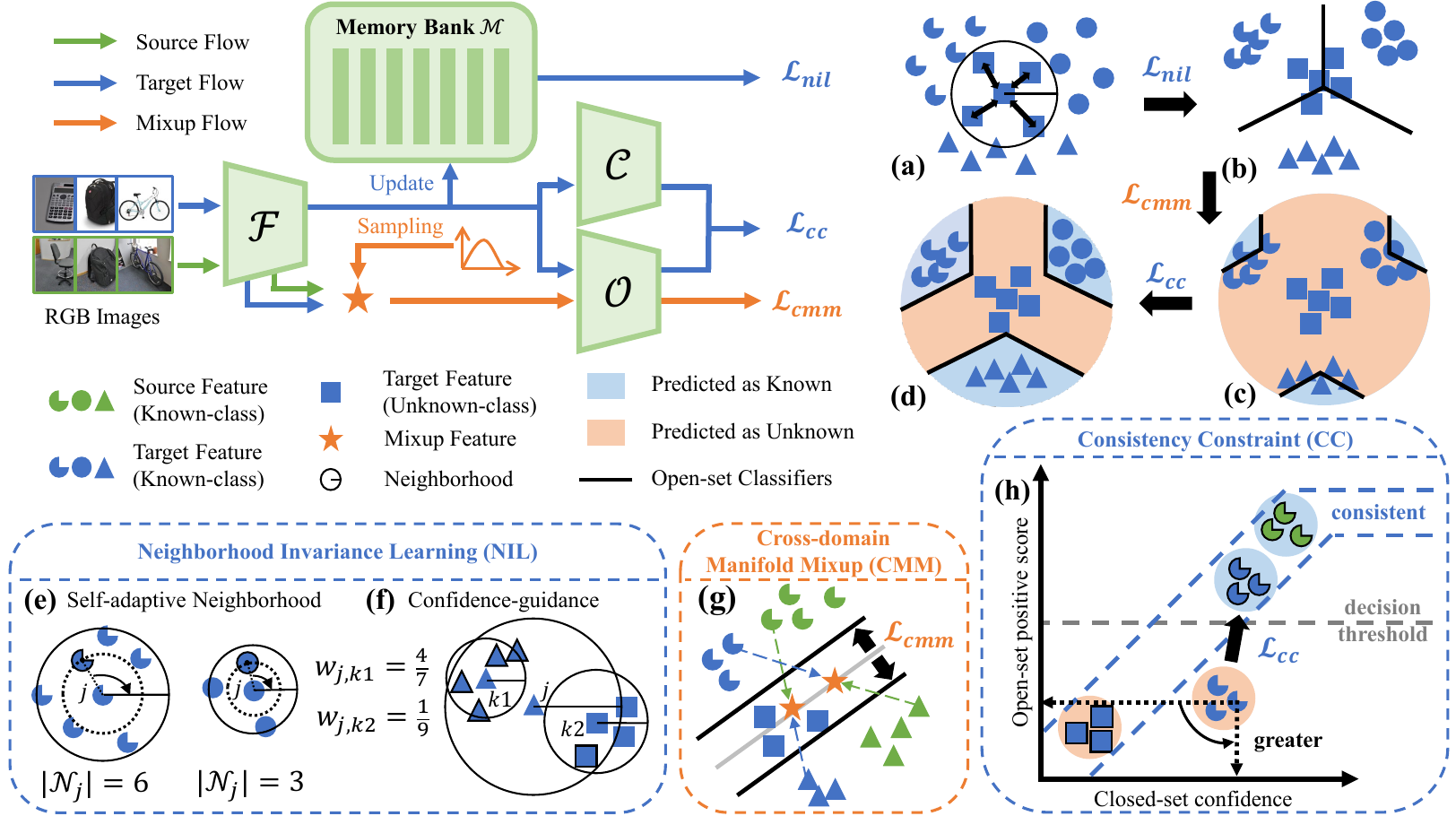}
    \caption{
    Schematics of MLNet.
    (a-d) Illustrating the effect of different losses.
    (e) Neighborhood is adaptive to the sample size of different classes.
    (f) Neighbors with more similar neighborhoods have higher confidence.
    (g) We simulate unknown-class samples across domains and utilize them to supervise the open-set classifiers for better unknown-class identification.
    (h) By optimizing $\mathcal{L}_{cc}$, misidentified known-class samples are corrected while unknown-class identification is not affected.
    } 
    \label{fig:schematic}
\end{figure*}

\subsection{Neighborhood Invariance Learning (NIL)}

Invariance learning can help to learn discriminative features for unsupervised person re-identification~\cite{zhong2019invariance,ding2020adaptive}, but has not been exploited in UniDA as far as we know.
If the feature embedding is appropriately trained on the source and target domains, target samples within a local neighborhood share the same class with high probability, which we call invariant neighborhood.
Thus, we resort to this neighborhood invariance for reducing the intra-domain variations as illustrated in Figure~\ref{fig:schematic}(a)(b), such that discriminative features can be learned to better separate different classes in the target domain.

Our method maintains a memory bank $\mathcal{M}\in \mathbb{R}^{N_t\times D}$ for features of all the target samples, where each slot $\bm{m}_{j} \in \mathbb{R}^D$ stores the feature of corresponding sample $x_j^t$. 
The memory is updated inplace every mini-batch and $l_2$-normalized, i.e.,
\begin{equation}
\label{eq:update}
    \bm{m}_j \leftarrow z_j^t / \|z_j^t\|_2, \text{ where } z_j^t = \mathcal{F}(x_j^t).
\end{equation}
Then, the pairwise similarity between the input query $x_j^t$ and the $k$-th sample in the target domain can then be obtained as,
\begin{equation}
\label{eq:similarity}
    s_{j,k}=\bm{m}_j\bm{m}_k.
\end{equation}

Following common practices, we normalize this similarity along all the target instances in the memory using Softmax function.
In this way, the memory bank is behaving like a non-parametric classifier or fully connected layer~\cite{wu2018unsupervised,zhong2019invariance}, so that 
the probability of $x_k^t$ sharing the same class with $x_j^t$ can be estimated, i.e.,
\begin{equation}
\label{eq:probability}
    p_{jk}=\frac{\exp(\tau \cdot s_{j,k})}{\sum_{n=1}^{N_t} \exp(\tau \cdot s_{j,n})},
\end{equation}
where $\tau$ is a scaling factor that modulates the sharpness of probability distribution.
After that, we reduce the intra-domain variations by maximizing the probabilities between each input sample and its neighbors.
However, the quality of neighbor search is crucial that heavily affects whether the outcome embedding space is neighborhood-invariant or not.

\paragraph{Self-adaptive Neighbor Search.}
A widely used strategy is to select $\mathcal{K}$-nearest neighbors, but it turned out to be severely limited by the sample size, i.e., the number of samples per class.
It is expensive to pick an appropriate $\mathcal{K}$-value for each dataset or even each class, considering there is also an imbalance of sample sizes within a dataset.
Inspired by the idea of self-paced learning~\cite{luo2022learning}, we turn to a more robust and adaptive approach based on the nearest neighbor, as shown in Figure~\ref{fig:schematic}(e).

Concretely, the neighborhood of input $x_j^t$ is defined as,
\begin{equation}
\label{eq:neighborhood}
    \mathcal{N}_j=\{k\ne j|s_{j,k} > \epsilon \cdot s_{j,\mathcal{N}_j[0]}\},
\end{equation}
where its range is based on the relative similarity ratio to the nearest neighbor $\mathcal{N}_j[0]$.
This design is advantageous for self-adaptation in two aspects. 
First, we find $\epsilon$ to be highly transferable which avoids setting different $\mathcal{K}$-value for different datasets.
Second, this can mitigate the imbalance of sample sizes within a dataset and promotes efficiency\footnote{Proof is provided in supplementary.}.

\paragraph{Confidence-guided Representation Learning.}
Another technique we employ to improve the robustness is assigning different weights to each neighbor in $\mathcal{N}_j$, as indicated in Figure~\ref{fig:schematic}(f).
Instead of directly computing the confidence by the feature embeddings of $x_k^t$ and $x_j^t$ as in Equations~\eqref{eq:similarity}\eqref{eq:probability}, the additional information about the neighborhood relationship between each other is utilized.
In practice, we use Jaccard distance to measure the confidence that $k$-th instance in the memory is a neighbor of input sample $x_j^t$, i.e.,
\begin{equation}
\label{eq:confidence}
    w_{jk} = \frac{|\mathcal{N}_j \cap \mathcal{N}_k|}{|\mathcal{N}_j \cup \mathcal{N}_k|}.
\end{equation}

Finally, our proposed confidence-guided invariant feature learning objective in the target domain can be written as,
\begin{equation}
\label{eq:nil}
    \mathcal{L}_{nil}(x_j^t)=-\frac{1}{|\mathcal{N}_j|}\sum_{k \in \mathcal{N}_j} w_{jk} \log p_{jk}.
\end{equation}
By minimizing the loss function~\eqref{eq:nil}, reliable neighbors with high confidence are constrained to have similar feature representations for reducing the intra-domain variations.

\subsection{Cross-domain Manifold Mixup (CMM)}

Though feature representations become more discriminative with neighborhood invariance learning, the problem of category-shift remains. 
Since only inter-class distances in the source domain are considered by HNCS, a number of target unknown-class samples would be wrongly classified as one of the known classes by OEM in the baseline.
We need to equip open-set classifiers with the ability to identify samples close to the decision boundary of known classes as unknown.
To achieve this and prevent from distribution shift, unknown-class samples are simulated by a novel cross-domain mixup scheme in a smoother way across domains based on the manifold mixup~\cite{verma2019manifold}.

Concretely, we first randomly sample an interpolation factor from a Beta distribution, i.e., $\lambda \sim Beta(\alpha, \alpha)$,
% \begin{equation}
% \label{eq:lambda}
%     \lambda \sim Beta(\alpha, \alpha),
% \end{equation}
where $\alpha$ is set to 2.0 for the Beta distribution.
The continuous feasible range of the interpolation factor $\lambda \in [0,1]$ allows our network to leverage arbitrary intermediate states of the mixups across domains to smooth the open-set classifiers.
Given the input source and target sample $x_i^s$ and $x_j^t$, the mixup feature vector across domains is obtained by,
\begin{equation}
\label{eq:mixup}
    z_{i,j,\lambda}^m = \lambda \mathcal{F}(x_i^s)+(1-\lambda)\mathcal{F}(x_j^t),
\end{equation}
as illustrated in Figure~\ref{fig:schematic}(g).
It can be proved that $z^m$ belongs to a certain known class with low probability\footnote{Proof is provided in supplementary.}, so the mixup features are good simulations of potential unknown-class samples.
We define the loss based on them to supervise the open-set classifier of corresponding source class, i.e.,
\begin{equation}
\label{eq:cmm}
    \mathcal{L}_{cmm}(x_i^s,y_i^s,x_j^t)= -\log (1-p_o(y_i^s|z_{i,j,\lambda}^m)),
\end{equation}
where $p_o(l|z)$ denotes the open-set positive score of input feature $z\in\mathbb{R}^D$ for the $l$-th class.
By minimizing $\mathcal{L}_{cmm}$, the positive scores of simulated mixup samples become smaller, such that unknown-class samples can be better identified.

\subsection{Mutual Learning via Consistency Constraint (CC)}

Despite the strong ability of CMM to identify unknown-class samples, the decision space of the known class will shrink or even degenerate into the trivial solution to identifying all the target samples as unknown.
This is because the mixups obtained using unknown-class samples in ODA and OPDA will overly squeeze the decision space for open-set classifiers to identify known classes.
As a result, the performance for known-class identification degrades significantly as illustrated in Figure~\ref{fig:schematic}(c).
To overcome this limitation, we propose to distinguish the misidentified known-class samples by the relation between the closed-set and open-set classifiers, which has never been studied in the literature.

Let's say a known-class sample $x_j^t$ with groundtruth label $l\in\mathcal{C}_s\cap \mathcal{C}_t$ is misidentified as unknown.
We observe that, it generally has high confidence for the maximum-probability known class $l$ in the closed-set classifier, but is with low positive score in the corresponding open-set sub-classifier.
Differently, for unknown-class samples, they usually have smaller closed-set confidence due to both the distribution and category shifts.
This means the closed-set confidence of an unknown-class sample is consistent with its low open-set positive score, as illustrated in Figure~\ref{fig:schematic}(h).
Hence, we can conclude that target samples with inconsistent closed-set confidence and open-set positive score are probably from the known classes misidentified as unknown.

Based on the above observations, we propose a novel consistency constraint for mutual learning between the closed-set and open-set classifiers, i.e.,
\begin{equation}
\label{eq:cc}
    \mathcal{L}_{cc}(x_j^t) = -\frac{1}{K}\sum_{l=1}^K p_c(l|x_j^t) \cdot p_o(l|x_j^t).
\end{equation}
The effect of minimizing the consistency constraint~\eqref{eq:cc} is analyzed as follows.
Since the closed-set confidence $p_c(l|x_j^t)$ is positive, the partial derivative w.r.t. the open-set positive score $p_o(l|x_j^t)$ is negative, i.e., $-p_c(l|x_j^t)/K <0$.
Therefore, the open-set positive score $p_o(l|x_j^t)$ increases by a magnitude proportional to the closed-set confidence $p_c(l|x_j^t)$ via gradient descend.
In this way, the wrong predictions for known-class samples can be corrected due to their high closed-set confidence. 
For the unknown-class samples, both $p_c(l|x_j^t)$ and $p_o(l|x_j^t)$ are consistently lower, so that the increment is slight and they will not be wrongly identified as the known-class.
Finally, the open-set classifiers and decision boundaries are optimized to better separate both the known and unknown classes as shown in Figure~\ref{fig:schematic}(d).

\begin{table*}[t]
    \centering
    \small
    \begin{tabular}{c *{20}{m{0.32cm}} c}
        \toprule
        \multirow{2}[2]{*}{\textbf{Method}} & \multicolumn{7}{c}{\textbf{Office(10/10/11)}} & \multicolumn{13}{c}{\textbf{OfficeHome(10/5/50)}} & \multirow{2}[2]{*}{\textbf{\makecell{VisDA\\(6/3/3)}}} \\
        \cmidrule(lr){2-8}\cmidrule(lr){9-21}
        & \makebox[0.32cm][c]{A2D}  & \makebox[0.32cm][c]{A2W}  & \makebox[0.32cm][c]{D2A}  & \makebox[0.32cm][c]{D2W}  & \makebox[0.32cm][c]{W2A}  & \makebox[0.32cm][c]{W2D}  & \makebox[0.32cm][c]{Avg} 
        & \makebox[0.32cm][c]{A2C}  & \makebox[0.32cm][c]{A2P}  & \makebox[0.32cm][c]{A2R}  & \makebox[0.32cm][c]{C2A}  & \makebox[0.32cm][c]{C2P}  & \makebox[0.32cm][c]{C2R}  & \makebox[0.32cm][c]{P2A}  & \makebox[0.32cm][c]{P2C}  & \makebox[0.32cm][c]{P2R}  & \makebox[0.32cm][c]{R2A}  & \makebox[0.32cm][c]{R2C}  & \makebox[0.32cm][c]{R2P}  & \makebox[0.32cm][c]{Avg} & \\
        \midrule
        UAN & \makebox[0.32cm][c]{59.7} & \makebox[0.32cm][c]{58.6} & \makebox[0.32cm][c]{60.1} & \makebox[0.32cm][c]{70.6} & \makebox[0.32cm][c]{60.3} & \makebox[0.32cm][c]{71.4} & \makebox[0.32cm][c]{63.5} & \makebox[0.32cm][c]{51.6} & \makebox[0.32cm][c]{51.7} & \makebox[0.32cm][c]{54.3} & \makebox[0.32cm][c]{61.7} & \makebox[0.32cm][c]{57.6} & \makebox[0.32cm][c]{61.9} & \makebox[0.32cm][c]{50.4} & \makebox[0.32cm][c]{47.6} & \makebox[0.32cm][c]{61.5} & \makebox[0.32cm][c]{62.9} & \makebox[0.32cm][c]{52.6} & \makebox[0.32cm][c]{65.2} & \makebox[0.32cm][c]{56.6} & \makebox[0.32cm][c]{30.5} \\
        CMU & \makebox[0.32cm][c]{68.1} & \makebox[0.32cm][c]{67.3} & \makebox[0.32cm][c]{71.4} & \makebox[0.32cm][c]{79.3} & \makebox[0.32cm][c]{72.2} & \makebox[0.32cm][c]{80.4} & \makebox[0.32cm][c]{73.1} & \makebox[0.32cm][c]{56.0} & \makebox[0.32cm][c]{56.6} & \makebox[0.32cm][c]{59.2} & \makebox[0.32cm][c]{67.0} & \makebox[0.32cm][c]{64.3} & \makebox[0.32cm][c]{67.8} & \makebox[0.32cm][c]{54.7} & \makebox[0.32cm][c]{51.1} & \makebox[0.32cm][c]{66.4} & \makebox[0.32cm][c]{68.2} & \makebox[0.32cm][c]{57.9} & \makebox[0.32cm][c]{69.7} & \makebox[0.32cm][c]{61.6} & \makebox[0.32cm][c]{34.6} \\
        DANCE & \makebox[0.32cm][c]{79.6} & \makebox[0.32cm][c]{75.8} & \makebox[0.32cm][c]{82.9} & \makebox[0.32cm][c]{90.9} & \makebox[0.32cm][c]{77.6} & \makebox[0.32cm][c]{87.1} & \makebox[0.32cm][c]{82.3} & \makebox[0.32cm][c]{61.0} & \makebox[0.32cm][c]{60.4} & \makebox[0.32cm][c]{64.9} & \makebox[0.32cm][c]{65.7} & \makebox[0.32cm][c]{58.8} & \makebox[0.32cm][c]{61.8} & \makebox[0.32cm][c]{73.1} & \makebox[0.32cm][c]{61.2} & \makebox[0.32cm][c]{66.6} & \makebox[0.32cm][c]{67.7} & \makebox[0.32cm][c]{62.4} & \makebox[0.32cm][c]{63.7} & \makebox[0.32cm][c]{63.9} & \makebox[0.32cm][c]{42.8} \\
        DCC & \makebox[0.32cm][c]{88.5} & \makebox[0.32cm][c]{78.5} & \makebox[0.32cm][c]{70.2} & \makebox[0.32cm][c]{79.3} & \makebox[0.32cm][c]{75.9} & \makebox[0.32cm][c]{88.6} & \makebox[0.32cm][c]{80.2} & \makebox[0.32cm][c]{58.0} & \makebox[0.32cm][c]{54.1} & \makebox[0.32cm][c]{58.0} & \makebox[0.32cm][c]{74.6} & \makebox[0.32cm][c]{70.6} & \makebox[0.32cm][c]{77.5} & \makebox[0.32cm][c]{64.3} & \makebox[0.32cm][c]{73.6} & \makebox[0.32cm][c]{74.9} & \makebox[0.32cm][c]{81.0} & \makebox[0.32cm][c]{75.1} & \makebox[0.32cm][c]{80.4} & \makebox[0.32cm][c]{70.2} & \makebox[0.32cm][c]{43.0} \\
        GATE & \makebox[0.32cm][c]{87.7} & \makebox[0.32cm][c]{81.6} & \makebox[0.32cm][c]{84.2} & \makebox[0.32cm][c]{94.8} & \makebox[0.32cm][c]{83.4} & \makebox[0.32cm][c]{94.1} & \makebox[0.32cm][c]{87.6} & \makebox[0.32cm][c]{63.8} & \makebox[0.32cm][c]{75.9} & \makebox[0.32cm][c]{81.4} & \makebox[0.32cm][c]{74.0} & \makebox[0.32cm][c]{72.1} & \makebox[0.32cm][c]{79.8} & \makebox[0.32cm][c]{74.7} & \makebox[0.32cm][c]{70.3} & \makebox[0.32cm][c]{82.7} & \makebox[0.32cm][c]{79.1} & \makebox[0.32cm][c]{71.5} & \makebox[0.32cm][c]{81.7} & \makebox[0.32cm][c]{75.6} & \makebox[0.32cm][c]{56.4} \\
        TNT & \makebox[0.32cm][c]{85.7} & \makebox[0.32cm][c]{80.4} & \makebox[0.32cm][c]{83.8} & \makebox[0.32cm][c]{92.0} & \makebox[0.32cm][c]{79.1} & \makebox[0.32cm][c]{91.2} & \makebox[0.32cm][c]{85.4} & \makebox[0.32cm][c]{61.9} & \makebox[0.32cm][c]{74.6} & \makebox[0.32cm][c]{80.2} & \makebox[0.32cm][c]{73.5} & \makebox[0.32cm][c]{71.4} & \makebox[0.32cm][c]{79.6} & \makebox[0.32cm][c]{74.2} & \makebox[0.32cm][c]{69.5} & \makebox[0.32cm][c]{82.7} & \makebox[0.32cm][c]{77.3} & \makebox[0.32cm][c]{70.1} & \makebox[0.32cm][c]{81.2} & \makebox[0.32cm][c]{74.7} & \makebox[0.32cm][c]{55.3} \\
        UniOT & \makebox[0.32cm][c]{87.0} & \makebox[0.32cm][c]{88.5} & \makebox[0.32cm][c]{88.4} & \makebox[0.32cm][c]{98.8} & \makebox[0.32cm][c]{87.6} & \makebox[0.32cm][c]{96.6} & \makebox[0.32cm][c]{91.2} & \makebox[0.32cm][c]{67.3} & \makebox[0.32cm][c]{80.5} & \makebox[0.32cm][c]{86.0} & \makebox[0.32cm][c]{73.5} & \makebox[0.32cm][c]{77.3} & \makebox[0.32cm][c]{84.3} & \makebox[0.32cm][c]{75.5} & \makebox[0.32cm][c]{63.3} & \makebox[0.32cm][c]{86.0} & \makebox[0.32cm][c]{77.8} & \makebox[0.32cm][c]{65.4} & \makebox[0.32cm][c]{81.9} & \makebox[0.32cm][c]{76.6} & \makebox[0.32cm][c]{57.3} \\
        CPR & \makebox[0.32cm][c]{84.4} & \makebox[0.32cm][c]{81.4} & \makebox[0.32cm][c]{85.5} & \makebox[0.32cm][c]{93.4} & \makebox[0.32cm][c]{91.3} & \makebox[0.32cm][c]{96.8} & \makebox[0.32cm][c]{88.8} & \makebox[0.32cm][c]{59.0} & \makebox[0.32cm][c]{77.1} & \makebox[0.32cm][c]{83.7} & \makebox[0.32cm][c]{69.7} & \makebox[0.32cm][c]{68.1} & \makebox[0.32cm][c]{75.4} & \makebox[0.32cm][c]{74.6} & \makebox[0.32cm][c]{56.1} & \makebox[0.32cm][c]{78.9} & \makebox[0.32cm][c]{80.5} & \makebox[0.32cm][c]{63.0} & \makebox[0.32cm][c]{81.0} & \makebox[0.32cm][c]{72.3} & \makebox[0.32cm][c]{58.2} \\
        NCAL & \makebox[0.32cm][c]{85.3} & \makebox[0.32cm][c]{85.3} & \makebox[0.32cm][c]{88.0} & \makebox[0.32cm][c]{94.0} & \makebox[0.32cm][c]{87.9} & \makebox[0.32cm][c]{95.5} & \makebox[0.32cm][c]{89.3} & \makebox[0.32cm][c]{59.1} & \makebox[0.32cm][c]{88.3} & \makebox[0.32cm][c]{87.3} & \makebox[0.32cm][c]{72.1} & \makebox[0.32cm][c]{73.2} & \makebox[0.32cm][c]{81.0} & \makebox[0.32cm][c]{76.3} & \makebox[0.32cm][c]{57.4} & \makebox[0.32cm][c]{88.4} & \makebox[0.32cm][c]{81.1} & \makebox[0.32cm][c]{62.0} & \makebox[0.32cm][c]{85.4} & \makebox[0.32cm][c]{75.9} & \makebox[0.32cm][c]{62.9} \\
        \midrule
        Baseline & \makebox[0.32cm][c]{83.9} & \makebox[0.32cm][c]{78.5} & \makebox[0.32cm][c]{81.3} & \makebox[0.32cm][c]{95.4} & \makebox[0.32cm][c]{83.2} & \makebox[0.32cm][c]{96.4} & \makebox[0.32cm][c]{86.5} & \makebox[0.32cm][c]{62.2} & \makebox[0.32cm][c]{79.0} & \makebox[0.32cm][c]{80.0} & \makebox[0.32cm][c]{69.2} & \makebox[0.32cm][c]{70.5} & \makebox[0.32cm][c]{76.5} & \makebox[0.32cm][c]{70.9} & \makebox[0.32cm][c]{59.5} & \makebox[0.32cm][c]{80.9} & \makebox[0.32cm][c]{76.8} & \makebox[0.32cm][c]{62.8} & \makebox[0.32cm][c]{79.6} & \makebox[0.32cm][c]{72.3} & \makebox[0.32cm][c]{49.6} \\
        Ours & \makebox[0.32cm][c]{90.4} & \makebox[0.32cm][c]{93.7} & \makebox[0.32cm][c]{89.7} & \makebox[0.32cm][c]{96.2} & \makebox[0.32cm][c]{88.4} & \makebox[0.32cm][c]{98.3} & \makebox[0.32cm][c]{\textbf{92.8}} & \makebox[0.32cm][c]{68.2} & \makebox[0.32cm][c]{83.8} & \makebox[0.32cm][c]{85.0} & \makebox[0.32cm][c]{73.6} & \makebox[0.32cm][c]{78.2} & \makebox[0.32cm][c]{82.2} & \makebox[0.32cm][c]{75.2} & \makebox[0.32cm][c]{64.7} & \makebox[0.32cm][c]{85.1} & \makebox[0.32cm][c]{78.8} & \makebox[0.32cm][c]{69.9} & \makebox[0.32cm][c]{83.9} & \makebox[0.32cm][c]{\textbf{77.4}} & \makebox[0.32cm][c]{\textbf{69.9}} \\
        \bottomrule
    \end{tabular}
    \caption{H-score comparison in the OPDA setting. Some results are referred to previous work~\cite{chen2022geometric}. Due to page limit, the comparisons with state-of-the-arts under other settings (i.e., CDA, ODA and PDA) are given in the supplementary.}
    \label{tab:opda}
\end{table*}
\begin{table}[t]
    \centering
    \small
    \begin{tabular}{ccm{0.42cm}m{0.42cm}m{0.42cm}m{0.42cm}m{0.42cm}m{0.42cm}}
        \toprule
        \multirow{2}[2]{*}{\textbf{Dataset}} & \multirow{2}[2]{*}{\textbf{Method}} & \multicolumn{3}{c}{\textbf{Accuracy}} & \multicolumn{3}{c}{\textbf{H-Score}} \\
        \cmidrule(lr){3-5}\cmidrule(lr){6-8}
        & & \makebox[0.42cm][c]{CDA} & \makebox[0.42cm][c]{PDA} & \makebox[0.42cm][c]{Avg} & \makebox[0.42cm][c]{ODA} & \makebox[0.42cm][c]{OPDA} & \makebox[0.42cm][c]{Avg} \\
        \midrule
        \multirow{3}{*}{Office} 
        & Baseline & \makebox[0.42cm][c]{82.7} & \makebox[0.42cm][c]{90.0} & \makebox[0.42cm][c]{86.4} & \makebox[0.42cm][c]{91.6} & \makebox[0.42cm][c]{86.5} & \makebox[0.42cm][c]{89.1} \\
        & Ours  & \makebox[0.42cm][c]{\textbf{86.9}} & \makebox[0.42cm][c]{\textbf{94.4}} & \makebox[0.42cm][c]{\textbf{90.7}} & \makebox[0.42cm][c]{\textbf{92.8}} & \makebox[0.42cm][c]{\textbf{92.8}} & \makebox[0.42cm][c]{\textbf{92.8}} \\
        & (Inc.) & \makebox[0.42cm][c]{+4.2} & \makebox[0.42cm][c]{+4.4} & \makebox[0.42cm][c]{+4.3} & \makebox[0.42cm][c]{+1.2} & \makebox[0.42cm][c]{+6.3} & \makebox[0.42cm][c]{+3.7} \\
        \midrule
        \multirow{3}{*}{\makecell[c]{Office\\Home}} 
        & Baseline & \makebox[0.42cm][c]{63.9} & \makebox[0.42cm][c]{65.1} & \makebox[0.42cm][c]{64.5} & \makebox[0.42cm][c]{64.1} & \makebox[0.42cm][c]{72.3} & \makebox[0.42cm][c]{68.2} \\
        & Ours & \makebox[0.42cm][c]{\textbf{69.2}} & \makebox[0.42cm][c]{\textbf{78.1}} & \makebox[0.42cm][c]{\textbf{73.7}} & \makebox[0.42cm][c]{\textbf{67.0}} & \makebox[0.42cm][c]{\textbf{77.4}}  & \makebox[0.42cm][c]{\textbf{72.2}} \\
        & (Inc.) & \makebox[0.42cm][c]{+5.3} & \makebox[0.42cm][c]{+13.0} & \makebox[0.42cm][c]{+9.2} & \makebox[0.42cm][c]{+2.9} & \makebox[0.42cm][c]{+5.1} & \makebox[0.42cm][c]{+4.0} \\
        \midrule
        \multirow{3}{*}{VisDA} 
        & Baseline & \makebox[0.42cm][c]{58.5} & \makebox[0.42cm][c]{60.8} & \makebox[0.42cm][c]{59.7} & \makebox[0.42cm][c]{61.6} & \makebox[0.42cm][c]{49.6} & \makebox[0.42cm][c]{55.6} \\
        & Ours & \makebox[0.42cm][c]{\textbf{73.0}} & \makebox[0.42cm][c]{\textbf{80.4}} & \makebox[0.42cm][c]{\textbf{76.7}} & \makebox[0.42cm][c]{\textbf{63.9}} & \makebox[0.42cm][c]{\textbf{69.9}}  & \makebox[0.42cm][c]{\textbf{66.9}} \\
        & (Inc.) & \makebox[0.42cm][c]{+14.5} & \makebox[0.42cm][c]{+19.6} & \makebox[0.42cm][c]{+17.0} & \makebox[0.42cm][c]{+2.3} & \makebox[0.42cm][c]{+20.3} & \makebox[0.42cm][c]{+11.3} \\
        \bottomrule
    \end{tabular}
    \caption{Overall comparison against baseline. Average performances count for average classification accuracy in CDA and PDA, and H-score in ODA and OPDA.}
    \label{tab:overall}
\end{table}
\subsection{Optimization}

The overall training objective of our method is given by,
\begin{equation}
\label{eq:all}
\begin{split}
    \mathcal{L}_{all} = \mathcal{L}_{base} + \mathbb{E}_{\substack{(x_i^s,y_i^s,x_j^t)\\ \sim(\mathcal{D}_s\cup\mathcal{D}_t)}} [\beta_1\mathcal{L}_{nil}(x_j^t) & \\
    +\beta_2\mathcal{L}_{cmm}(x_i^s,y_i^s,x_j^t) + \eta \mathcal{L}_{cc} (x_j^t)] & , 
\end{split}
\end{equation}
where $\beta_1, \beta_2, \eta$ are weights that trade off different losses.
They are determined by experiments on Office~\cite{saenko2010adapting} and generalized to others.
In our experiments, we fix $\beta_1$ and perform a grid search on the values of $\beta_2$ and $\eta$.
\section{Experiments}

\subsection{Experimental Setup}

\noindent\textbf{Datasets.}
We conduct experiments on three UniDA benchmarks including Office~\cite{saenko2010adapting}, OfficeHome~\cite{venkateswara2017deep} and VisDA~\cite{peng2017visda}.
Office consists of 31 categories and contains around 4,700 images across three domains, namely Amazon(A), DSLR(D) and Webcam(W).
OfficeHome includes 15,500 images of 65 categories across four domains, while VisDA is a large-scale dataset that contains around 150,000 synthetic and 50,000 real-world images of 12 categories.
We use $(|\mathcal{C}_s\cap\mathcal{C}_t|/|\mathcal{C}_s-\mathcal{C}_t|/|\mathcal{C}_t-\mathcal{C}_s|)$ to represent the split for datasets. 
For more details, please refer to the supplementary.

\noindent\textbf{Evaluation protocols.}
Following previous studies~\cite{li2021domain,chen2022geometric}, we use the same evaluation protocols as those used in different settings.
In CDA and PDA, we calculate the classification accuracy over all target samples as the performance metric.
In ODA and OPDA, we compute the H-score to account for the trade-off between known-class and unknown-class identification, i.e. the harmonic mean of the average of per-class accuracies on known classes and accuracy on the unknown class.

\noindent\textbf{Implementation details.}
We adopt the same network architecture and optimization setting as the baseline.
The feature extractor $\mathcal{F}$ is a ResNet-50~\cite{he2016deep} pretrained on ImageNet~\cite{deng2009imagenet}.
% Both closed-set and open-set classifiers are a fully connected layer.
The SGD optimizer with nesterov momentum and inverse scheduler is employed and the batch size is set to 36.
The learning rates for the feature extractor $\mathcal{F}$ and all the other components are 0.001 and 0.01, respectively.
We train 50 epochs for all datasets and settings, and $\mathcal{L}_{nil}$ is disabled in the first epoch.
The scaling factor $\tau$ is set to 10.0 as the same in previous work~\cite{luo2022learning}.
The neighborhood relative ratio $\epsilon=0.875$ and loss weights $\beta_1=0.5$, $\beta_2=0.1$ are fixed across all datasets and settings.
Through extensive experiments, we find that the difficulty in achieving consistency is related to the sample size as detailed in ablation studies, so $\eta$ is set to 0.16 by default and doubled on VisDA.

\subsection{Comparison with State-of-the-arts}

\begin{figure*}[t]
    \centering
    \subfloat{\includegraphics[height=2.7cm]{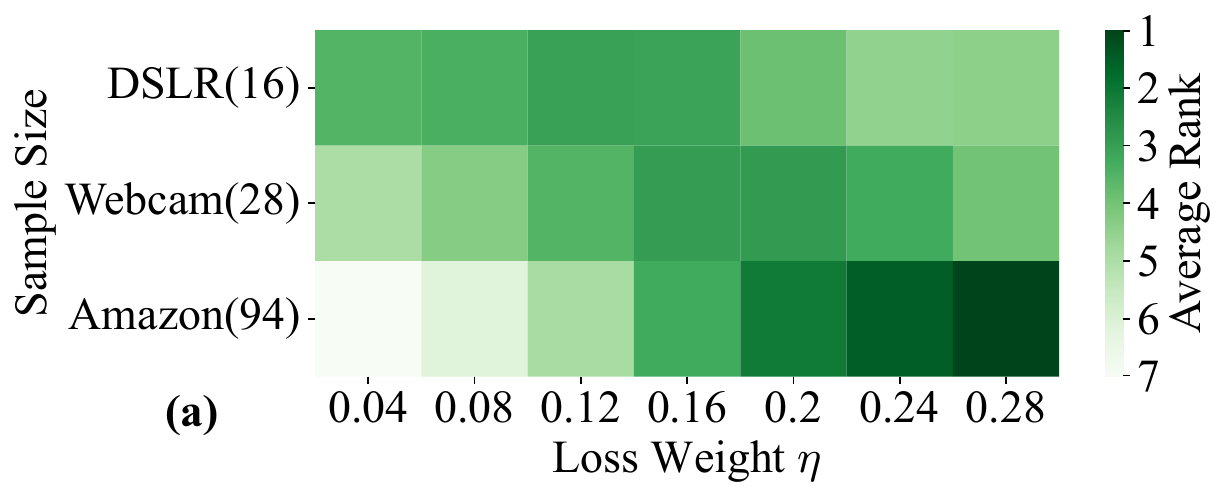}}
    \subfloat{\includegraphics[height=3cm]{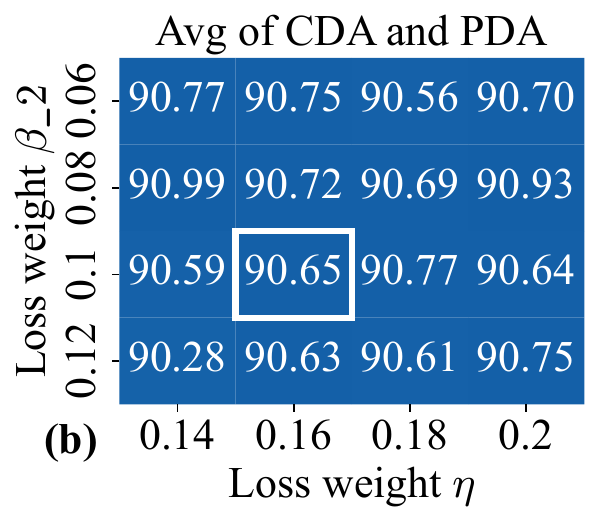}}
    \subfloat{\includegraphics[height=3cm]{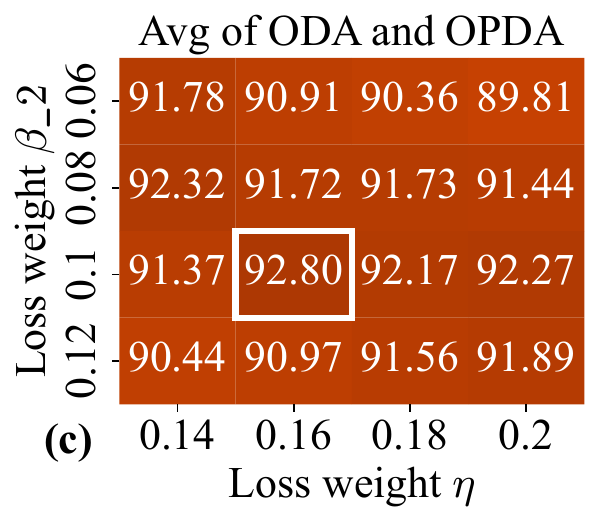}}
    \subfloat{\includegraphics[height=2.9cm]{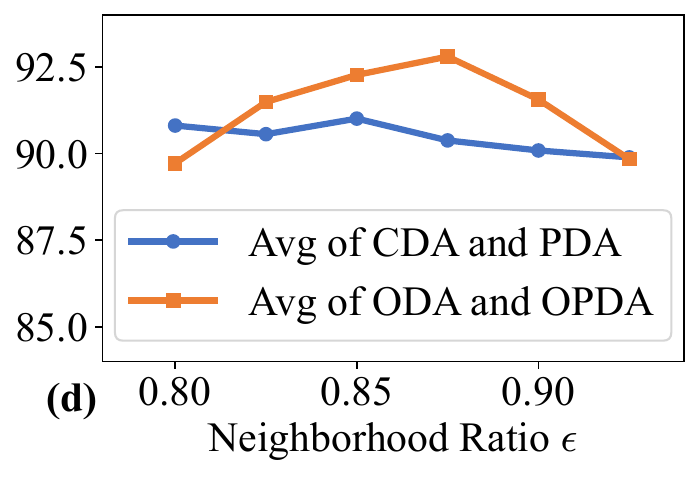}}
    \caption{Ablation studies for (a) consistency constraint and (b-d) hyperparameter sensitivity on the Office dataset. 
    (a) Each row collects results for all settings with a specific target domain (e.g., Amazon with sample size of 94) when varying the loss weight $\eta$.
    A higher average ranking (up to 1) indicates higher performance across the 8 experiments (4 DA settings and 2 sources).
    }
    \label{fig:sensitivity}
\end{figure*}

We compare MLNet with state-of-the-arts across all settings, including UAN~\cite{you2019universal}, CMU~\cite{fu2020learning}, DANCE~\cite{saito2020universal}, DCC~\cite{li2021domain}, OVANet~\cite{saito2021ovanet}, GATE~\cite{chen2022geometric}, MATHS~\cite{chen2022mutual}, TNT~\cite{chen2022evidential}, UniOT~\cite{chang2022unified}, CPR~\cite{hur2023learning} and NCAL~\cite{su2023neighborhood}.
For fair comparison, we reproduce the baseline (i.e., OVANet) in the same environment.

\noindent\textbf{Overall comparison.}
Table~\ref{tab:overall} presents the overall comparison against the baseline across all settings.
These results show that the substantial performance improvement of our proposed MLNet over baseline.
Especially in the challenging VisDA dataset, the average improvements are 17.0\% and 11.3\% for the accuracy and H-score, respectively.

\noindent\textbf{OPDA and ODA settings.}
From Table~\ref{tab:opda}, we can see that MLNet achieves new state-of-the-art results in the challenging OPDA setting across all datasets.
When compared to UniOT which is tailored for OPDA, our method still outperforms it by around 1\% on Office and OfficeHome, while the improvement is incredible 12.6\% on VisDA.
The results in the supplementary for ODA show that our method achieves competitive results on OfficeHome and VisDA, and constantly gives the highest H-score on Office.

\noindent\textbf{PDA and CDA settings.}
In the supplementary, the comparison under the PDA setting shows that MLNet outperforms the state-of-the-art methods averagely by 4.9\% and 4.8\% on the OfficeHome and VisDA datasets, respectively.
For the CDA setting, results show that our method achieves competitive performance compared to existing works.

\begin{table}[t]
    \centering
    \small
    \begin{tabular}{m{2.2cm}*{8}{m{0.3cm}}}
        \toprule
        \multirow{2}[2]{*}{\makebox[2.2cm][l]{\textbf{Method}}} & \multirow{2}[2]{*}{\makebox[0.3cm][c]{\textbf{CDA}}} & \multicolumn{3}{c}{\textbf{ODA}} & \multicolumn{3}{c}{\textbf{OPDA}} & \multirow{2}[2]{*}{\makebox[0.3cm][c]{\textbf{PDA}}} \\
        \cmidrule(lr){3-5}\cmidrule(lr){6-8}
        & & \makebox[0.3cm][c]{\emph{K}} & \makebox[0.3cm][c]{\emph{U}} & \makebox[0.3cm][c]{\emph{H}} & \makebox[0.3cm][c]{\emph{K}} & \makebox[0.3cm][c]{\emph{U}} & \makebox[0.3cm][c]{\emph{H}} & \\
        \midrule
         \makebox[2.2cm][l]{Ours w/o NIL}  & \makebox[0.3cm][c]{79.5} & \makebox[0.3cm][c]{11.9} & \makebox[0.3cm][c]{100.0} & \makebox[0.3cm][c]{17.1} & \makebox[0.3cm][c]{27.1} & \makebox[0.3cm][c]{99.3} & \makebox[0.3cm][c]{35.3} & \makebox[0.3cm][c]{85.9} \\
         \makebox[2.2cm][l]{Ours w/o CMM}  & \makebox[0.3cm][c]{85.7} & \makebox[0.3cm][c]{97.4} & \makebox[0.3cm][c]{27.2}  & \makebox[0.3cm][c]{41.3} & \makebox[0.3cm][c]{96.3} & \makebox[0.3cm][c]{34.6} & \makebox[0.3cm][c]{50.2} & \makebox[0.3cm][c]{93.0} \\
         \makebox[2.2cm][l]{Ours w/o CC}   & \makebox[0.3cm][c]{86.7} & \makebox[0.3cm][c]{3.9}  & \makebox[0.3cm][c]{99.5}  & \makebox[0.3cm][c]{6.9}  & \makebox[0.3cm][c]{64.4} & \makebox[0.3cm][c]{96.5} & \makebox[0.3cm][c]{75.7} & \makebox[0.3cm][c]{94.1} \\
         \makebox[2.2cm][l]{Ours w/o Conf.}& \makebox[0.3cm][c]{86.5} & \makebox[0.3cm][c]{88.7} & \makebox[0.3cm][c]{93.3}  & \makebox[0.3cm][c]{90.8} & \makebox[0.3cm][c]{93.1} & \makebox[0.3cm][c]{88.5} & \makebox[0.3cm][c]{90.8} & \makebox[0.3cm][c]{91.9} \\
         \makebox[2.2cm][l]{Ours w/ KNN}   & \makebox[0.3cm][c]{85.3} & \makebox[0.3cm][c]{51.6} & \makebox[0.3cm][c]{98.6}  & \makebox[0.3cm][c]{62.1} & \makebox[0.3cm][c]{68.4} & \makebox[0.3cm][c]{93.3} & \makebox[0.3cm][c]{77.0} & \makebox[0.3cm][c]{79.7} \\
         \makebox[2.2cm][l]{Ours w/ SMM}   & \makebox[0.3cm][c]{85.4} & \makebox[0.3cm][c]{97.0} & \makebox[0.3cm][c]{66.2}  & \makebox[0.3cm][c]{77.2} & \makebox[0.3cm][c]{95.5} & \makebox[0.3cm][c]{59.2} & \makebox[0.3cm][c]{72.7} & \makebox[0.3cm][c]{92.9} \\
         \makebox[2.2cm][l]{Ours (full)}   & \makebox[0.3cm][c]{86.9} & \makebox[0.3cm][c]{89.6} & \makebox[0.3cm][c]{95.9}  & \makebox[0.3cm][c]{92.8} & \makebox[0.3cm][c]{92.3} & \makebox[0.3cm][c]{93.5} & \makebox[0.3cm][c]{92.8} & \makebox[0.3cm][c]{94.4} \\
        \bottomrule
    \end{tabular}
    \caption{Ablation studies of different components on the Office dataset. Note that \emph{K, U, H} refer to the known accuracy, unknown accuracy and H-score, respectively. KNN and SMM are short for $\mathcal{K}$-nearest neighbors and source-domain manifold mixup~\cite{chen2022geometric}, respectively.}
    \label{tab:ablation} 
\end{table}

\subsection{Ablation Studies}

In Table~\ref{tab:ablation}, we present the ablation studies on the Office dataset across all settings.
Since H-score is a harmonic mean sensitive to extreme values, it is only high when both accuracies are high.
Therefore, removing any loss function (Rows 1-3) will result in a severe imbalance of trade-off between known-class and unknown-class identification.

\noindent\textbf{Effect of Confidence-guidance.}
We also try removing the confidence $w_{ij}$ in Equation~\eqref{eq:nil},
where the results show a slight decrease in the performance of different settings (Row 4), suggesting that exploiting the relationship between neighborhoods is helpful in improving the robustness.

\noindent\textbf{Effect of Self-adaptive Neighborhood.}
In Row 5, we instead use $\mathcal{K}$-nearest neighbors to determine neighborhood, where $\mathcal{K}$ is set to 5 as the same in GATE and the performance degrades severely when category-shift exists, which validates the superiority of our self-adaptive approach.

\noindent\textbf{Effect of Cross-domain Mixup.}
To evaluate the contribution of our proposed CMM, we conduct ablation experiments using source-domain manifold mixup as utilized in GATE (Row 6).
The results indicate that our cross-domain scheme is more effective since we can leverage arbitrary intermediate states across domains to smooth the transfer.

\begin{table}[t]
    \centering
    \small
    \begin{tabular}{l*{6}{m{0.45cm}}}
        \toprule
        \multirow{2}[2]{*}{\textbf{Method}} & \multicolumn{3}{c}{\textbf{Accuracy}} & \multicolumn{3}{c}{\textbf{H-Score}} \\
        \cmidrule(lr){2-4}\cmidrule(lr){5-7}
        & \makebox[0.45cm][c]{CDA} & \makebox[0.45cm][c]{PDA} & \makebox[0.45cm][c]{Avg} & \makebox[0.45cm][c]{ODA} & \makebox[0.45cm][c]{OPDA} & \makebox[0.45cm][c]{Avg} \\
        \midrule
        Baseline    & \makebox[0.45cm][c]{82.7} & \makebox[0.45cm][c]{90.0} & \makebox[0.45cm][c]{86.4} & \makebox[0.45cm][c]{91.6} & \makebox[0.45cm][c]{86.5} & \makebox[0.45cm][c]{89.1} \\
        CC via Cosine        & \makebox[0.45cm][c]{86.7} & \makebox[0.45cm][c]{94.8} & \makebox[0.45cm][c]{90.8} & \makebox[0.45cm][c]{85.9} & \makebox[0.45cm][c]{84.9} & \makebox[0.45cm][c]{85.4} \\
        CC via JS-Div        & \makebox[0.45cm][c]{86.6} & \makebox[0.45cm][c]{94.5} & \makebox[0.45cm][c]{90.6} & \makebox[0.45cm][c]{86.2} & \makebox[0.45cm][c]{90.4} & \makebox[0.45cm][c]{88.3} \\
        CC via KL-Div (C)    & \makebox[0.45cm][c]{86.5} & \makebox[0.45cm][c]{95.1} & \makebox[0.45cm][c]{90.8} & \makebox[0.45cm][c]{87.7} & \makebox[0.45cm][c]{90.8} & \makebox[0.45cm][c]{89.3} \\
        CC via KL-Div (O)    & \makebox[0.45cm][c]{86.3} & \makebox[0.45cm][c]{94.8} & \makebox[0.45cm][c]{90.6} & \makebox[0.45cm][c]{86.9} & \makebox[0.45cm][c]{89.2} & \makebox[0.45cm][c]{88.1} \\
        Ours                & \makebox[0.45cm][c]{86.9} & \makebox[0.45cm][c]{94.4} & \makebox[0.45cm][c]{90.7} & \makebox[0.45cm][c]{92.8} & \makebox[0.45cm][c]{92.8} & \makebox[0.45cm][c]{92.8} \\
        \bottomrule
    \end{tabular}
    \caption{Ablation studies of $\mathcal{L}_{cc}$ on the Office dataset. 
    We normalize open-set positive scores across all classes using Softmax to obtain probability distribution for methods in Rows 3-5. 
    Since $D_{KL}(P\|Q)$ is asymmetric, we refer to the case when $P$ denotes closed-set probability and normalized open-set positive scores as KL-Div (C) and (O), respectively.
    }
    \label{tab:cc}
\end{table}

\noindent\textbf{Effect of Consistency Constraint.}
Table~\ref{tab:cc} presents the overall comparison of different choices on $\mathcal{L}_{cc}$, including cosine distance, JS and KL divergence.
We can observe that bridging the closed-set and open-set classifiers is indeed beneficial (Rows 2-5) compared to the baseline in CDA and PDA.
Only our $\mathcal{L}_{cc}$ can further mitigate the inconsistency between known-class samples and increase the H-score in ODA and OPDA (Row 6).
In Figure~\ref{fig:sensitivity}(a), we collect the average rank versus sample sizes across different settings when varying the weight of consistency constraint $\eta$.
We observe that the larger the sample size, the greater the weights to achieve the better results.
To ensure the robustness of our method, we follow the practice in previous works~\cite{chen2022evidential,chen2022geometric,chang2022unified} to fix the hyperparameter across different settings and datasets by default and simply double it in the large-scale VisDA for better performance.

\noindent\textbf{Hyperparameter Sensitivity.}
This experiment is performed with $\beta_1=0.5$. More results on $\beta_1=0.1$ and $1.0$ are provided in the supplementary.
In Figure~\ref{fig:sensitivity}(b)(c), we conduct a grid search on the loss weights of $\beta_2,\eta$, where the accuracies in CDA and PDA keep almost unchanged.
While the average performance of ODA and OPDA is also stable, the H-score is sensitive to the trade-off between known and unknown classes.
Figure~\ref{fig:sensitivity}(d) presents the experiments on varying neighborhood ratio $\epsilon$.
Similarly, both accuracy and H-score are only slightly varied by less than 1\% and 3\%, respectively, which shows the robustness of our method.
\section{Conclusion}

In this paper, we propose a novel Mutual Learning Network for universal domain adaptation by incorporating neighborhood invariance and cross-domain mixup.
To trade off the misidentified known-class samples, a novel consistency constraint is employed between the closed-set and open-set classifiers.
Extensive experiments validate that our method outperforms the baseline significantly and achieve state-of-the-art results in most cases across all settings in UniDA.

\section{Acknowledgments}
This work was supported in part by the National Natural Science Foundation of China (U22A2095, 62276281) and the Science and Technology Program of Guangzhou (202002030371).

\bibliography{aaai24}

\begin{thebibliography}{46}
\providecommand{\natexlab}[1]{#1}

\bibitem[{Busto and Gall(2017)}]{busto2017open}
Busto, P.~P.; and Gall, J. 2017.
\newblock Open Set Domain Adaptation.
\newblock In \emph{ICCV}, 754–763.

\bibitem[{Cao et~al.(2018)Cao, Ma, Long, and Wang}]{cao2018partial}
Cao, Z.; Ma, L.; Long, M.; and Wang, J. 2018.
\newblock Partial Adversarial Domain Adaptation.
\newblock In \emph{ECCV}, 135–150.

\bibitem[{Chang et~al.(2022)Chang, Shi, Tuan, and Wang}]{chang2022unified}
Chang, W.; Shi, Y.; Tuan, H.; and Wang, J. 2022.
\newblock Unified Optimal Transport Framework for Universal Domain Adaptation.
\newblock In \emph{NeurIPS}.

\bibitem[{Chen et~al.(2022{\natexlab{a}})Chen, Chen, Wei, Jin, Tan, Jin, and Chen}]{chen2022reusing}
Chen, L.; Chen, H.; Wei, Z.; Jin, X.; Tan, X.; Jin, Y.; and Chen, E. 2022{\natexlab{a}}.
\newblock Reusing the Task-specific Classifier as a Discriminator: Discriminator-free Adversarial Domain Adaptation.
\newblock In \emph{CVPR}, 7181–7190.

\bibitem[{Chen et~al.(2022{\natexlab{b}})Chen, Du, Lou, He, Bai, and Deng}]{chen2022mutual}
Chen, L.; Du, Q.; Lou, Y.; He, J.; Bai, T.; and Deng, M. 2022{\natexlab{b}}.
\newblock Mutual Nearest Neighbor Contrast and Hybrid Prototype Self-Training for Universal Domain Adaptation.
\newblock \emph{AAAI}, 36(6): 6248–6257.

\bibitem[{Chen et~al.(2022{\natexlab{c}})Chen, Lou, He, Bai, and Deng}]{chen2022evidential}
Chen, L.; Lou, Y.; He, J.; Bai, T.; and Deng, M. 2022{\natexlab{c}}.
\newblock Evidential Neighborhood Contrastive Learning for Universal Domain Adaptation.
\newblock \emph{AAAI}, 36(6): 6258–6267.

\bibitem[{Chen et~al.(2022{\natexlab{d}})Chen, Lou, He, Bai, and Deng}]{chen2022geometric}
Chen, L.; Lou, Y.; He, J.; Bai, T.; and Deng, M. 2022{\natexlab{d}}.
\newblock Geometric Anchor Correspondence Mining with Uncertainty Modeling for Universal Domain Adaptation.
\newblock In \emph{CVPR}, 16134–16143.

\bibitem[{Deng and Jia(2023)}]{deng2023universal}
Deng, B.; and Jia, K. 2023.
\newblock Universal Domain Adaptation from Foundation Models.
\newblock \emph{arXiv:2305.11092}.

\bibitem[{Deng et~al.(2009)Deng, Dong, Socher, Li, Li, and Fei-Fei}]{deng2009imagenet}
Deng, J.; Dong, W.; Socher, R.; Li, L.-J.; Li, K.; and Fei-Fei, L. 2009.
\newblock ImageNet: A large-scale hierarchical image database.
\newblock In \emph{CVPR}, 248–255.

\bibitem[{Ding et~al.(2020)Ding, Fan, Xu, and Yang}]{ding2020adaptive}
Ding, Y.; Fan, H.; Xu, M.; and Yang, Y. 2020.
\newblock Adaptive Exploration for Unsupervised Person Re-identification.
\newblock \emph{TOMM}, 16(1): 1–19.

\bibitem[{Fu et~al.(2020)Fu, Cao, Long, and Wang}]{fu2020learning}
Fu, B.; Cao, Z.; Long, M.; and Wang, J. 2020.
\newblock Learning to Detect Open Classes for Universal Domain Adaptation.
\newblock In \emph{ECCV}, 567–583.

\bibitem[{Grandvalet and Bengio(2004)}]{grandvalet2004semi}
Grandvalet, Y.; and Bengio, Y. 2004.
\newblock Semi-supervised Learning by Entropy Minimization.
\newblock In \emph{NeurIPS}, volume~17.

\bibitem[{He et~al.(2016)He, Zhang, Ren, and Sun}]{he2016deep}
He, K.; Zhang, X.; Ren, S.; and Sun, J. 2016.
\newblock Deep Residual Learning for Image Recognition.
\newblock In \emph{CVPR}, 770–778.

\bibitem[{Hur et~al.(2023)Hur, Shin, Park, Woo, and Kweon}]{hur2023learning}
Hur, S.; Shin, I.; Park, K.; Woo, S.; and Kweon, I.~S. 2023.
\newblock Learning Classifiers of Prototypes and Reciprocal Points for Universal Domain Adaptation.
\newblock In \emph{WACV}, 531–540.

\bibitem[{Jing, Liu, and Ding(2021)}]{jing2021towards}
Jing, T.; Liu, H.; and Ding, Z. 2021.
\newblock Towards Novel Target Discovery Through Open-Set Domain Adaptation.
\newblock In \emph{ICCV}, 9322–9331.

\bibitem[{Kouw and Loog(2021)}]{kouw2021review}
Kouw, W.~M.; and Loog, M. 2021.
\newblock A Review of Domain Adaptation without Target Labels.
\newblock \emph{TPAMI}, 43(3): 766–785.

\bibitem[{Kundu et~al.(2020)Kundu, Venkat, V., and Babu}]{kundu2020universal}
Kundu, J.~N.; Venkat, N.; V., R.~M.; and Babu, R.~V. 2020.
\newblock Universal Source-Free Domain Adaptation.
\newblock In \emph{CVPR}, 4544–4553.

\bibitem[{LeCun, Bengio, and Hinton(2015)}]{lecun2015deep}
LeCun, Y.; Bengio, Y.; and Hinton, G. 2015.
\newblock Deep learning.
\newblock \emph{Nature}, 521(7553): 436–444.

\bibitem[{Li et~al.(2021)Li, Kang, Zhu, Wei, and Yang}]{li2021domain}
Li, G.; Kang, G.; Zhu, Y.; Wei, Y.; and Yang, Y. 2021.
\newblock Domain Consensus Clustering for Universal Domain Adaptation.
\newblock In \emph{CVPR}, 9757–9766.

\bibitem[{Liang et~al.(2021)Liang, Hu, Feng, and He}]{liang2021umad}
Liang, J.; Hu, D.; Feng, J.; and He, R. 2021.
\newblock UMAD: Universal Model Adaptation under Domain and Category Shift.
\newblock \emph{arXiv:2112.08553}.

\bibitem[{Lin et~al.(2022)Lin, Zhou, Qiu, and Zheng}]{lin2022adversarial}
Lin, K.-Y.; Zhou, J.; Qiu, Y.; and Zheng, W.-S. 2022.
\newblock Adversarial Partial Domain Adaptation by Cycle Inconsistency.
\newblock In \emph{ECCV}, 530–548.

\bibitem[{Luo, Song, and Zhang(2022)}]{luo2022learning}
Luo, C.; Song, C.; and Zhang, Z. 2022.
\newblock Learning to Adapt Across Dual Discrepancy for Cross-Domain Person Re-Identification.
\newblock \emph{TPAMI}, 45(2): 1963–1980.

\bibitem[{Mao et~al.(2019)Mao, Ma, Yang, Chen, and Li}]{mao2019virtual}
Mao, X.; Ma, Y.; Yang, Z.; Chen, Y.; and Li, Q. 2019.
\newblock Virtual Mixup Training for Unsupervised Domain Adaptation.
\newblock \emph{arXiv:1905.04215}.

\bibitem[{Olsson et~al.(2021)Olsson, Tranheden, Pinto, and Svensson}]{olsson2021classmix}
Olsson, V.; Tranheden, W.; Pinto, J.; and Svensson, L. 2021.
\newblock Classmix: Segmentation-based data augmentation for semi-supervised learning.
\newblock In \emph{WACV}, 1369--1378.

\bibitem[{Peng et~al.(2017)Peng, Usman, Kaushik, Hoffman, Wang, and Saenko}]{peng2017visda}
Peng, X.; Usman, B.; Kaushik, N.; Hoffman, J.; Wang, D.; and Saenko, K. 2017.
\newblock VisDA: The Visual Domain Adaptation Challenge.
\newblock \emph{arXiv:1710.06924}.

\bibitem[{Qu et~al.(2023)Qu, Zou, R{\"o}hrbein, Lu, Chen, Tao, and Jiang}]{qu2023upcycling}
Qu, S.; Zou, T.; R{\"o}hrbein, F.; Lu, C.; Chen, G.; Tao, D.; and Jiang, C. 2023.
\newblock Upcycling models under domain and category shift.
\newblock In \emph{CVPR}, 20019--20028.

\bibitem[{Rangwani et~al.(2022)Rangwani, Aithal, Mishra, Jain, and Babu}]{rangwani2022closer}
Rangwani, H.; Aithal, S.~K.; Mishra, M.; Jain, A.; and Babu, R.~V. 2022.
\newblock A Closer Look at Smoothness in Domain Adversarial Training.
\newblock In \emph{ICML}, 18378–18399.

\bibitem[{Saenko et~al.(2010)Saenko, Kulis, Fritz, and Darrell}]{saenko2010adapting}
Saenko, K.; Kulis, B.; Fritz, M.; and Darrell, T. 2010.
\newblock Adapting Visual Category Models to New Domains.
\newblock In \emph{ECCV}, 213–226.

\bibitem[{Saito et~al.(2020)Saito, Kim, Sclaroff, and Saenko}]{saito2020universal}
Saito, K.; Kim, D.; Sclaroff, S.; and Saenko, K. 2020.
\newblock Universal Domain Adaptation through Self Supervision.
\newblock In \emph{NeurIPS}, volume~33, 16282–16292.

\bibitem[{Saito and Saenko(2021)}]{saito2021ovanet}
Saito, K.; and Saenko, K. 2021.
\newblock OVANet: One-vs-All Network for Universal Domain Adaptation.
\newblock In \emph{ICCV}, 9000–9009.

\bibitem[{Saito et~al.(2018)Saito, Yamamoto, Ushiku, and Harada}]{saito2018open}
Saito, K.; Yamamoto, S.; Ushiku, Y.; and Harada, T. 2018.
\newblock Open Set Domain Adaptation by Backpropagation.
\newblock In \emph{ECCV}, 153–168.

\bibitem[{Selvaraju et~al.(2017)Selvaraju, Cogswell, Das, Vedantam, Parikh, and Batra}]{selvaraju2017grad}
Selvaraju, R.~R.; Cogswell, M.; Das, A.; Vedantam, R.; Parikh, D.; and Batra, D. 2017.
\newblock Grad-cam: Visual explanations from deep networks via gradient-based localization.
\newblock In \emph{ICCV}, 618--626.

\bibitem[{Shen et~al.(2023)Shen, Lu, Hu, and Ma}]{shen2023collaborative}
Shen, M.; Lu, Y.; Hu, Y.; and Ma, A.~J. 2023.
\newblock Collaborative Learning of Diverse Experts for Source-free Universal Domain Adaptation.
\newblock In \emph{ACM MM}, 2054--2065.

\bibitem[{Su et~al.(2023)Su, Han, He, Wei, He, and Yin}]{su2023neighborhood}
Su, W.; Han, Z.; He, R.; Wei, B.; He, X.; and Yin, Y. 2023.
\newblock Neighborhood-based credibility anchor learning for universal domain adaptation.
\newblock \emph{Pattern Recognition}, 142: 109686.

\bibitem[{Tranheden et~al.(2021)Tranheden, Olsson, Pinto, and Svensson}]{tranheden2021dacs}
Tranheden, W.; Olsson, V.; Pinto, J.; and Svensson, L. 2021.
\newblock DACS: Domain Adaptation via Cross-domain Mixed Sampling.
\newblock In \emph{WACV}, 1379–1389.

\bibitem[{Van~der Maaten and Hinton(2008)}]{van2008visualizing}
Van~der Maaten, L.; and Hinton, G. 2008.
\newblock Visualizing data using t-SNE.
\newblock \emph{Journal of Machine Learning Research}, 9(11).

\bibitem[{Venkateswara et~al.(2017)Venkateswara, Eusebio, Chakraborty, and Panchanathan}]{venkateswara2017deep}
Venkateswara, H.; Eusebio, J.; Chakraborty, S.; and Panchanathan, S. 2017.
\newblock Deep Hashing Network for Unsupervised Domain Adaptation.
\newblock In \emph{CVPR}, 5018–5027.

\bibitem[{Verma et~al.(2019)Verma, Lamb, Beckham, Najafi, Mitliagkas, Lopez-Paz, and Bengio}]{verma2019manifold}
Verma, V.; Lamb, A.; Beckham, C.; Najafi, A.; Mitliagkas, I.; Lopez-Paz, D.; and Bengio, Y. 2019.
\newblock Manifold Mixup: Better Representations by Interpolating Hidden States.
\newblock In \emph{ICML}, 6438–6447.

\bibitem[{Wu, Inkpen, and El-Roby(2020)}]{wu2020dual}
Wu, Y.; Inkpen, D.; and El-Roby, A. 2020.
\newblock Dual Mixup Regularized Learning for Adversarial Domain Adaptation.
\newblock In \emph{ECCV}, 540–555.

\bibitem[{Wu et~al.(2018)Wu, Xiong, Yu, and Lin}]{wu2018unsupervised}
Wu, Z.; Xiong, Y.; Yu, S.~X.; and Lin, D. 2018.
\newblock Unsupervised Feature Learning via Non-parametric Instance Discrimination.
\newblock In \emph{CVPR}, 3733–3742.

\bibitem[{Yan et~al.(2020)Yan, Song, Li, Zou, and Ren}]{yan2020improve}
Yan, S.; Song, H.; Li, N.; Zou, L.; and Ren, L. 2020.
\newblock Improve Unsupervised Domain Adaptation with Mixup Training.
\newblock \emph{arXiv:2001.00677}.

\bibitem[{Yang et~al.(2021)Yang, Wang, Gao, Shrivastava, Weinberger, Chao, and Lim}]{yang2021deep}
Yang, L.; Wang, Y.; Gao, M.; Shrivastava, A.; Weinberger, K.~Q.; Chao, W.-L.; and Lim, S.-N. 2021.
\newblock Deep Co-Training with Task Decomposition for Semi-Supervised Domain Adaptation.
\newblock In \emph{ICCV}, 8906–8916.

\bibitem[{Yang et~al.(2022)Yang, Wang, Wang, Jui, and van~de Weijer}]{yang2022one}
Yang, S.; Wang, Y.; Wang, K.; Jui, S.; and van~de Weijer, J. 2022.
\newblock One Ring to Bring Them All: Towards Open-Set Recognition under Domain Shift.
\newblock \emph{arXiv:2206.03600}.

\bibitem[{You et~al.(2019)You, Long, Cao, Wang, and Jordan}]{you2019universal}
You, K.; Long, M.; Cao, Z.; Wang, J.; and Jordan, M.~I. 2019.
\newblock Universal Domain Adaptation.
\newblock In \emph{CVPR}, 2720–2729.

\bibitem[{Zhang et~al.(2018)Zhang, Cisse, Dauphin, and Lopez-Paz}]{zhang2018mixup}
Zhang, H.; Cisse, M.; Dauphin, Y.~N.; and Lopez-Paz, D. 2018.
\newblock mixup: Beyond Empirical Risk Minimization.
\newblock In \emph{ICLR}.

\bibitem[{Zhong et~al.(2019)Zhong, Zheng, Luo, Li, and Yang}]{zhong2019invariance}
Zhong, Z.; Zheng, L.; Luo, Z.; Li, S.; and Yang, Y. 2019.
\newblock Invariance Matters: Exemplar Memory for Domain Adaptive Person Re-Identification.
\newblock In \emph{CVPR}, 598–607.

\end{thebibliography}

\clearpage

\section{Theoretical Analysis}
\begin{figure*}[t]
    \centering
    \subfloat[$\beta_1=0.1$]{\includegraphics[height=3.5cm]{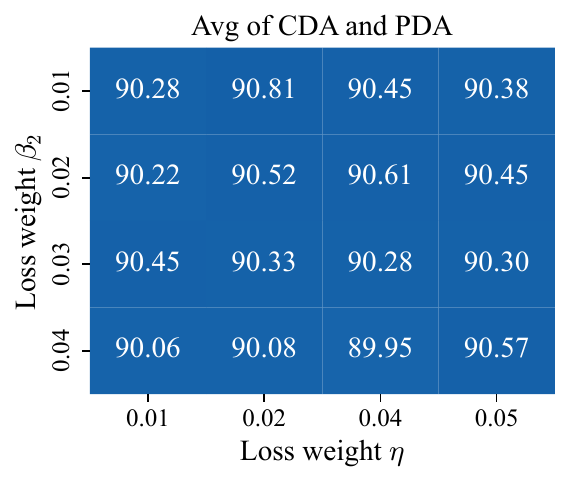}
    \includegraphics[height=3.5cm]{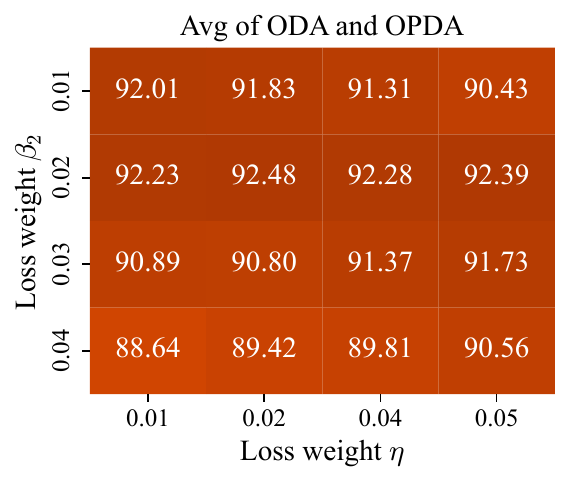}}
    \hspace{6pt}
    \subfloat[$\beta_1=1.0$]{\includegraphics[height=3.5cm]{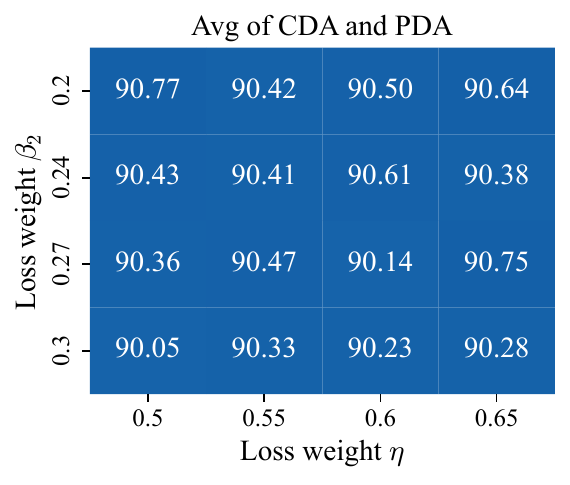}
    \includegraphics[height=3.5cm]{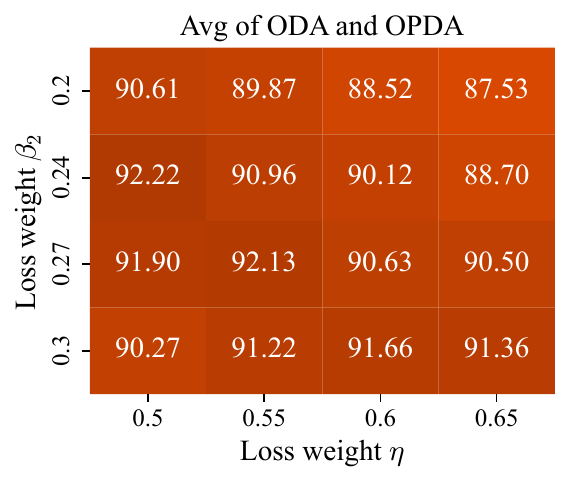}}
    \caption{
        Sensitivity analysis when varying the loss weight $\beta_1$ for neighborhood invariance learning on the Office dataset.
    }
    \label{fig:beta1}
\end{figure*}

\subsection{Self-adaptive Neighbor Search}
To thoroughly analyze the effectiveness of our proposed self-adaptive neighbor search, we provide a proof for the claim that, \emph{compared with the widely used $\mathcal{K}$-nearest neighbors, the imbalance of sample sizes within a dataset is mitigated by our proposed self-adaptive neighbor search}.

\begin{proof}
Denote the ratio of neighbors to the sample size of the $i$-th class as,
\begin{equation}
\label{eq:ratio}
    \beta_i = \frac{k_i}{n_i}, 
\end{equation}
where $k_i$ and $n_i$ refer to the number of the neighbors and all the $i$-th class samples, respectively.
Without loss of generality, assume that the data is imbalanced between the $i$-th and $j$-th classes with $n_i > n_j$.
The relative neighbor ratio between the $i$-th and the $j$-th classes is denoted as,
\begin{equation}
\label{eq:relative}
    \gamma_{ij} = \frac{\beta_i}{\beta_j} = \frac{n_j}{n_i} \cdot \frac{k_i}{k_j}.
\end{equation}
Then, $\gamma_{ij} = 1$ holds if and only if the search of neighbors for different classes (i.e., $i$ and $j$) is balanced.

Considering the widely used $\mathcal{K}$-nearest neighbors, $k_i$ and $k_j$ both equal to $\mathcal{K}$ in Equation~\eqref{eq:relative}, so the relative neighbor ratio is degenerated as,
\begin{equation}
\label{eq:knn}
    \hat{\gamma}_{ij} =  \frac{n_j}{n_i}, 
\end{equation}
However, given that $n_i > n_j$, we have $\hat{\gamma}_{ij} < 1$, which means that the $\mathcal{K}$-nearest neighbors tends to sample the $j$-th class data more frequently, resulting in the model biased during training.
In our proposed self-adaptive neighbor search, the amount of neighbor samples selected in the $i$-th class is positively proportional to the sample size of the $i$-th class under the assumption of neighborhood invariance that data points within a local neighborhood are more likely to share the same class label, i.e., 
\begin{equation}
\label{eq:propto}
    k_i \propto  n_i.
\end{equation}
Therefore, we can infer that $k_i > k_j$ with $n_i > n_j$.
Then, we have the following inequality,
\begin{equation}
\label{eq:inequal}
    \frac{n_j}{n_i} \cdot \frac{k_i}{k_j} > \frac{n_j}{n_i} \cdot \frac{k}{k} > 0.
\end{equation}
From Equation~\eqref{eq:inequal}, the amount of neighbors selected for the $i$-th class is enlarged by a coefficient $\frac{k_i}{k_j} > 1$.
So the relative neighbor ratio $\gamma_{ij}$ is closer to 1 than the corresponding $\hat{\gamma}_{ij}$, for more balanced neighbor search. 
Thus, the imbalance problem of neighbor search between the $i$-th and $j$-th class is mitigated.
Since the $i$-th class has more samples than the $j$-th class (i.e., $n_i > n_j$), sampling more neighbors for the $i$-th class also helps to ensure that the confidence level of neighbors is essentially the same across classes.
\end{proof}

\subsection{Cross-domain Manifold Mixup}
Without label information in the target domain, the label of the mixup feature vector cannot be determined as in the vanilla mixup scheme~\cite{zhang2018mixup}.
Nevertheless, we provide a proof for the fact that \emph{the mixup feature vector belongs to a certain known class with low probability, and thus can help to improve unknown-class identification}.

\begin{proof}
Denote $y_i^s$ as the label of the $i$-th sample $x_i^s$ in the source domain and $y_j^t$ as the label of the $j$-th sample $x_j^t$ in the target domain.
Let the total number of categories in the source and target domains be  $K=|\mathcal{C}_s|$ and $K' = |\mathcal{C}_t|$, respectively, while the number of shared categories is defined as $K_s = |\mathcal{C}_s \cap \mathcal{C}_t|$.
The mixup feature vector is computed by $z_{i,j,\lambda}^m = \lambda\mathcal{F}(x_i^s)  + (1-\lambda) \mathcal{F}(x_j^t)$, where $\mathcal{F}$ is the feature extractor and $\lambda$ is randomly sampled from a Beta distribution.
Then, $z_{i,j,\lambda}^m$ is from one of the known classes if and only if $x_i^s$ and $x_j^t$ are from the same known class, i.e., $y_i^s = y_j^t$ and $y_i^s, y_j^t \in \mathcal{C}_s \cap \mathcal{C}_t$.
The probability of such event is given by,
\begin{equation}
\begin{split}
    \label{eq:event}
    p(y_i^s = y_j^t \in \mathcal{C}_s \cap \mathcal{C}_t) & = \frac{\#\{y_i^s = y_j^t \in \mathcal{C}_s \cap \mathcal{C}_t\}}{\#\{y_i^s \in \mathcal{C}_s, y_j^t \in \mathcal{C}_t\}} \\
    & = \frac{C_{K_s}^1}{C_{K}^1 \cdot C_{K'}^1} = \frac{K_s}{K K'},
\end{split}
\end{equation}
where $\#$ denotes the number of elementary events and $C$ is the combination number.
Since $K_s \leq \min\{K, K'\}$, $p(y_i^s = y_j^t \in \mathcal{C}_s \cap \mathcal{C}_t)$ is a small number.
For example, if $K = 20, K' = 21, K_s = 10$ in the OPDA setting on the Office dataset, then $p(y_i^s = y_j^t \in \mathcal{C}_s \cap \mathcal{C}_t) \approx 0.024$.
Thus, the mixup feature vector $z_{i,j,\lambda}^m$ belongs to a certain known class with low probability.
In contrast, the probability that $z_{i,j,\lambda}^m$ is NOT from one of the known classes is a large number given by $1 - p(y_i^s = y_j^t \in \mathcal{C}_s \cap \mathcal{C}_t)$.
This means the mixup feature vector can be considered as an unknown-class sample with high probability for learning the open-set classifiers.
\end{proof}

\section{Supplementary Experiments}
\begin{figure*}[t]
    \centering
    \subfloat[Known-class]{\includegraphics[height=3cm]{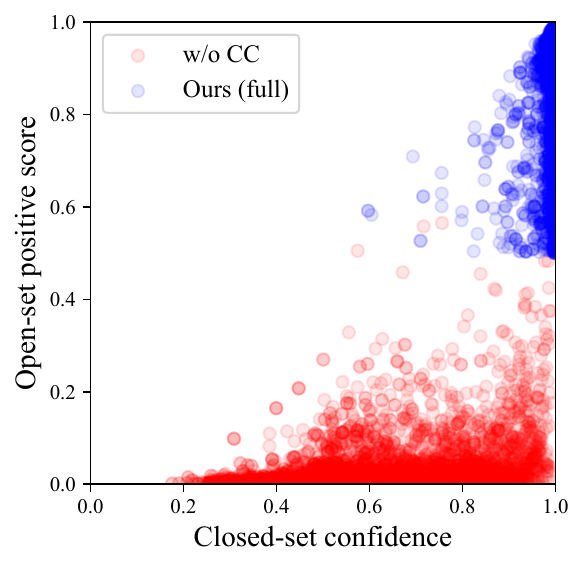}}
    \subfloat[Unknown-class]{\includegraphics[height=3cm]{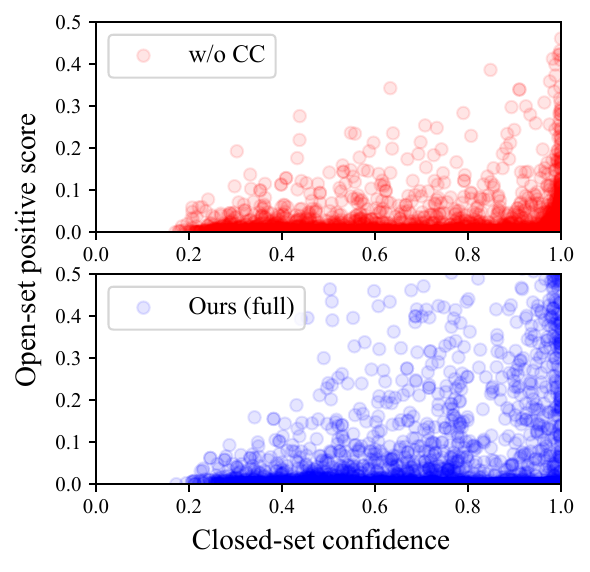}}
    \subfloat[Ablation with UCR]{\includegraphics[height=2.95cm]{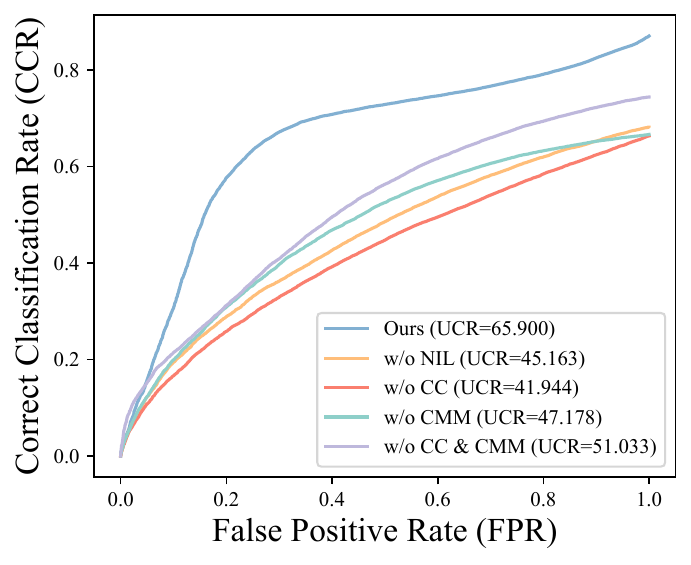}}
    \subfloat[w/ KNN]{\includegraphics[height=2.95cm]{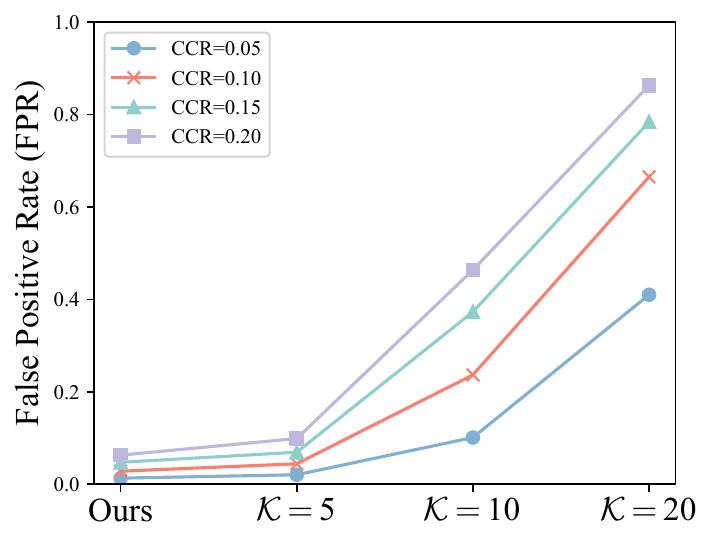}}
    \subfloat[Extreme $\epsilon$]{\includegraphics[height=2.95cm]{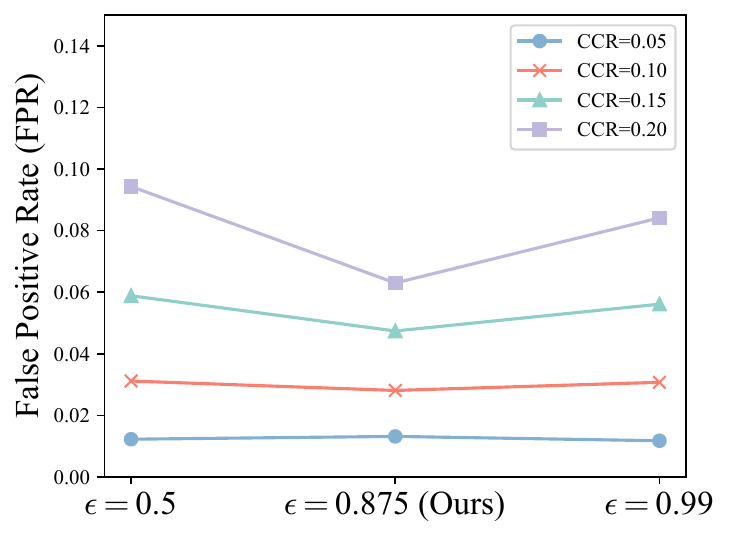}}
    \caption{
    (a)(b) Visualizations of known- and unknown-class distributions w/o and w/ CC. 
    (c)(d)(e) More ablation results.
    CCR and FPR are short for Correct Classification Rate (higher is better) and False Positive Rate (lower is better), respectively.
    UCR score refers to the area under the (CCR vs FPR) curve in open-set settings~\cite{deng2023universal}.
    }
    \label{fig:supplementary}
\end{figure*}

\subsection{Datasets}
Our experiments are conducted on three publicly available UniDA benchmarks, including Office~\cite{saenko2010adapting}, OfficeHome~\cite{venkateswara2017deep} and VisDA~\cite{peng2017visda}.
In Tables~\ref{tab:office}-\ref{tab:visda}, we present the split details across all settings in UniDA for each of these three dataset.
We use $(|\mathcal{C}_s\cap\mathcal{C}_t|/|\mathcal{C}_s-\mathcal{C}_t|/|\mathcal{C}_t-\mathcal{C}_s|)$ to represent the split, which are the numbers of shared, source private and target private categories, respectively.
\textbf{Office} consists of 31 categories and contains around 4,700 images across three domains, namely Amazon (A), DSLR (D) and Webcam (W). 
\textbf{OfficeHome} consists of 65 categories and contains around 15,500 images across four domains, namely Artistic (A), Clip-art (C), Product (P) and Real-world (R).
\textbf{VisDA} is a large-scale dataset consisting of 12 categories and containing around 150,000 synthetic and 50,000 real-world images in the source and target domains, respectively.

\subsection{Comparison with State-of-the-arts}
Due to page limit, we place the comparisons with state-of-the-arts in the ODA, PDA and CDA settings here, including UAN~\cite{you2019universal}, CMU~\cite{fu2020learning}, DANCE~\cite{saito2020universal}, DCC~\cite{li2021domain}, OVANet~\cite{saito2021ovanet}, GATE~\cite{chen2022geometric}, MATHS~\cite{chen2022mutual}, TNT~\cite{chen2022evidential}, UniOT~\cite{chang2022unified}, CPR~\cite{hur2023learning} and NCAL~\cite{su2023neighborhood}.

\noindent\textbf{ODA Setting.}
Table~\ref{tab:oda} present overall comparison with state-of-the-arts in the ODA setting.
Our method consistently yields the best H-score on the Office dataset and outperforms previous works except for the baseline by at least 3\%.
While on the OfficeHome and VisDA datasets, we would like to highlight that competitive results are obtained when compared to state-of-the-arts and the improvements over the baseline are significant 2-3\%.

\noindent\textbf{PDA Setting.}
In Table~\ref{tab:pda}, the results show that our method achieves a new state-of-the-art and beats all previous approaches in the PDA setting across three benchmarks.
Especially on the OfficeHome and VisDA datasets, our method outperforms previous studies by a large margin of about 5\%.

\noindent\textbf{CDA Setting.}
From the results in Table~\ref{tab:cda}, our method achieves significant improvements against the baseline, i.e., 4.2\%, 5.3\% and 14.5\% on the three datasets, respectively.
When compared to other state-of-the-arts, although slightly inferior to GATE, we beat almost everyone else across three benchmarks.
We would like to highlight that the assumption of CDA that category-shift does not exist and $\mathcal{C}_s=\mathcal{C}_t$ is not always practical in real-world scenarios, while our baseline pays more attention to the identification of unknown classes and trains one-vs-all classifiers for each known class.

\subsection{Hyperparameter Sensitivity}
In our experiments, we fix the loss weight $\beta_1=0.5$ for neighbor invariance learning and perform a grid search on the values of $\beta_2$ and $\eta$ for cross-domain manifold mixup and consistency constraint, respectively.
The hyperparameter sensitivity experiments in the manuscript has proven that our method is robust enough to the choices of loss weights.
For completeness, we additionally conduct two grid searches fixing $\beta_1=0.1$ and $\beta_1=1.0$ in Figure~\ref{fig:beta1}, where the average accuracy in CDA and PDA settings, and H-score in ODA and OPDA settings are shown.
The results indicate that our method still gives best performance when varying $\beta_1$ on the Office dataset.
Both accuracy and H-score are only slightly varied by around 1\% and 3\%, respectively.

\subsection{More Ablation Studies}

\noindent\textbf{Consistency Constraint (CC).}
In Figure~\ref{fig:supplementary}(a), we randomly sample 5,000 known-class images on the VisDA dataset under the OPDA setting to validate Figure~\ref{fig:schematic}(h) in the manuscript.
It shows that there are indeed a large number of data points (in red) with high closed-set confidence but low open-set positive score when the CC is absent.
Such inconsistency is reduced by the CC for more accurate known-class predictions (in blue).
Meanwhile, another 5,000 unknown-class images are randomly sampled and visualized in Figure~\ref{fig:supplementary}(b).
We can see that the overconfidence problem does exist, but $p_o(l|x_j^t)$ does not ``increase remarkably" for wrong predictions of the unknown class.
This is because the Cross-domain Manifold Mixup (CMM) gives a strong bootstrap by simulating unknown-class samples for a good trade-off of the open-set classifier.
Quantitatively, taking the known class as positive and the unknown class as negative, the recall and precision increase from 9.64\% to 62.36\%\textbf{(+52.72)} and 84.47\% to 86.20\%\textbf{(+1.73)}, respectively.
It demonstrates that our CC is able to substantially improve the classification of known classes without compromising the identification of unknown classes.

\noindent\textbf{(CCR vs FPR) curve.}
In Figure~\ref{fig:supplementary}(c), we use (CCR vs FPR) curve~\cite{deng2023universal} as an evaluation metric for a more persuasive ablation study.
For example, when the CC is enabled, the CCR improves significantly by correctly predicting much more known-class samples w.r.t. the same FPR, which is in line with the illustration in Figure~\ref{fig:supplementary}(a).

\noindent\textbf{Neighborhood Invariance Learning (NIL).}
Our proposed NIL differs from existing neighbor-based methods like DANCE~\cite{saito2020universal} and GLC~\cite{qu2023upcycling} by online clustering (instead of offline) when reducing the intra-class distances in the target domain.
This improves training efficiency and avoids offline clustering such as K-Means for too long time when the data volume is large.
Compared to the GATE using KNN, we experiment with more $\mathcal{K}$-values in Figure~\ref{fig:supplementary}(d), which shows a high sensitivity to the neighborhood hyperparameter.
In contrast, even if we take extreme $\epsilon$, we can still maintain good robustness as shown in Figure~\ref{fig:supplementary}(e).
We believe this benefits from the self-adaptation during training.
The binary decision boundaries of each open-set classifier changes w.r.t. different classes for better identification of unknown classes, compared to the ($|\mathcal{C}_s|$+1)-dimension universal classifier used in GATE.
Another advantage of our method is the consistent hyperparameter $\epsilon$ across different datasets, whereas GATE sets a smaller $\mathcal{K}$-value for the Office dataset.

\subsection{More t-SNE Visualizations}
\begin{figure*}[t]
    \centering
    \subfloat[A2D]{\includegraphics[width=.32\linewidth]{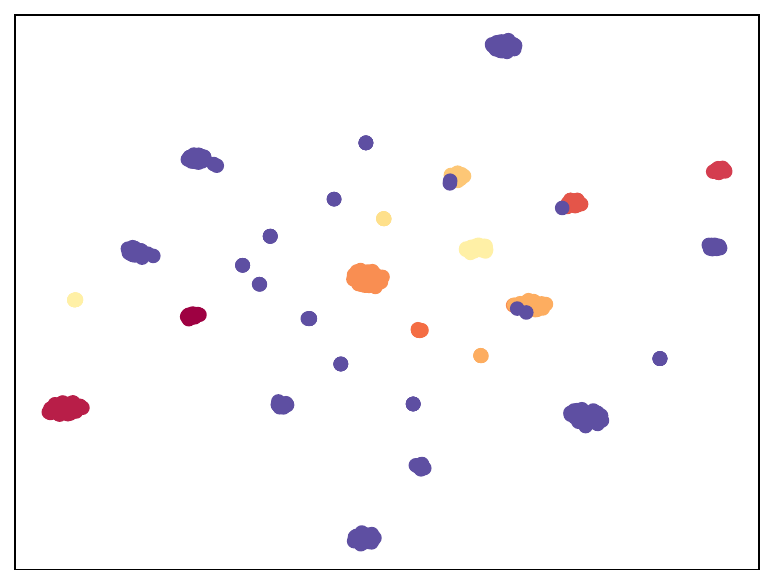}}
    \subfloat[A2W]{\includegraphics[width=.32\linewidth]{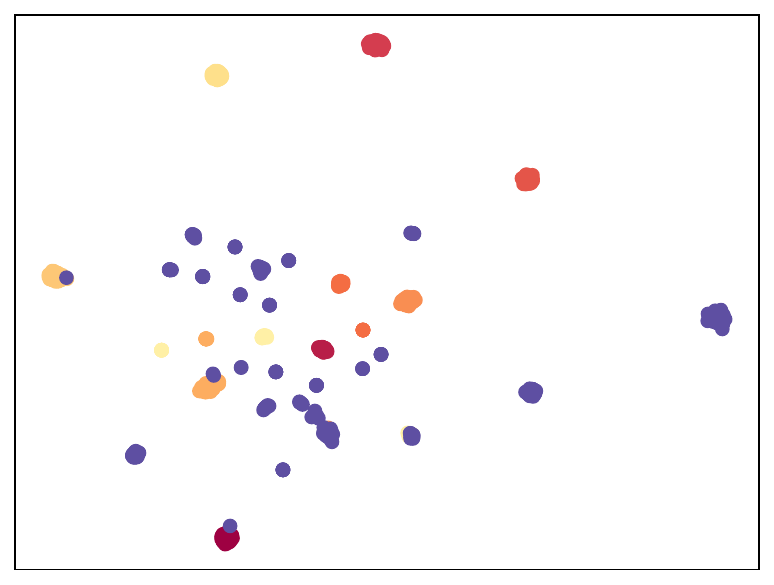}}
    \subfloat[D2A]{\includegraphics[width=.32\linewidth]{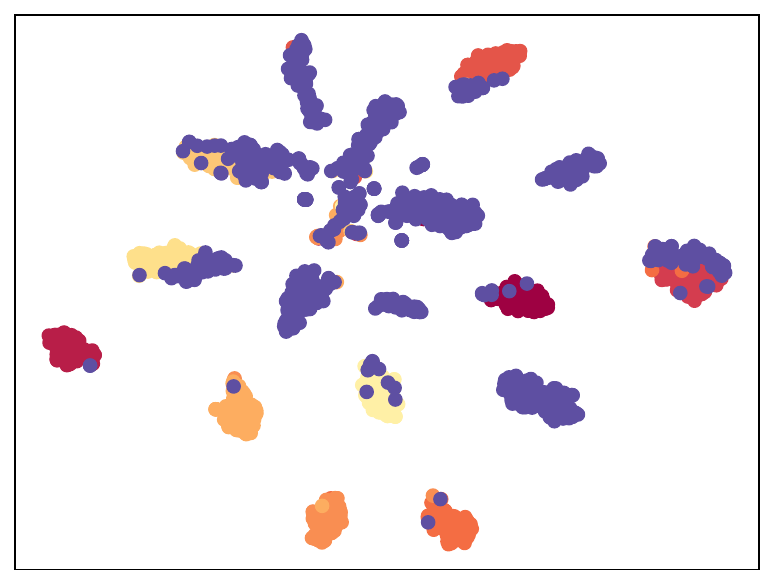}}\\
    \subfloat[D2W]{\includegraphics[width=.32\linewidth]{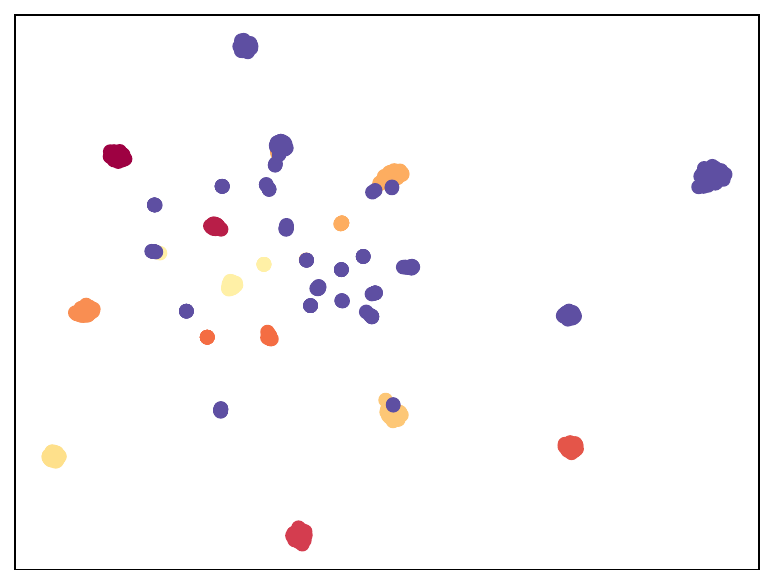}}
    \subfloat[W2A]{\includegraphics[width=.32\linewidth]{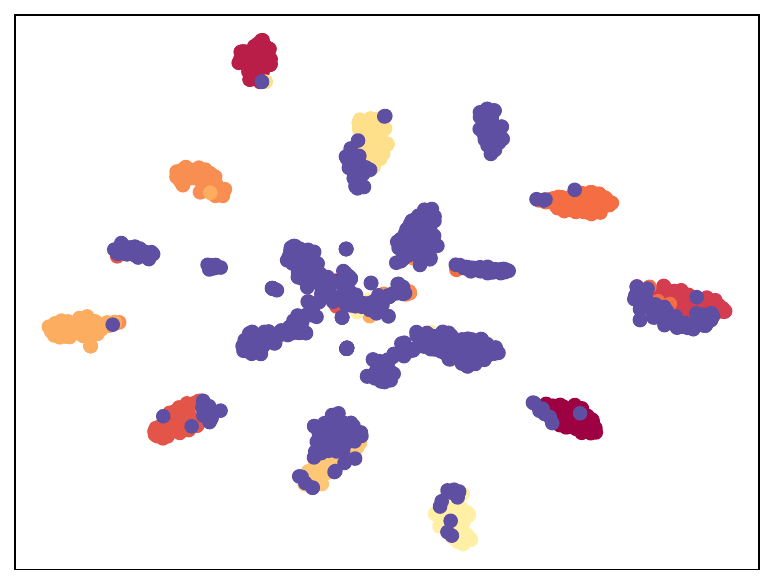}}
    \subfloat[W2D]{\includegraphics[width=.32\linewidth]{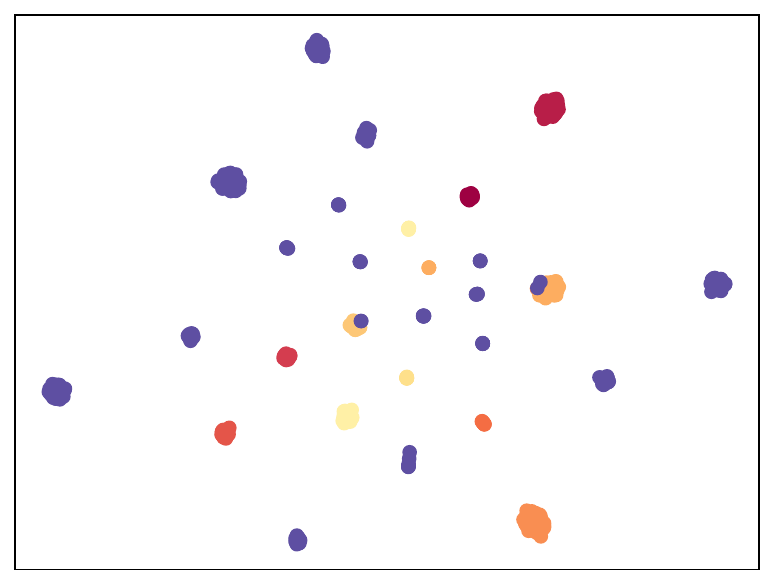}}
    \caption{t-SNE visualization by our method in the OPDA setting on the Office dataset.}
    \label{fig:opda1}
\end{figure*}

\begin{figure*}[t]
    \centering
    \subfloat[A2D]{\includegraphics[width=.32\linewidth]{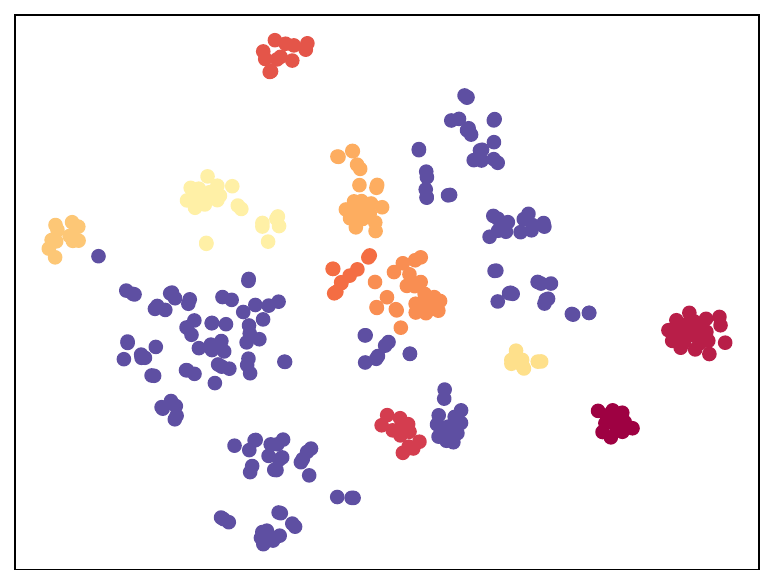}}
    \subfloat[A2W]{\includegraphics[width=.32\linewidth]{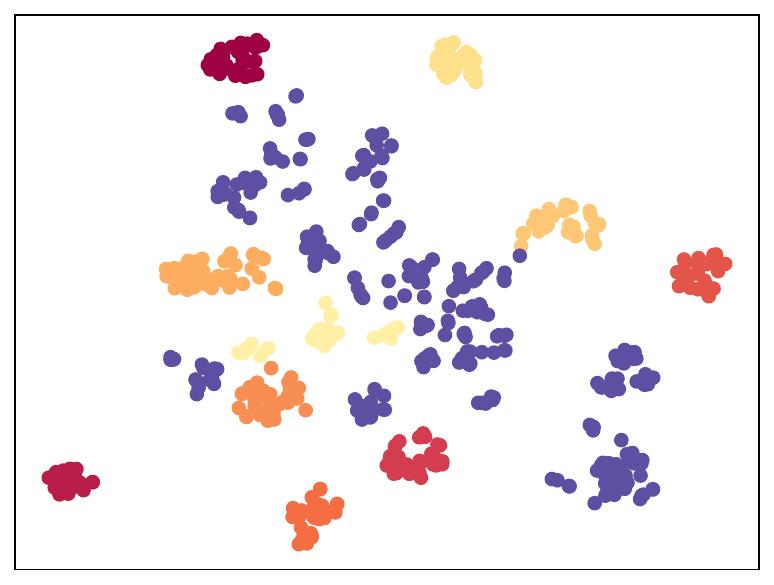}}
    \subfloat[D2A]{\includegraphics[width=.32\linewidth]{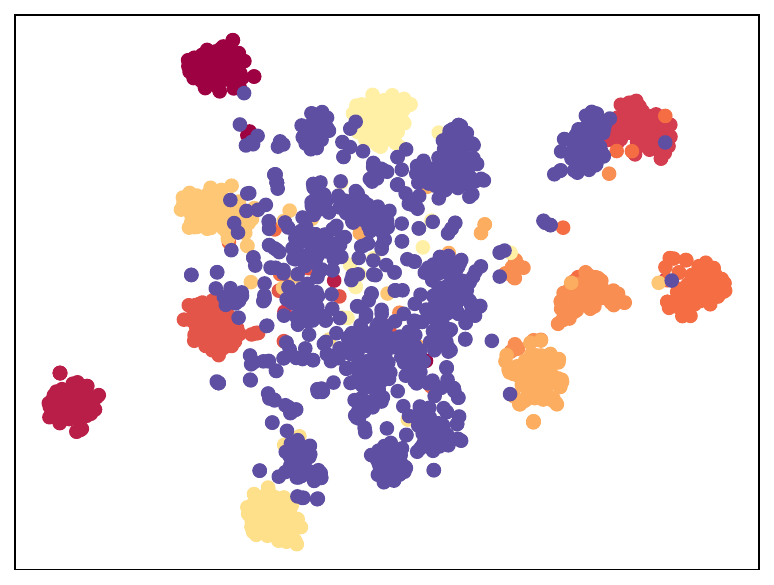}}\\
    \subfloat[D2W]{\includegraphics[width=.32\linewidth]{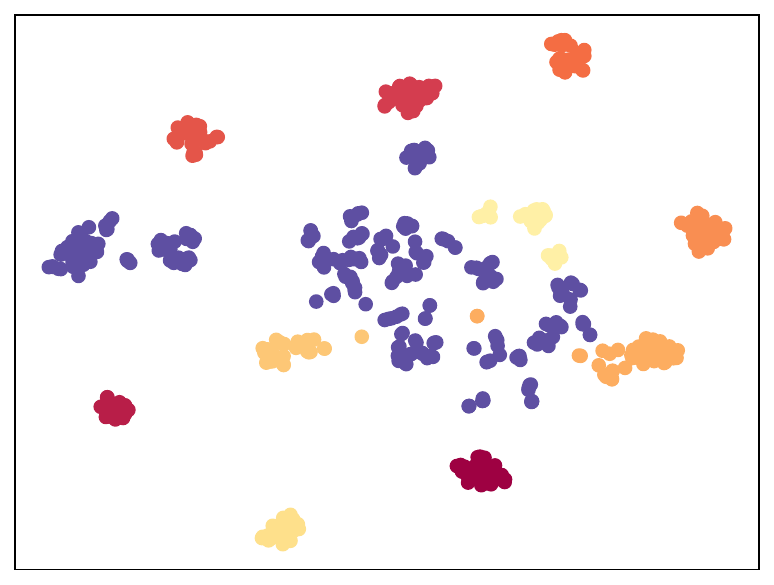}}
    \subfloat[W2A]{\includegraphics[width=.32\linewidth]{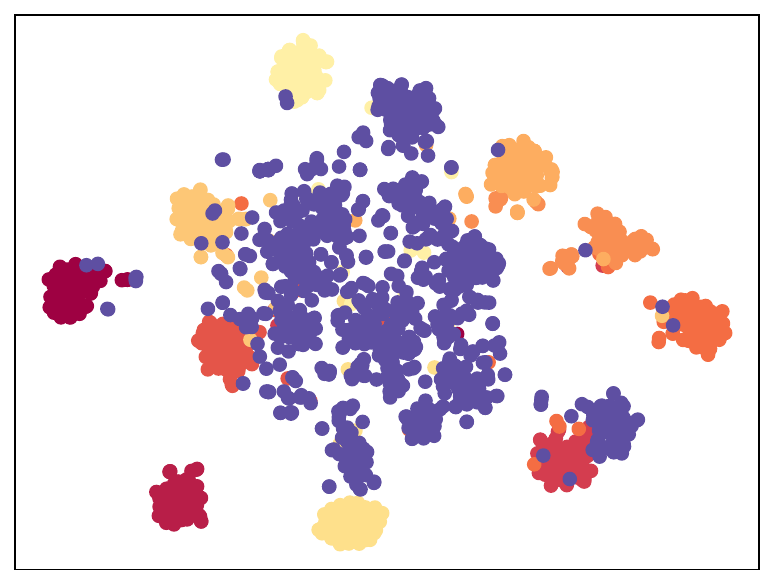}}
    \subfloat[W2D]{\includegraphics[width=.32\linewidth]{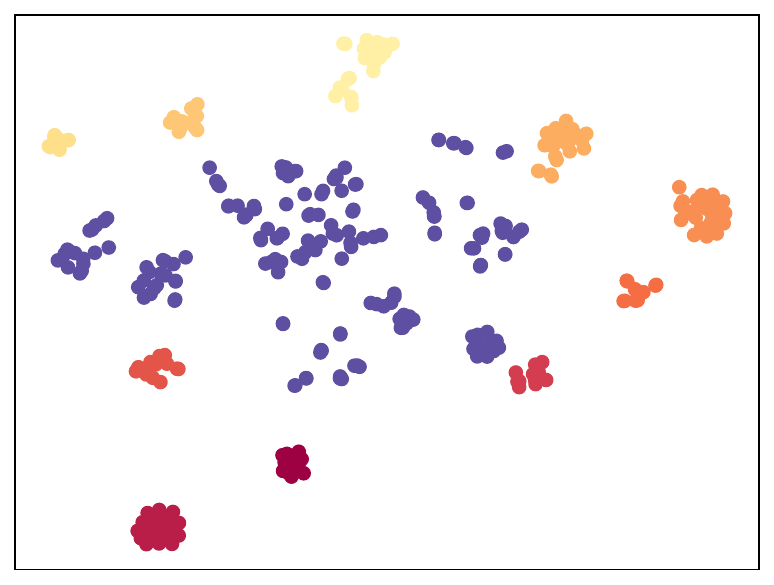}}
    \caption{t-SNE visualization by the baseline in the OPDA setting on the Office dataset.}
    \label{fig:opda2}
\end{figure*}

\begin{figure*}[t]
    \centering
    \subfloat[A2D]{\includegraphics[width=.32\linewidth]{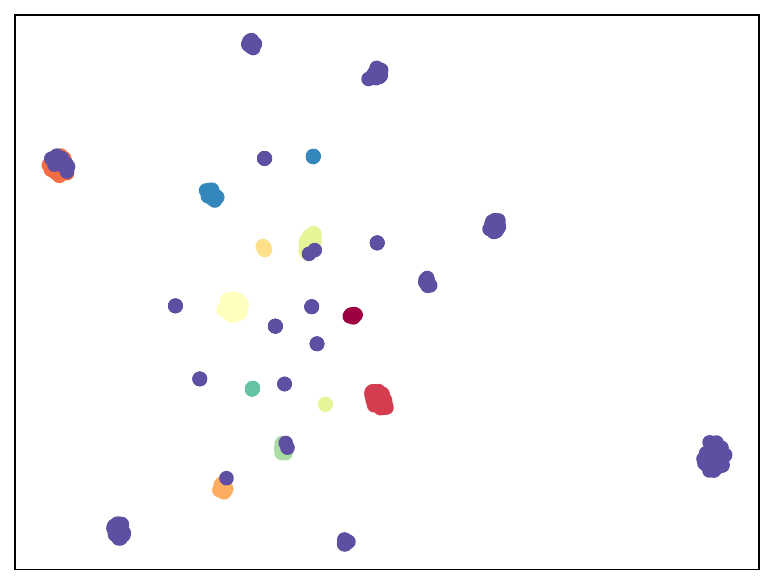}}
    \subfloat[A2W]{\includegraphics[width=.32\linewidth]{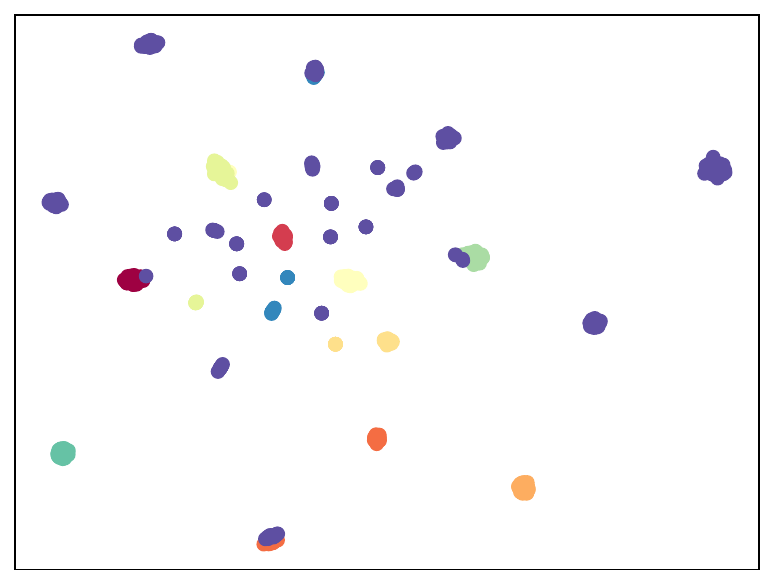}}
    \subfloat[D2A]{\includegraphics[width=.32\linewidth]{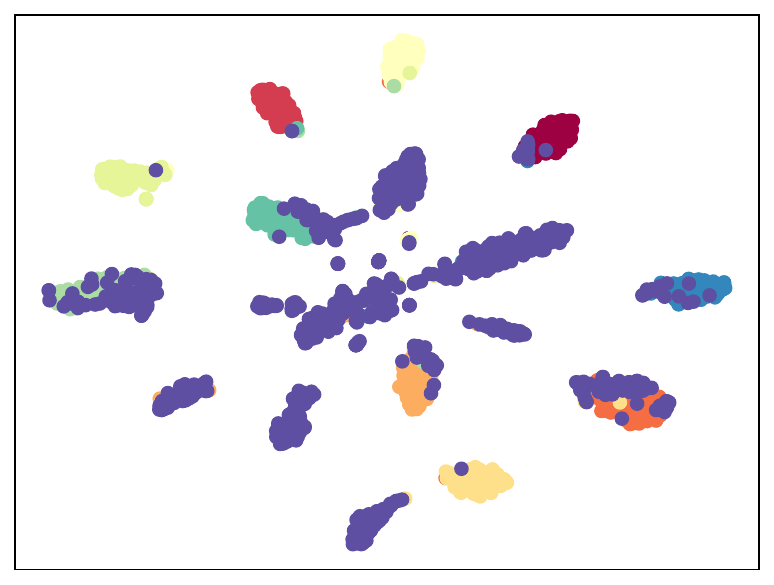}}\\
    \subfloat[D2W]{\includegraphics[width=.32\linewidth]{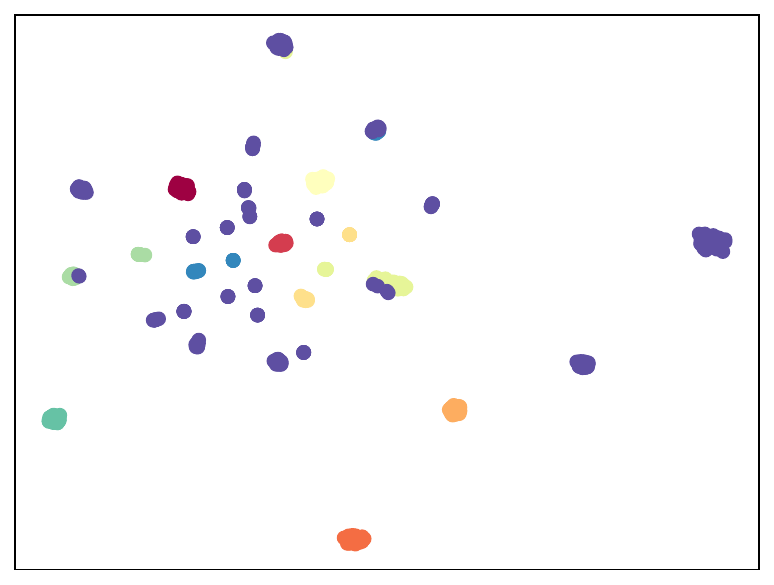}}
    \subfloat[W2A]{\includegraphics[width=.32\linewidth]{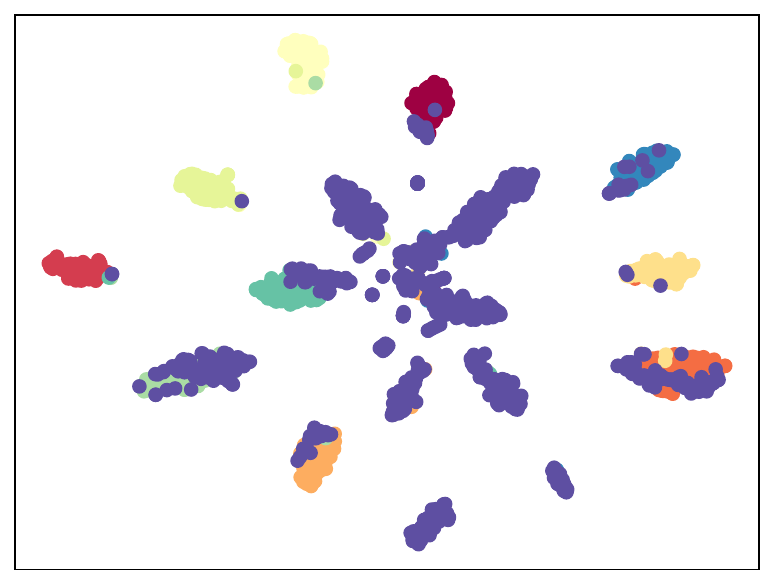}}
    \subfloat[W2D]{\includegraphics[width=.32\linewidth]{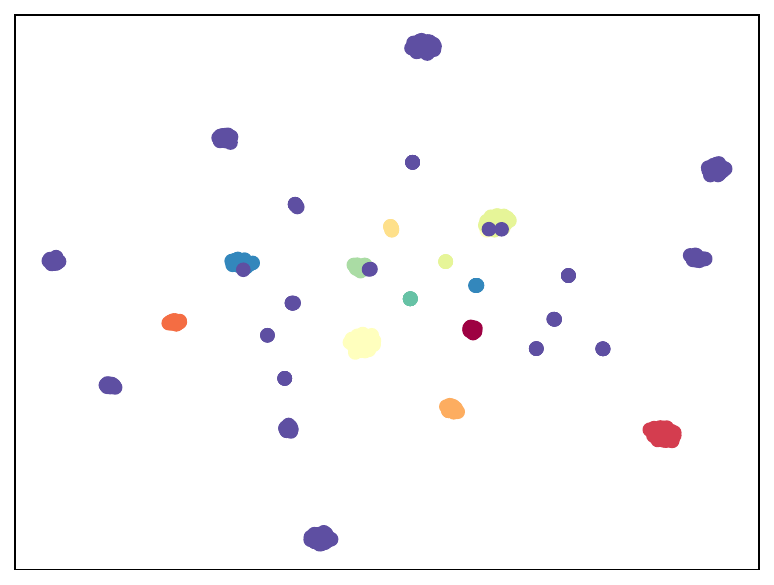}}
    \caption{t-SNE visualization by our method in the ODA setting on the Office dataset.}
    \label{fig:oda1}
\end{figure*}

\begin{figure*}[t]
    \centering
    \subfloat[A2D]{\includegraphics[width=.32\linewidth]{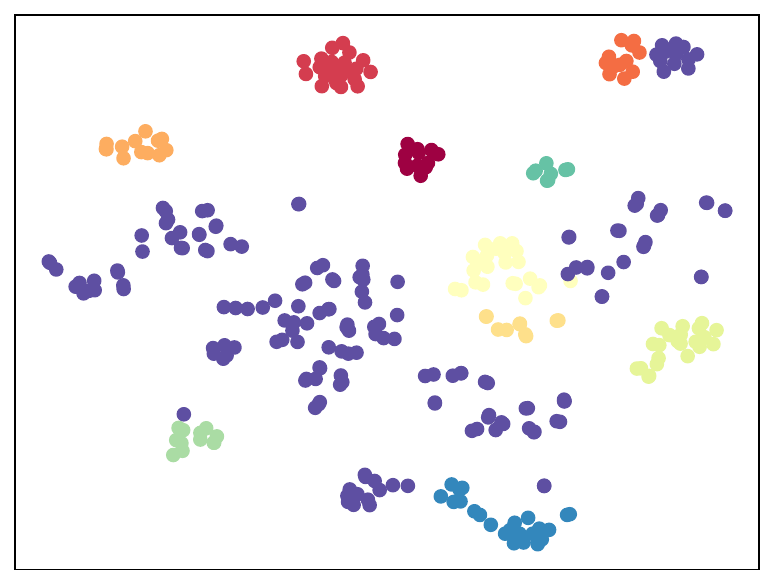}}
    \subfloat[A2W]{\includegraphics[width=.32\linewidth]{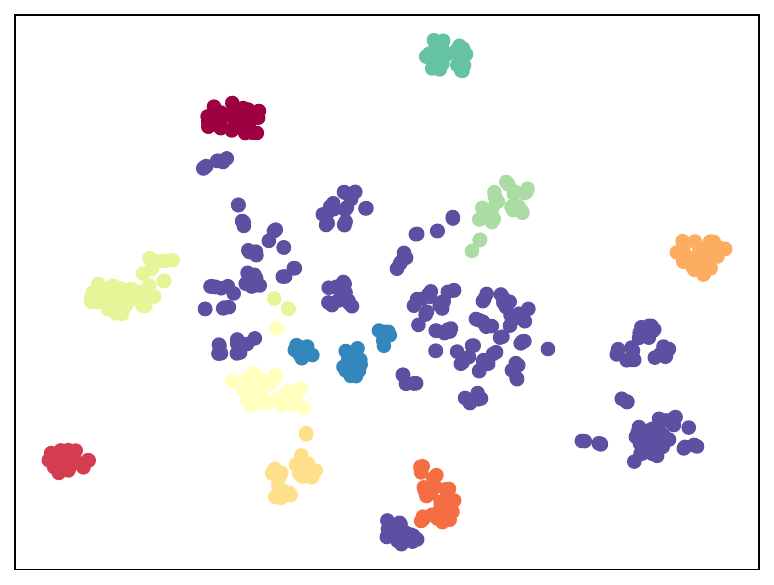}}
    \subfloat[D2A]{\includegraphics[width=.32\linewidth]{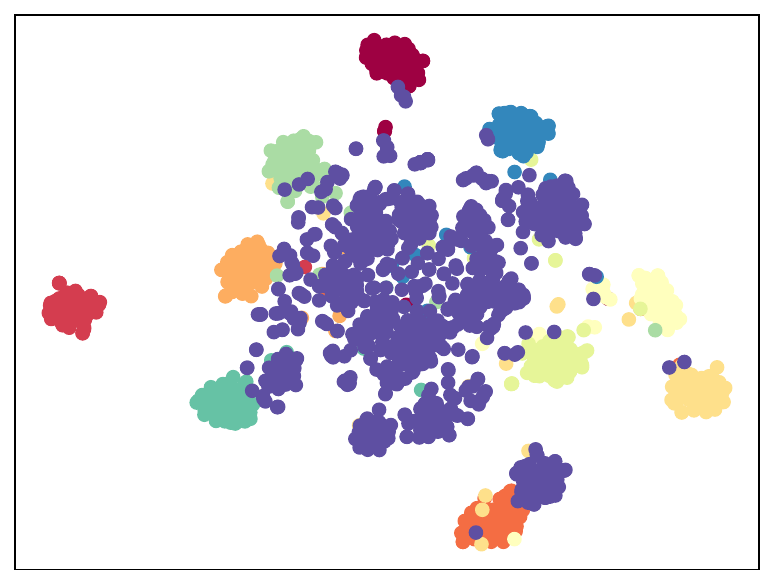}}\\
    \subfloat[D2W]{\includegraphics[width=.32\linewidth]{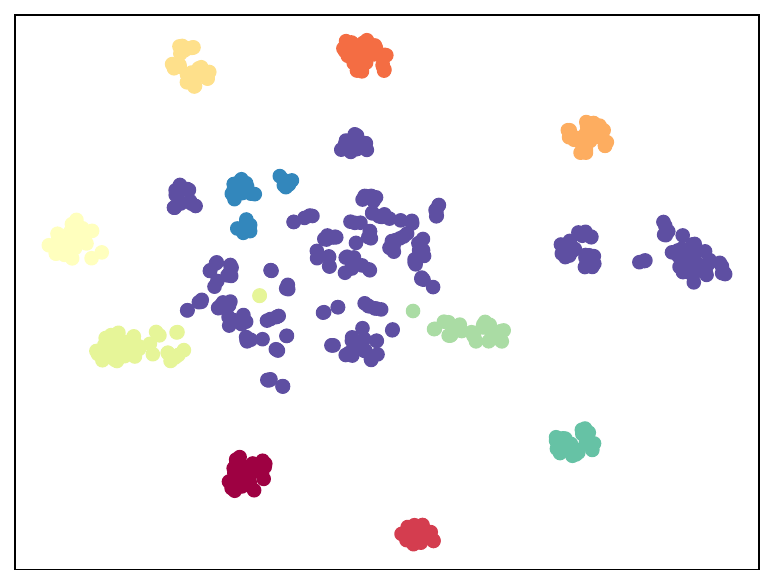}}
    \subfloat[W2A]{\includegraphics[width=.32\linewidth]{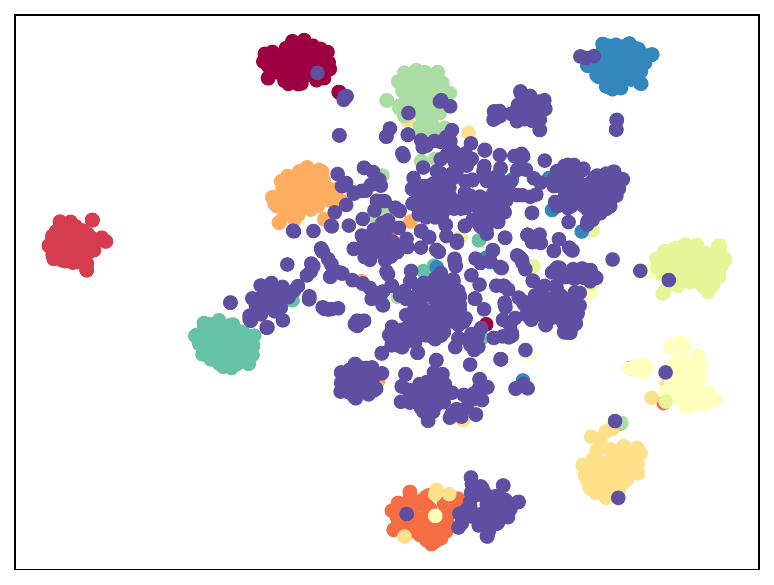}}
    \subfloat[W2D]{\includegraphics[width=.32\linewidth]{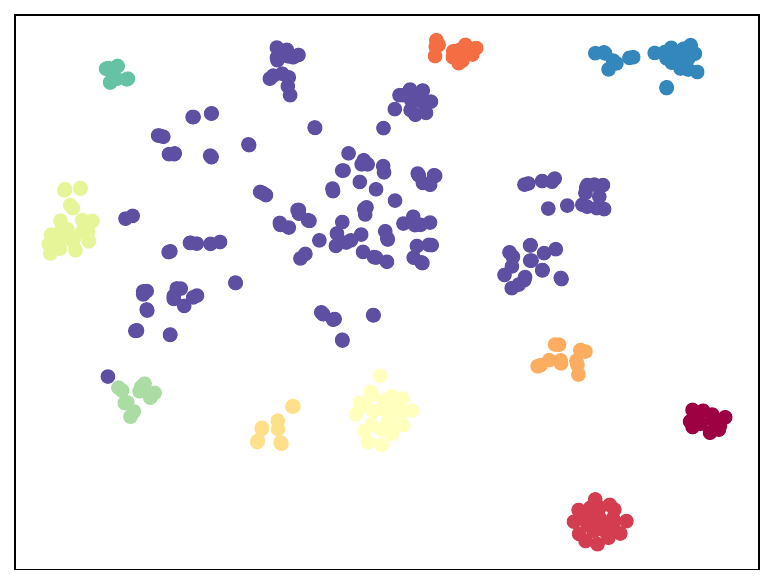}}
    \caption{t-SNE visualization by the baseline in the ODA setting on the Office dataset.}
    \label{fig:oda2}
\end{figure*}

\begin{figure*}[t]
    \centering
    \subfloat[A2D]{\includegraphics[width=.32\linewidth]{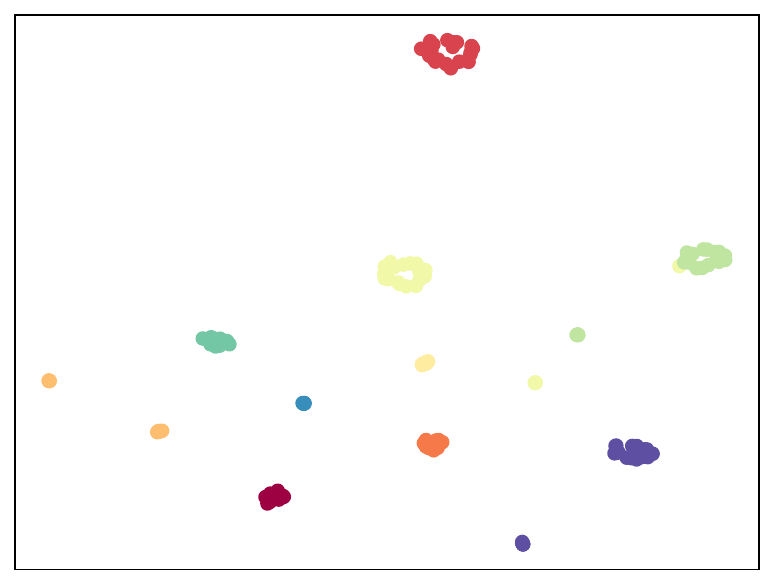}}
    \subfloat[A2W]{\includegraphics[width=.32\linewidth]{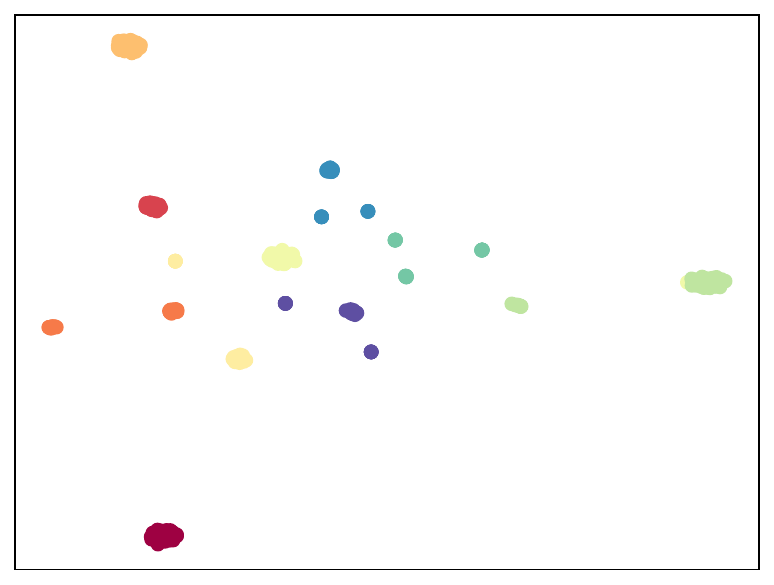}}
    \subfloat[D2A]{\includegraphics[width=.32\linewidth]{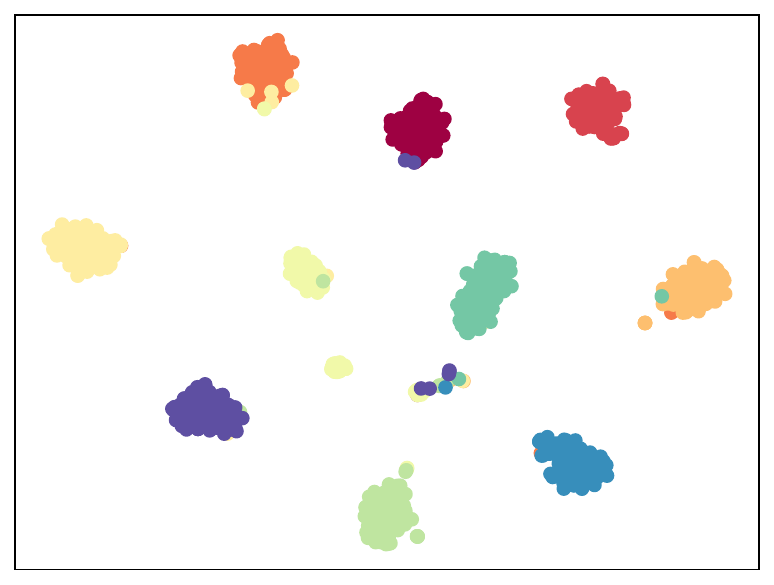}}\\
    \subfloat[D2W]{\includegraphics[width=.32\linewidth]{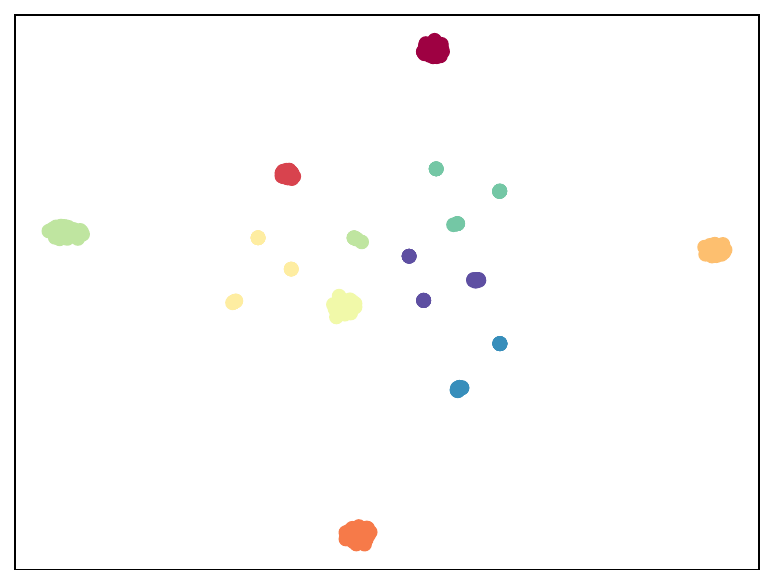}}
    \subfloat[W2A]{\includegraphics[width=.32\linewidth]{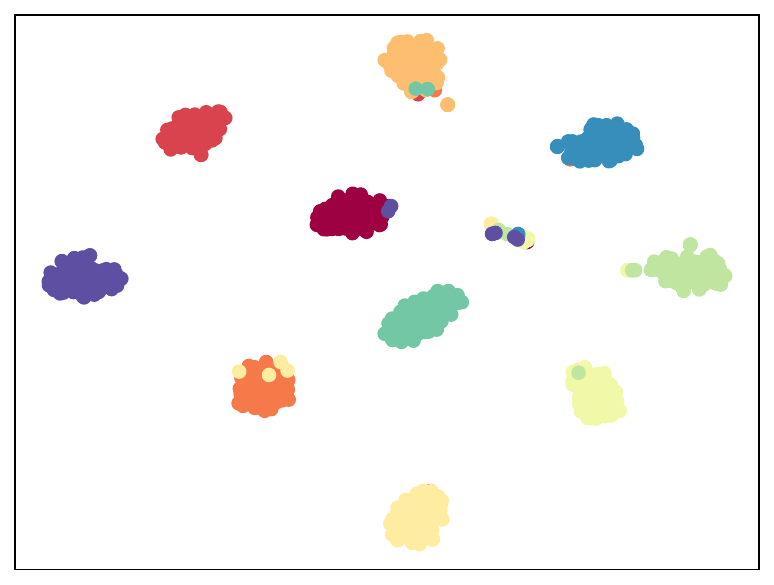}}
    \subfloat[W2D]{\includegraphics[width=.32\linewidth]{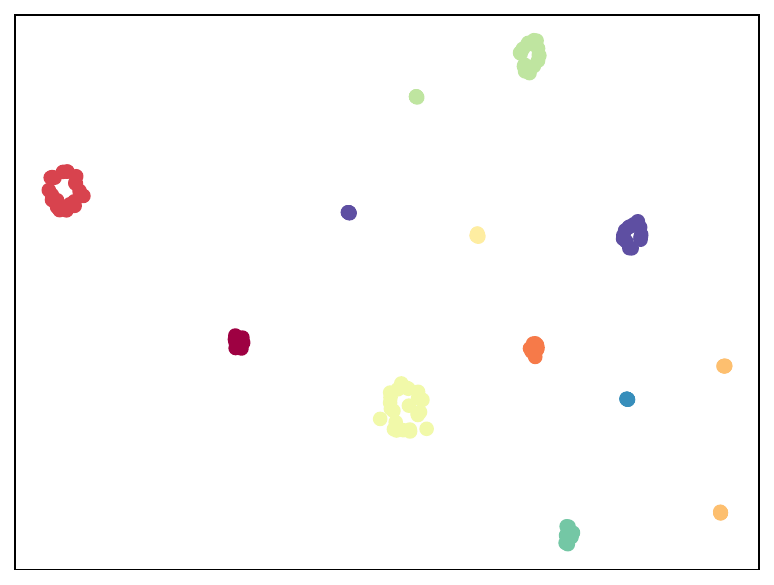}}
    \caption{t-SNE visualization by our method in the PDA setting on the Office dataset.}
    \label{fig:pda1}
\end{figure*}

\begin{figure*}[t]
    \centering
    \subfloat[A2D]{\includegraphics[width=.32\linewidth]{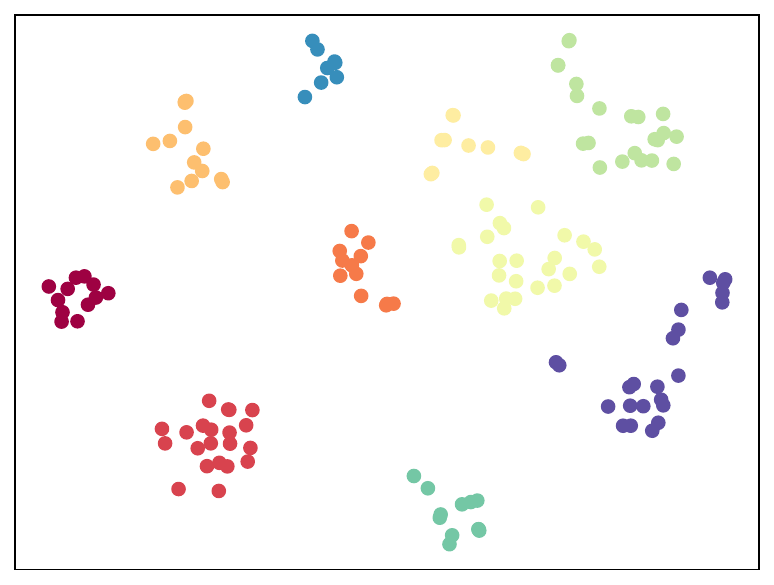}}
    \subfloat[A2W]{\includegraphics[width=.32\linewidth]{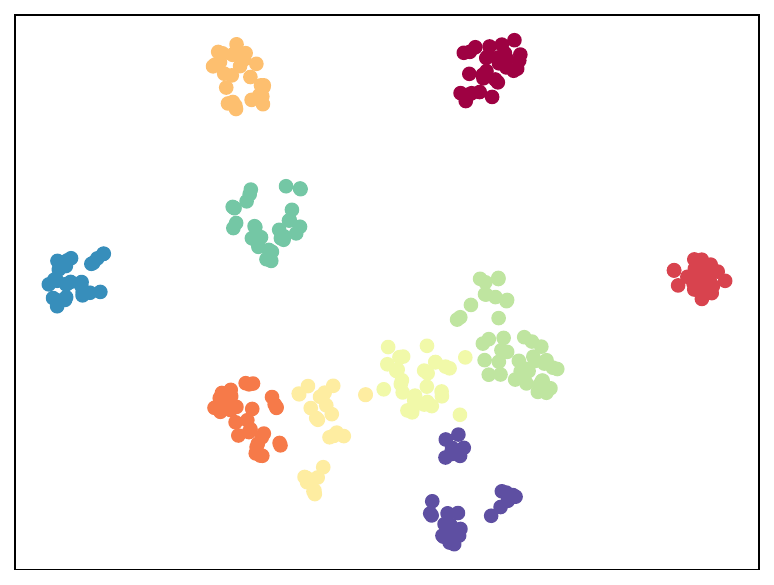}}
    \subfloat[D2A]{\includegraphics[width=.32\linewidth]{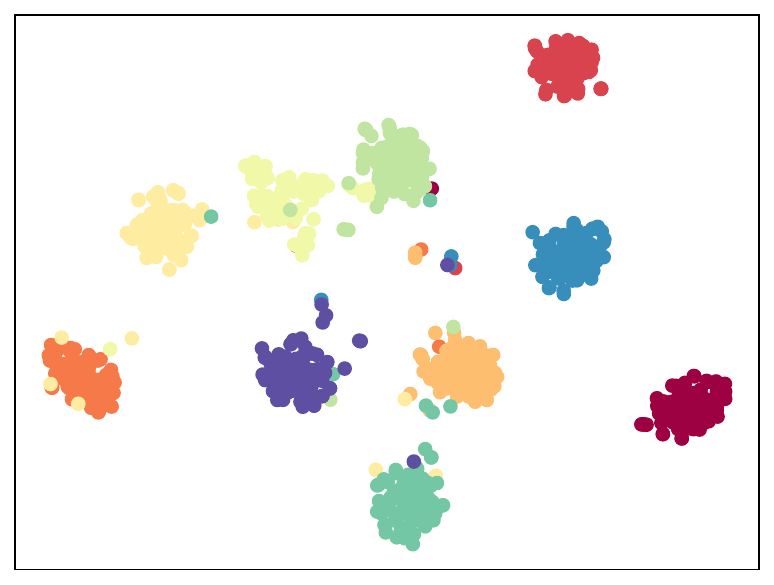}}\\
    \subfloat[D2W]{\includegraphics[width=.32\linewidth]{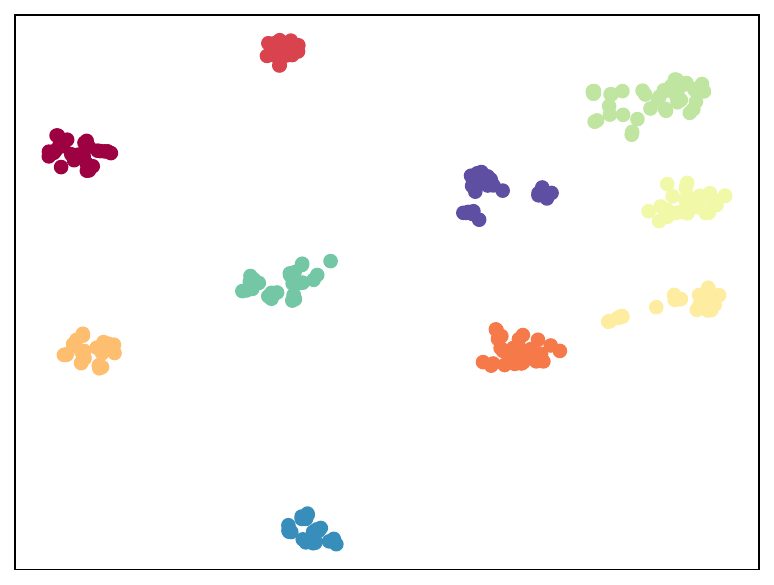}}
    \subfloat[W2A]{\includegraphics[width=.32\linewidth]{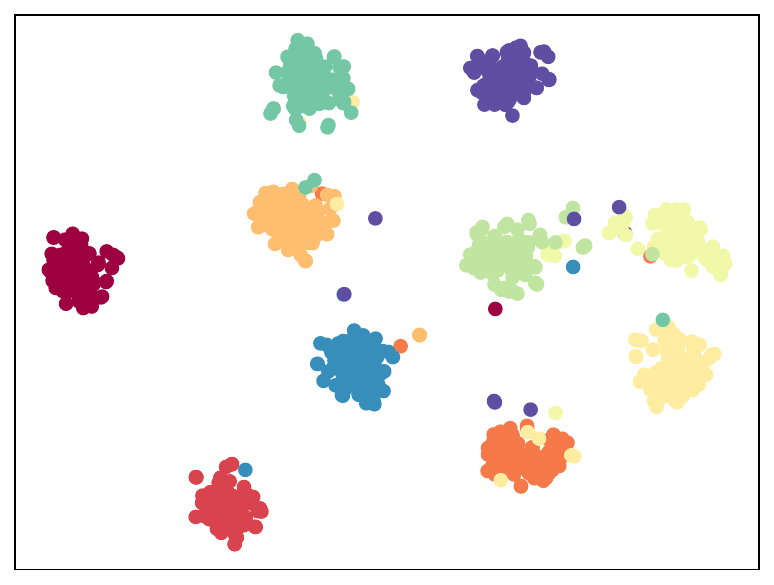}}
    \subfloat[W2D]{\includegraphics[width=.32\linewidth]{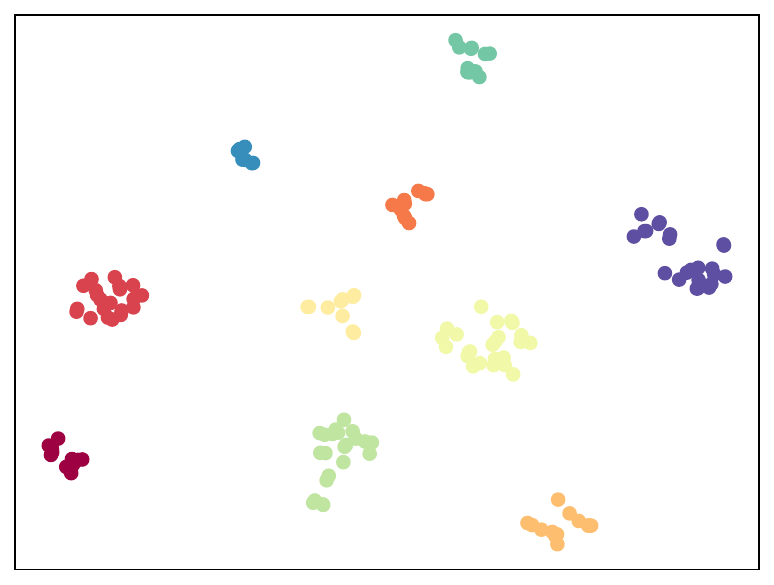}}
    \caption{t-SNE visualization by the baseline in the PDA setting on the Office dataset.}
    \label{fig:pda2}
\end{figure*}

\begin{figure*}[t]
    \centering
    \subfloat[A2D]{\includegraphics[width=.32\linewidth]{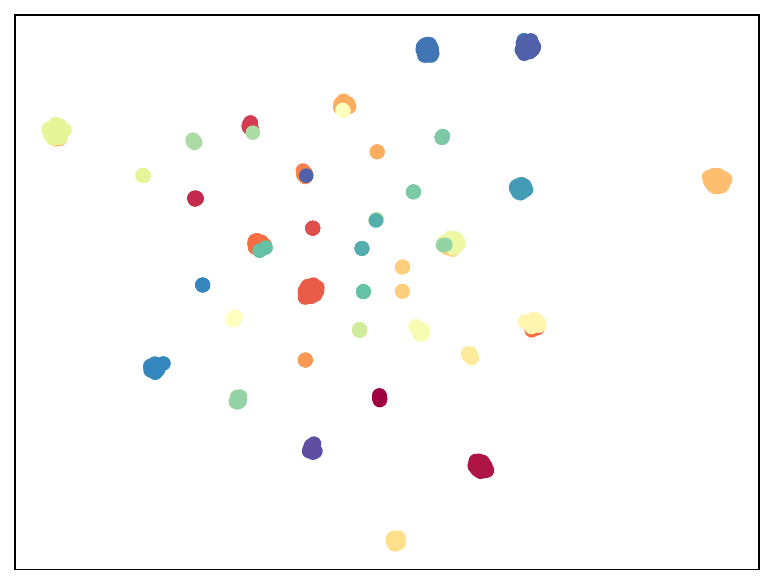}}
    \subfloat[A2W]{\includegraphics[width=.32\linewidth]{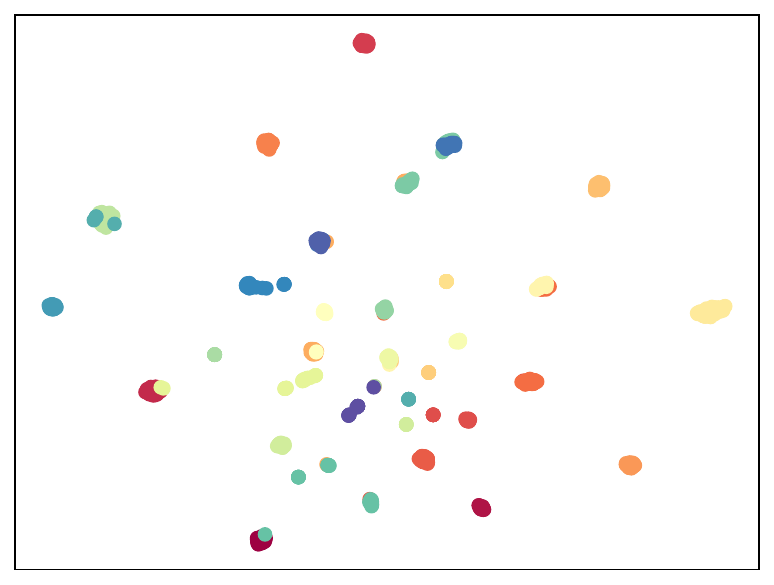}}
    \subfloat[D2A]{\includegraphics[width=.32\linewidth]{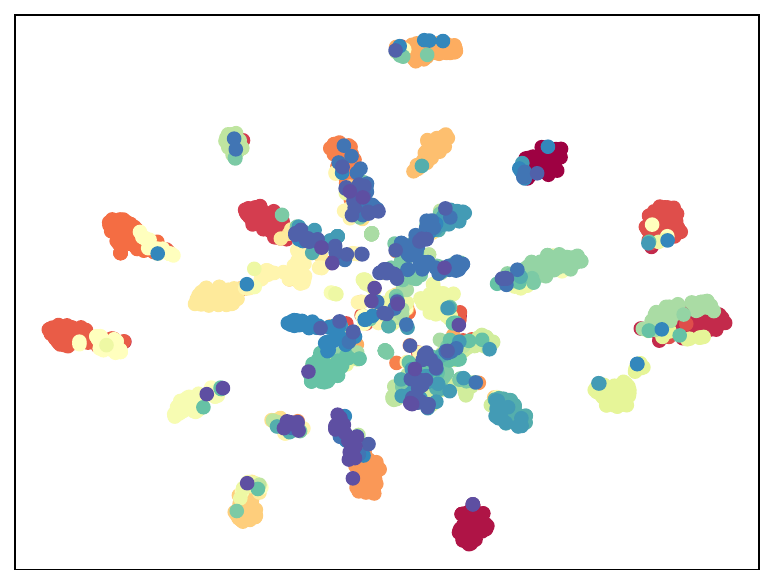}}\\
    \subfloat[D2W]{\includegraphics[width=.32\linewidth]{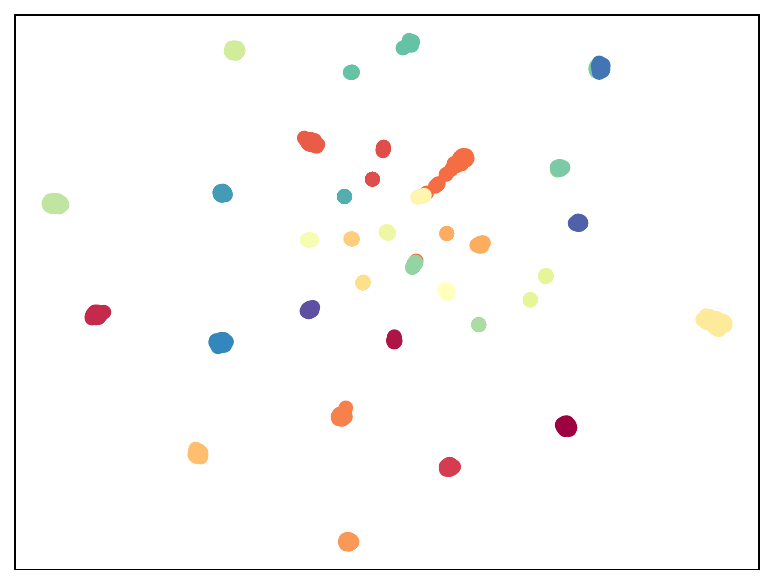}}
    \subfloat[W2A]{\includegraphics[width=.32\linewidth]{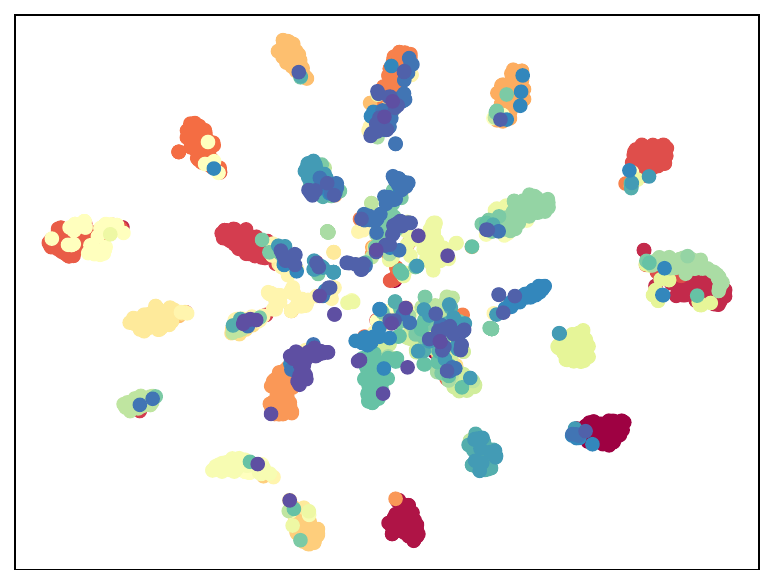}}
    \subfloat[W2D]{\includegraphics[width=.32\linewidth]{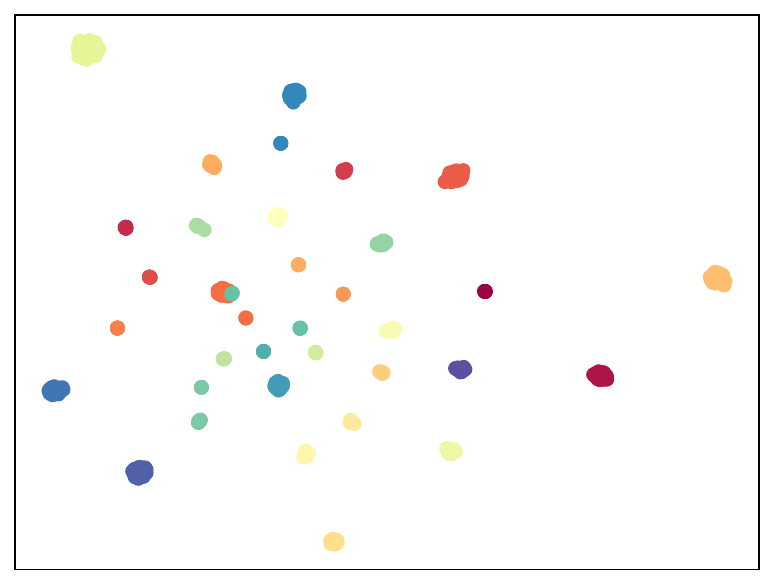}}
    \caption{t-SNE visualization by our method in the CDA setting on the Office dataset.}
    \label{fig:cda1}
\end{figure*}

\begin{figure*}[t]
    \centering
    \subfloat[A2D]{\includegraphics[width=.32\linewidth]{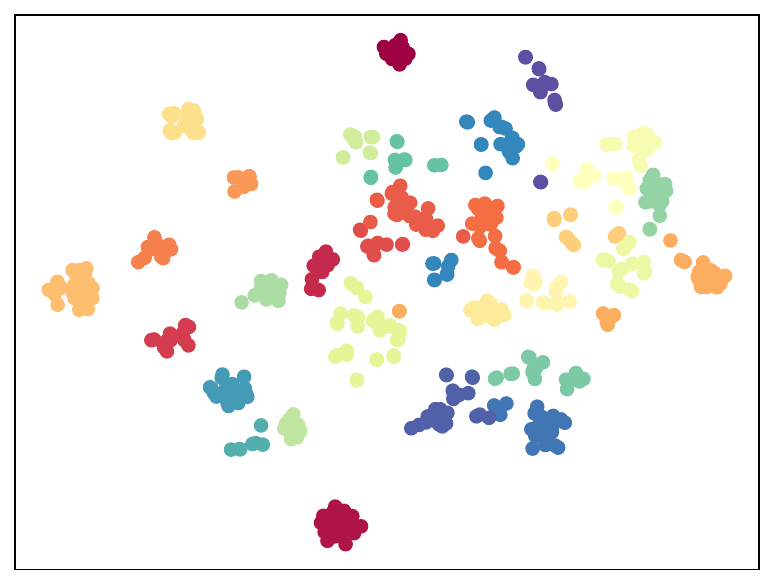}}
    \subfloat[A2W]{\includegraphics[width=.32\linewidth]{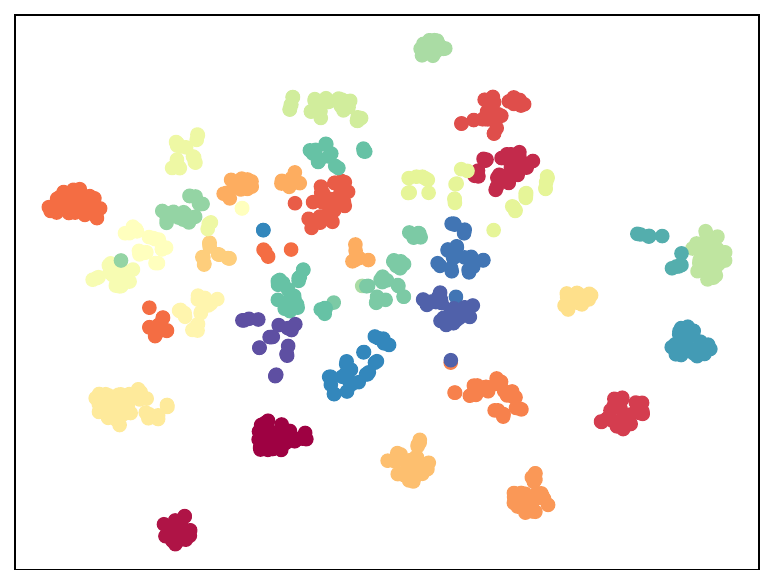}}
    \subfloat[D2A]{\includegraphics[width=.32\linewidth]{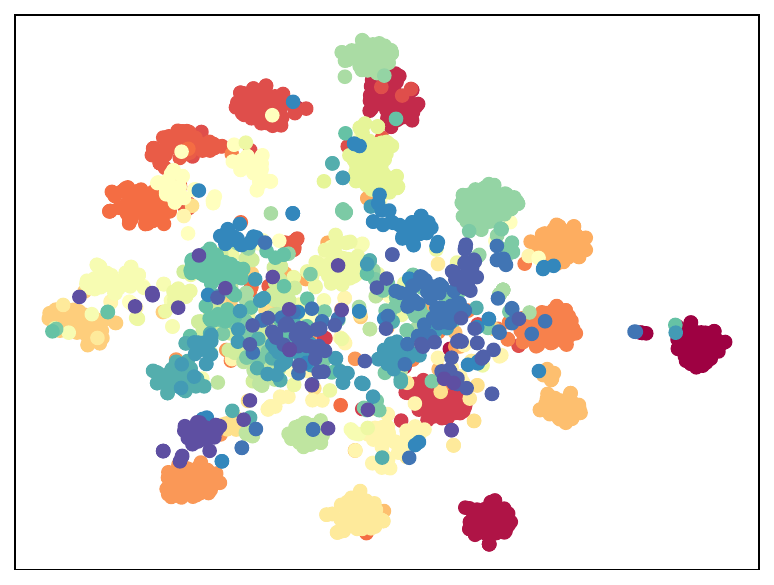}}\\
    \subfloat[D2W]{\includegraphics[width=.32\linewidth]{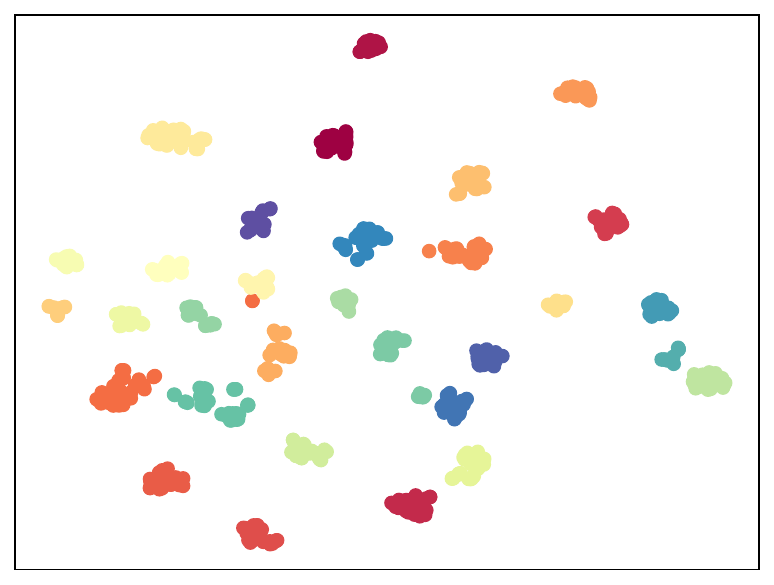}}
    \subfloat[W2A]{\includegraphics[width=.32\linewidth]{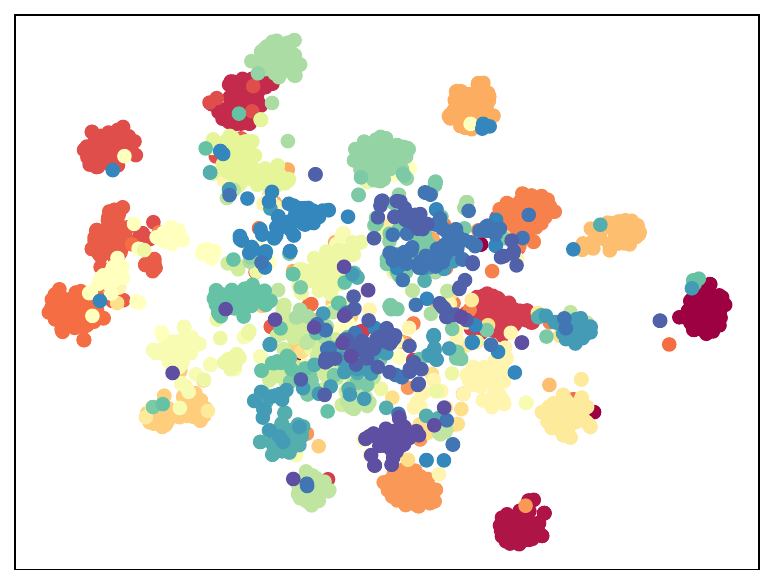}}
    \subfloat[W2D]{\includegraphics[width=.32\linewidth]{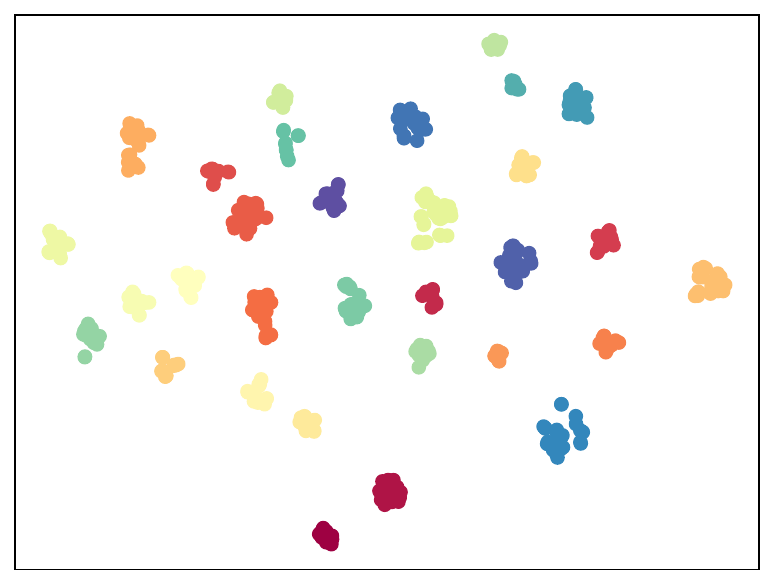}}
    \caption{t-SNE visualization by the baseline in the CDA setting on the Office dataset.}
    \label{fig:cda2}
\end{figure*}

Figures~\ref{fig:opda1}-\ref{fig:cda2} illustrate the comparison of t-SNE visualizations~\cite{van2008visualizing} across all the four settings in UniDA between the baseline and our method.
In the CDA and PDA settings, our method has smaller intra-domain variations, where target samples within a local neighborhood are more likely to share the same class label.
In the ODA and OPDA settings, unknown-class samples (data points in purple) can be approximated by neighborhood invariance in our method to achieve the effect of the clustering algorithm so that they do not interact with known-class samples, while in the baseline unknown-class samples are scattered throughout the embedding space, which is not favorable for unknown-class identification.

\begin{table*}[t]
    \centering
    \small
    \begin{tabular}{c*{4}{m{1.1cm}}c*{4}{m{1.1cm}}}
        \toprule
        \multirow{1}[2]{*}{\textbf{Categories}} & \makebox[1.1cm]{\makecell{\textbf{CDA}\\(31/0/0)}} & \makebox[1.1cm]{\makecell{\textbf{ODA}\\(10/0/11)}} & \makebox[1.1cm]{\makecell{\textbf{OPDA}\\(10/10/11)}} & \makebox[1.1cm]{\makecell{\textbf{PDA}\\(10/21/0)}} &  \multirow{1}[2]{*}{\textbf{Categories}} & \makebox[1.1cm]{\makecell{\textbf{CDA}\\(31/0/0)}} & \makebox[1.1cm]{\makecell{\textbf{ODA}\\(10/0/11)}} & \makebox[1.1cm]{\makecell{\textbf{OPDA}\\(10/10/11)}} & \makebox[1.1cm]{\makecell{\textbf{PDA}\\(10/21/0)}}\\
        \midrule
        back\_pack          & \makebox[1.1cm]{$\RIGHTcircle$} & \makebox[1.1cm]{$\RIGHTcircle$} & \makebox[1.1cm]{$\RIGHTcircle$} & \makebox[1.1cm]{$\RIGHTcircle$}     & mouse               & \makebox[1.1cm]{$\RIGHTcircle$} & \makebox[1.1cm]{$\RIGHTcircle$} & \makebox[1.1cm]{$\RIGHTcircle$} & \makebox[1.1cm]{$\RIGHTcircle$} \\
        bike                & \makebox[1.1cm]{$\RIGHTcircle$} & \makebox[1.1cm]{$\RIGHTcircle$} & \makebox[1.1cm]{$\RIGHTcircle$} & \makebox[1.1cm]{$\RIGHTcircle$}     & mug                 & \makebox[1.1cm]{$\RIGHTcircle$} & \makebox[1.1cm]{$\RIGHTcircle$} & \makebox[1.1cm]{$\RIGHTcircle$} & \makebox[1.1cm]{$\RIGHTcircle$} \\
        bike\_helmet        & \makebox[1.1cm]{$\RIGHTcircle$} &  & \makebox[1.1cm]{$\Circle$} & \makebox[1.1cm]{$\Circle$}                             & paper\_notebook     & \makebox[1.1cm]{$\RIGHTcircle$} &  & \makebox[1.1cm]{$\Circle$} & \makebox[1.1cm]{$\Circle$} \\
        bookcase            & \makebox[1.1cm]{$\RIGHTcircle$} &  & \makebox[1.1cm]{$\Circle$} & \makebox[1.1cm]{$\Circle$}                             & pen                 & \makebox[1.1cm]{$\RIGHTcircle$} & \makebox[1.1cm]{$\CIRCLE$} & \makebox[1.1cm]{$\CIRCLE$} & \makebox[1.1cm]{$\Circle$} \\
        bottle              & \makebox[1.1cm]{$\RIGHTcircle$} &  & \makebox[1.1cm]{$\Circle$} & \makebox[1.1cm]{$\Circle$}                             & phone               & \makebox[1.1cm]{$\RIGHTcircle$} & \makebox[1.1cm]{$\CIRCLE$} & \makebox[1.1cm]{$\CIRCLE$} & \makebox[1.1cm]{$\Circle$} \\
        calculator          & \makebox[1.1cm]{$\RIGHTcircle$} & \makebox[1.1cm]{$\RIGHTcircle$} & \makebox[1.1cm]{$\RIGHTcircle$} & \makebox[1.1cm]{$\RIGHTcircle$}     & printer             & \makebox[1.1cm]{$\RIGHTcircle$} & \makebox[1.1cm]{$\CIRCLE$} & \makebox[1.1cm]{$\CIRCLE$} & \makebox[1.1cm]{$\Circle$} \\
        desk\_chair         & \makebox[1.1cm]{$\RIGHTcircle$} &  & \makebox[1.1cm]{$\Circle$} & \makebox[1.1cm]{$\Circle$}                             & projector           & \makebox[1.1cm]{$\RIGHTcircle$} & \makebox[1.1cm]{$\RIGHTcircle$} & \makebox[1.1cm]{$\RIGHTcircle$} & \makebox[1.1cm]{$\RIGHTcircle$} \\
        desk\_lamp          & \makebox[1.1cm]{$\RIGHTcircle$} &  & \makebox[1.1cm]{$\Circle$} & \makebox[1.1cm]{$\Circle$}                             & punchers            & \makebox[1.1cm]{$\RIGHTcircle$} & \makebox[1.1cm]{$\CIRCLE$} & \makebox[1.1cm]{$\CIRCLE$} & \makebox[1.1cm]{$\Circle$} \\
        desktop\_computer   & \makebox[1.1cm]{$\RIGHTcircle$} &  & \makebox[1.1cm]{$\Circle$} & \makebox[1.1cm]{$\Circle$}                             & ring\_binder        & \makebox[1.1cm]{$\RIGHTcircle$} & \makebox[1.1cm]{$\CIRCLE$} & \makebox[1.1cm]{$\CIRCLE$} & \makebox[1.1cm]{$\Circle$} \\
        file\_cabinet       & \makebox[1.1cm]{$\RIGHTcircle$} &  & \makebox[1.1cm]{$\Circle$} & \makebox[1.1cm]{$\Circle$}                             & ruler               & \makebox[1.1cm]{$\RIGHTcircle$} & \makebox[1.1cm]{$\CIRCLE$} & \makebox[1.1cm]{$\CIRCLE$} & \makebox[1.1cm]{$\Circle$} \\
        headphones          & \makebox[1.1cm]{$\RIGHTcircle$} & \makebox[1.1cm]{$\RIGHTcircle$} & \makebox[1.1cm]{$\RIGHTcircle$} & \makebox[1.1cm]{$\RIGHTcircle$}     & scissors            & \makebox[1.1cm]{$\RIGHTcircle$} & \makebox[1.1cm]{$\CIRCLE$} & \makebox[1.1cm]{$\CIRCLE$} & \makebox[1.1cm]{$\Circle$} \\
        keyboard            & \makebox[1.1cm]{$\RIGHTcircle$} & \makebox[1.1cm]{$\RIGHTcircle$} & \makebox[1.1cm]{$\RIGHTcircle$} & \makebox[1.1cm]{$\RIGHTcircle$}     & speaker             & \makebox[1.1cm]{$\RIGHTcircle$} & \makebox[1.1cm]{$\CIRCLE$} & \makebox[1.1cm]{$\CIRCLE$} & \makebox[1.1cm]{$\Circle$} \\
        laptop\_computer    & \makebox[1.1cm]{$\RIGHTcircle$} & \makebox[1.1cm]{$\RIGHTcircle$} & \makebox[1.1cm]{$\RIGHTcircle$} & \makebox[1.1cm]{$\RIGHTcircle$}     & stapler             & \makebox[1.1cm]{$\RIGHTcircle$} & \makebox[1.1cm]{$\CIRCLE$} & \makebox[1.1cm]{$\CIRCLE$} & \makebox[1.1cm]{$\Circle$} \\
        letter\_tray        & \makebox[1.1cm]{$\RIGHTcircle$} &  & \makebox[1.1cm]{$\Circle$} & \makebox[1.1cm]{$\Circle$}                             & tape\_dispenser     & \makebox[1.1cm]{$\RIGHTcircle$} & \makebox[1.1cm]{$\CIRCLE$} & \makebox[1.1cm]{$\CIRCLE$} & \makebox[1.1cm]{$\Circle$} \\
        mobile\_phone       & \makebox[1.1cm]{$\RIGHTcircle$} &  & \makebox[1.1cm]{$\Circle$} & \makebox[1.1cm]{$\Circle$}                             & trash\_can          & \makebox[1.1cm]{$\RIGHTcircle$} & \makebox[1.1cm]{$\CIRCLE$} & \makebox[1.1cm]{$\CIRCLE$} & \makebox[1.1cm]{$\Circle$} \\
        monitor             & \makebox[1.1cm]{$\RIGHTcircle$} & \makebox[1.1cm]{$\RIGHTcircle$} & \makebox[1.1cm]{$\RIGHTcircle$} & \makebox[1.1cm]{$\RIGHTcircle$}     &                     &  &  &  &  \\
        \bottomrule
    \end{tabular}
    \caption{Split on the Office dataset. $\RIGHTcircle, \Circle, \CIRCLE$ indicate shared, source private and target private categories, respectively.}
    \label{tab:office}
\end{table*}
\begin{table*}[t]
    \centering
    \small
    \begin{tabular}{c*{4}{m{1.1cm}}c*{4}{m{1.1cm}}}
        \toprule
        \multirow{1}[2]{*}{\textbf{Categories}} & \makebox[1.1cm]{\makecell{\textbf{CDA}\\(65/0/0)}} & \makebox[1.1cm]{\makecell{\textbf{ODA}\\(25/0/40)}} & \makebox[1.1cm]{\makecell{\textbf{OPDA}\\(10/5/50)}} & \makebox[1.1cm]{\makecell{\textbf{PDA}\\(25/40/0)}} & \multirow{1}[2]{*}{\textbf{Categories}} & \makebox[1.1cm]{\makecell{\textbf{CDA}\\(65/0/0)}} & \makebox[1.1cm]{\makecell{\textbf{ODA}\\(25/0/40)}} & \makebox[1.1cm]{\makecell{\textbf{OPDA}\\(10/5/50)}} & \makebox[1.1cm]{\makecell{\textbf{PDA}\\(25/40/0)}}\\
        \midrule
        Alarm\_Clock    & \makebox[1.1cm]{$\RIGHTcircle$} & \makebox[1.1cm]{$\RIGHTcircle$} & \makebox[1.1cm]{$\RIGHTcircle$} & \makebox[1.1cm]{$\RIGHTcircle$} & Marker        & \makebox[1.1cm]{$\RIGHTcircle$} & \makebox[1.1cm]{$\CIRCLE$} & \makebox[1.1cm]{$\CIRCLE$} & \makebox[1.1cm]{$\Circle$} \\
        Backpack        & \makebox[1.1cm]{$\RIGHTcircle$} & \makebox[1.1cm]{$\RIGHTcircle$} & \makebox[1.1cm]{$\RIGHTcircle$} & \makebox[1.1cm]{$\RIGHTcircle$} & Monitor       & \makebox[1.1cm]{$\RIGHTcircle$} & \makebox[1.1cm]{$\CIRCLE$} & \makebox[1.1cm]{$\CIRCLE$} & \makebox[1.1cm]{$\Circle$} \\
        Batteries       & \makebox[1.1cm]{$\RIGHTcircle$} & \makebox[1.1cm]{$\RIGHTcircle$} & \makebox[1.1cm]{$\RIGHTcircle$} & \makebox[1.1cm]{$\RIGHTcircle$} & Mop           & \makebox[1.1cm]{$\RIGHTcircle$} & \makebox[1.1cm]{$\CIRCLE$} & \makebox[1.1cm]{$\CIRCLE$} & \makebox[1.1cm]{$\Circle$} \\
        Bed             & \makebox[1.1cm]{$\RIGHTcircle$} & \makebox[1.1cm]{$\RIGHTcircle$} & \makebox[1.1cm]{$\RIGHTcircle$} & \makebox[1.1cm]{$\RIGHTcircle$} & Mouse         & \makebox[1.1cm]{$\RIGHTcircle$} & \makebox[1.1cm]{$\CIRCLE$} & \makebox[1.1cm]{$\CIRCLE$} & \makebox[1.1cm]{$\Circle$} \\
        Bike            & \makebox[1.1cm]{$\RIGHTcircle$} & \makebox[1.1cm]{$\RIGHTcircle$} & \makebox[1.1cm]{$\RIGHTcircle$} & \makebox[1.1cm]{$\RIGHTcircle$} & Mug           & \makebox[1.1cm]{$\RIGHTcircle$} & \makebox[1.1cm]{$\CIRCLE$} & \makebox[1.1cm]{$\CIRCLE$} & \makebox[1.1cm]{$\Circle$} \\
        Bottle          & \makebox[1.1cm]{$\RIGHTcircle$} & \makebox[1.1cm]{$\RIGHTcircle$} & \makebox[1.1cm]{$\RIGHTcircle$} & \makebox[1.1cm]{$\RIGHTcircle$} & Notebook      & \makebox[1.1cm]{$\RIGHTcircle$} & \makebox[1.1cm]{$\CIRCLE$} & \makebox[1.1cm]{$\CIRCLE$} & \makebox[1.1cm]{$\Circle$} \\
        Bucket          & \makebox[1.1cm]{$\RIGHTcircle$} & \makebox[1.1cm]{$\RIGHTcircle$} & \makebox[1.1cm]{$\RIGHTcircle$} & \makebox[1.1cm]{$\RIGHTcircle$} & Oven          & \makebox[1.1cm]{$\RIGHTcircle$} & \makebox[1.1cm]{$\CIRCLE$} & \makebox[1.1cm]{$\CIRCLE$} & \makebox[1.1cm]{$\Circle$} \\
        Calculator      & \makebox[1.1cm]{$\RIGHTcircle$} & \makebox[1.1cm]{$\RIGHTcircle$} & \makebox[1.1cm]{$\RIGHTcircle$} & \makebox[1.1cm]{$\RIGHTcircle$} & Pan           & \makebox[1.1cm]{$\RIGHTcircle$} & \makebox[1.1cm]{$\CIRCLE$} & \makebox[1.1cm]{$\CIRCLE$} & \makebox[1.1cm]{$\Circle$} \\
        Calendar        & \makebox[1.1cm]{$\RIGHTcircle$} & \makebox[1.1cm]{$\RIGHTcircle$} & \makebox[1.1cm]{$\RIGHTcircle$} & \makebox[1.1cm]{$\RIGHTcircle$} & Paper\_Clip   & \makebox[1.1cm]{$\RIGHTcircle$} & \makebox[1.1cm]{$\CIRCLE$} & \makebox[1.1cm]{$\CIRCLE$} & \makebox[1.1cm]{$\Circle$} \\
        Candles         & \makebox[1.1cm]{$\RIGHTcircle$} & \makebox[1.1cm]{$\RIGHTcircle$} & \makebox[1.1cm]{$\RIGHTcircle$} & \makebox[1.1cm]{$\RIGHTcircle$} & Pen           & \makebox[1.1cm]{$\RIGHTcircle$} & \makebox[1.1cm]{$\CIRCLE$} & \makebox[1.1cm]{$\CIRCLE$} & \makebox[1.1cm]{$\Circle$} \\
        Chair           & \makebox[1.1cm]{$\RIGHTcircle$} & \makebox[1.1cm]{$\RIGHTcircle$} & \makebox[1.1cm]{$\Circle$} & \makebox[1.1cm]{$\RIGHTcircle$} & Pencil        & \makebox[1.1cm]{$\RIGHTcircle$} & \makebox[1.1cm]{$\CIRCLE$} & \makebox[1.1cm]{$\CIRCLE$} & \makebox[1.1cm]{$\Circle$} \\
        Clipboards      & \makebox[1.1cm]{$\RIGHTcircle$} & \makebox[1.1cm]{$\RIGHTcircle$} & \makebox[1.1cm]{$\Circle$} & \makebox[1.1cm]{$\RIGHTcircle$} & Postit\_Notes & \makebox[1.1cm]{$\RIGHTcircle$} & \makebox[1.1cm]{$\CIRCLE$} & \makebox[1.1cm]{$\CIRCLE$} & \makebox[1.1cm]{$\Circle$} \\
        Computer        & \makebox[1.1cm]{$\RIGHTcircle$} & \makebox[1.1cm]{$\RIGHTcircle$} & \makebox[1.1cm]{$\Circle$} & \makebox[1.1cm]{$\RIGHTcircle$} & Printer       & \makebox[1.1cm]{$\RIGHTcircle$} & \makebox[1.1cm]{$\CIRCLE$} & \makebox[1.1cm]{$\CIRCLE$} & \makebox[1.1cm]{$\Circle$} \\
        Couch           & \makebox[1.1cm]{$\RIGHTcircle$} & \makebox[1.1cm]{$\RIGHTcircle$} & \makebox[1.1cm]{$\Circle$} & \makebox[1.1cm]{$\RIGHTcircle$} & Push\_Pin     & \makebox[1.1cm]{$\RIGHTcircle$} & \makebox[1.1cm]{$\CIRCLE$} & \makebox[1.1cm]{$\CIRCLE$} & \makebox[1.1cm]{$\Circle$} \\
        Curtains        & \makebox[1.1cm]{$\RIGHTcircle$} & \makebox[1.1cm]{$\RIGHTcircle$} & \makebox[1.1cm]{$\Circle$} & \makebox[1.1cm]{$\RIGHTcircle$} & Radio         & \makebox[1.1cm]{$\RIGHTcircle$} & \makebox[1.1cm]{$\CIRCLE$} & \makebox[1.1cm]{$\CIRCLE$} & \makebox[1.1cm]{$\Circle$} \\
        Desk\_Lamp      & \makebox[1.1cm]{$\RIGHTcircle$} & \makebox[1.1cm]{$\RIGHTcircle$} & \makebox[1.1cm]{$\CIRCLE$} & \makebox[1.1cm]{$\RIGHTcircle$} & Refrigerator  & \makebox[1.1cm]{$\RIGHTcircle$} & \makebox[1.1cm]{$\CIRCLE$} & \makebox[1.1cm]{$\CIRCLE$} & \makebox[1.1cm]{$\Circle$} \\
        Drill           & \makebox[1.1cm]{$\RIGHTcircle$} & \makebox[1.1cm]{$\RIGHTcircle$} & \makebox[1.1cm]{$\CIRCLE$} & \makebox[1.1cm]{$\RIGHTcircle$} & Ruler         & \makebox[1.1cm]{$\RIGHTcircle$} & \makebox[1.1cm]{$\CIRCLE$} & \makebox[1.1cm]{$\CIRCLE$} & \makebox[1.1cm]{$\Circle$} \\
        Eraser          & \makebox[1.1cm]{$\RIGHTcircle$} & \makebox[1.1cm]{$\RIGHTcircle$} & \makebox[1.1cm]{$\CIRCLE$} & \makebox[1.1cm]{$\RIGHTcircle$} & Scissors      & \makebox[1.1cm]{$\RIGHTcircle$} & \makebox[1.1cm]{$\CIRCLE$} & \makebox[1.1cm]{$\CIRCLE$} & \makebox[1.1cm]{$\Circle$} \\
        Exit\_Sign      & \makebox[1.1cm]{$\RIGHTcircle$} & \makebox[1.1cm]{$\RIGHTcircle$} & \makebox[1.1cm]{$\CIRCLE$} & \makebox[1.1cm]{$\RIGHTcircle$} & Screwdriver   & \makebox[1.1cm]{$\RIGHTcircle$} & \makebox[1.1cm]{$\CIRCLE$} & \makebox[1.1cm]{$\CIRCLE$} & \makebox[1.1cm]{$\Circle$} \\
        Fan             & \makebox[1.1cm]{$\RIGHTcircle$} & \makebox[1.1cm]{$\RIGHTcircle$} & \makebox[1.1cm]{$\CIRCLE$} & \makebox[1.1cm]{$\RIGHTcircle$} & Shelf         & \makebox[1.1cm]{$\RIGHTcircle$} & \makebox[1.1cm]{$\CIRCLE$} & \makebox[1.1cm]{$\CIRCLE$} & \makebox[1.1cm]{$\Circle$} \\
        File\_Cabinet   & \makebox[1.1cm]{$\RIGHTcircle$} & \makebox[1.1cm]{$\RIGHTcircle$} & \makebox[1.1cm]{$\CIRCLE$} & \makebox[1.1cm]{$\RIGHTcircle$} & Sink          & \makebox[1.1cm]{$\RIGHTcircle$} & \makebox[1.1cm]{$\CIRCLE$} & \makebox[1.1cm]{$\CIRCLE$} & \makebox[1.1cm]{$\Circle$} \\
        Flipflops       & \makebox[1.1cm]{$\RIGHTcircle$} & \makebox[1.1cm]{$\RIGHTcircle$} & \makebox[1.1cm]{$\CIRCLE$} & \makebox[1.1cm]{$\RIGHTcircle$} & Sneakers      & \makebox[1.1cm]{$\RIGHTcircle$} & \makebox[1.1cm]{$\CIRCLE$} & \makebox[1.1cm]{$\CIRCLE$} & \makebox[1.1cm]{$\Circle$} \\
        Flowers         & \makebox[1.1cm]{$\RIGHTcircle$} & \makebox[1.1cm]{$\RIGHTcircle$} & \makebox[1.1cm]{$\CIRCLE$} & \makebox[1.1cm]{$\RIGHTcircle$} & Soda          & \makebox[1.1cm]{$\RIGHTcircle$} & \makebox[1.1cm]{$\CIRCLE$} & \makebox[1.1cm]{$\CIRCLE$} & \makebox[1.1cm]{$\Circle$} \\
        Folder          & \makebox[1.1cm]{$\RIGHTcircle$} & \makebox[1.1cm]{$\RIGHTcircle$} & \makebox[1.1cm]{$\CIRCLE$} & \makebox[1.1cm]{$\RIGHTcircle$} & Speaker       & \makebox[1.1cm]{$\RIGHTcircle$} & \makebox[1.1cm]{$\CIRCLE$} & \makebox[1.1cm]{$\CIRCLE$} & \makebox[1.1cm]{$\Circle$} \\
        Fork            & \makebox[1.1cm]{$\RIGHTcircle$} & \makebox[1.1cm]{$\RIGHTcircle$} & \makebox[1.1cm]{$\CIRCLE$} & \makebox[1.1cm]{$\RIGHTcircle$} & Spoon         & \makebox[1.1cm]{$\RIGHTcircle$} & \makebox[1.1cm]{$\CIRCLE$} & \makebox[1.1cm]{$\CIRCLE$} & \makebox[1.1cm]{$\Circle$} \\
        Glasses         & \makebox[1.1cm]{$\RIGHTcircle$} & \makebox[1.1cm]{$\CIRCLE$} & \makebox[1.1cm]{$\CIRCLE$} & \makebox[1.1cm]{$\Circle$}      & TV            & \makebox[1.1cm]{$\RIGHTcircle$} & \makebox[1.1cm]{$\CIRCLE$} & \makebox[1.1cm]{$\CIRCLE$} & \makebox[1.1cm]{$\Circle$} \\
        Hammer          & \makebox[1.1cm]{$\RIGHTcircle$} & \makebox[1.1cm]{$\CIRCLE$} & \makebox[1.1cm]{$\CIRCLE$} & \makebox[1.1cm]{$\Circle$}      & Table         & \makebox[1.1cm]{$\RIGHTcircle$} & \makebox[1.1cm]{$\CIRCLE$} & \makebox[1.1cm]{$\CIRCLE$} & \makebox[1.1cm]{$\Circle$} \\
        Helmet          & \makebox[1.1cm]{$\RIGHTcircle$} & \makebox[1.1cm]{$\CIRCLE$} & \makebox[1.1cm]{$\CIRCLE$} & \makebox[1.1cm]{$\Circle$}      & Telephone     & \makebox[1.1cm]{$\RIGHTcircle$} & \makebox[1.1cm]{$\CIRCLE$} & \makebox[1.1cm]{$\CIRCLE$} & \makebox[1.1cm]{$\Circle$} \\
        Kettle          & \makebox[1.1cm]{$\RIGHTcircle$} & \makebox[1.1cm]{$\CIRCLE$} & \makebox[1.1cm]{$\CIRCLE$} & \makebox[1.1cm]{$\Circle$}      & ToothBrush    & \makebox[1.1cm]{$\RIGHTcircle$} & \makebox[1.1cm]{$\CIRCLE$} & \makebox[1.1cm]{$\CIRCLE$} & \makebox[1.1cm]{$\Circle$} \\
        Keyboard        & \makebox[1.1cm]{$\RIGHTcircle$} & \makebox[1.1cm]{$\CIRCLE$} & \makebox[1.1cm]{$\CIRCLE$} & \makebox[1.1cm]{$\Circle$}      & Toys          & \makebox[1.1cm]{$\RIGHTcircle$} & \makebox[1.1cm]{$\CIRCLE$} & \makebox[1.1cm]{$\CIRCLE$} & \makebox[1.1cm]{$\Circle$} \\
        Knives          & \makebox[1.1cm]{$\RIGHTcircle$} & \makebox[1.1cm]{$\CIRCLE$} & \makebox[1.1cm]{$\CIRCLE$} & \makebox[1.1cm]{$\Circle$}      & Trash\_Can    & \makebox[1.1cm]{$\RIGHTcircle$} & \makebox[1.1cm]{$\CIRCLE$} & \makebox[1.1cm]{$\CIRCLE$} & \makebox[1.1cm]{$\Circle$} \\
        Lamp\_Shade     & \makebox[1.1cm]{$\RIGHTcircle$} & \makebox[1.1cm]{$\CIRCLE$} & \makebox[1.1cm]{$\CIRCLE$} & \makebox[1.1cm]{$\Circle$}      & Webcam        & \makebox[1.1cm]{$\RIGHTcircle$} & \makebox[1.1cm]{$\CIRCLE$} & \makebox[1.1cm]{$\CIRCLE$} & \makebox[1.1cm]{$\Circle$} \\
        Laptop          & \makebox[1.1cm]{$\RIGHTcircle$} & \makebox[1.1cm]{$\CIRCLE$} & \makebox[1.1cm]{$\CIRCLE$} & \makebox[1.1cm]{$\Circle$}      &               &  &  &  &  \\
        \bottomrule
    \end{tabular}
    \caption{Split on the OfficeHome dataset. $\RIGHTcircle, \Circle, \CIRCLE$ indicate shared, source private and target private categories, respectively.}
    \label{tab:officehome}
\end{table*}
\begin{table*}[t]
    \centering
    \small
    \begin{tabular}{c*{4}{m{1.1cm}}c*{4}{m{1.1cm}}}
        \toprule
        \multirow{1}[2]{*}{\textbf{Categories}} & \makebox[1.1cm]{\makecell{\textbf{CDA}\\(12/0/0)}} & \makebox[1.1cm]{\makecell{\textbf{ODA}\\(6/0/6)}} & \makebox[1.1cm]{\makecell{\textbf{OPDA}\\(6/3/3)}} & \makebox[1.1cm]{\makecell{\textbf{PDA}\\(6/6/0)}} & \multirow{1}[2]{*}{\textbf{Categories}} & \makebox[1.1cm]{\makecell{\textbf{CDA}\\(12/0/0)}} & \makebox[1.1cm]{\makecell{\textbf{ODA}\\(6/0/6)}} & \makebox[1.1cm]{\makecell{\textbf{OPDA}\\(6/3/3)}} & \makebox[1.1cm]{\makecell{\textbf{PDA}\\(6/6/0)}}\\
        \midrule
        aeroplane   & \makebox[1.1cm]{$\RIGHTcircle$} & \makebox[1.1cm]{$\RIGHTcircle$} & \makebox[1.1cm]{$\RIGHTcircle$} & \makebox[1.1cm]{$\RIGHTcircle$} & motorcycle    & \makebox[1.1cm]{$\RIGHTcircle$} & \makebox[1.1cm]{$\CIRCLE$} & \makebox[1.1cm]{$\Circle$} & \makebox[1.1cm]{$\Circle$} \\
        bicycle     & \makebox[1.1cm]{$\RIGHTcircle$} & \makebox[1.1cm]{$\RIGHTcircle$} & \makebox[1.1cm]{$\RIGHTcircle$} & \makebox[1.1cm]{$\RIGHTcircle$} & person        & \makebox[1.1cm]{$\RIGHTcircle$} & \makebox[1.1cm]{$\CIRCLE$} & \makebox[1.1cm]{$\Circle$} & \makebox[1.1cm]{$\Circle$} \\
        bus         & \makebox[1.1cm]{$\RIGHTcircle$} & \makebox[1.1cm]{$\RIGHTcircle$} & \makebox[1.1cm]{$\RIGHTcircle$} & \makebox[1.1cm]{$\RIGHTcircle$} & plant         & \makebox[1.1cm]{$\RIGHTcircle$} & \makebox[1.1cm]{$\CIRCLE$} & \makebox[1.1cm]{$\Circle$} & \makebox[1.1cm]{$\Circle$} \\
        car         & \makebox[1.1cm]{$\RIGHTcircle$} & \makebox[1.1cm]{$\RIGHTcircle$} & \makebox[1.1cm]{$\RIGHTcircle$} & \makebox[1.1cm]{$\RIGHTcircle$} & skateboard    & \makebox[1.1cm]{$\RIGHTcircle$} & \makebox[1.1cm]{$\CIRCLE$} & \makebox[1.1cm]{$\CIRCLE$} & \makebox[1.1cm]{$\Circle$} \\
        horse       & \makebox[1.1cm]{$\RIGHTcircle$} & \makebox[1.1cm]{$\RIGHTcircle$} & \makebox[1.1cm]{$\RIGHTcircle$} & \makebox[1.1cm]{$\RIGHTcircle$} & train         & \makebox[1.1cm]{$\RIGHTcircle$} & \makebox[1.1cm]{$\CIRCLE$} & \makebox[1.1cm]{$\CIRCLE$} & \makebox[1.1cm]{$\Circle$} \\
        knife       & \makebox[1.1cm]{$\RIGHTcircle$} & \makebox[1.1cm]{$\RIGHTcircle$} & \makebox[1.1cm]{$\RIGHTcircle$} & \makebox[1.1cm]{$\RIGHTcircle$} & truck         & \makebox[1.1cm]{$\RIGHTcircle$} & \makebox[1.1cm]{$\CIRCLE$} & \makebox[1.1cm]{$\CIRCLE$} & \makebox[1.1cm]{$\Circle$} \\
        \bottomrule
    \end{tabular}
    \caption{Split on the VisDA dataset. $\RIGHTcircle, \Circle, \CIRCLE$ indicate shared, source private and target private categories, respectively.}
    \label{tab:visda}
\end{table*}
\begin{table*}[t]
    \centering
    \small
    \begin{tabular}{c *{20}{m{0.32cm}} c}
        \toprule
        \multirow{2}[2]{*}{\textbf{Method}} & \multicolumn{7}{c}{\textbf{Office(10/0/11)}} & \multicolumn{13}{c}{\textbf{OfficeHome(25/0/40)}} & \multirow{2}[2]{*}{\textbf{\makecell{VisDA\\(6/0/6)}}} \\
        \cmidrule(lr){2-8}\cmidrule(lr){9-21}
        & \makebox[0.32cm][c]{A2D}  & \makebox[0.32cm][c]{A2W}  & \makebox[0.32cm][c]{D2A}  & \makebox[0.32cm][c]{D2W}  & \makebox[0.32cm][c]{W2A}  & \makebox[0.32cm][c]{W2D}  & \makebox[0.32cm][c]{Avg} 
        & \makebox[0.32cm][c]{A2C}  & \makebox[0.32cm][c]{A2P}  & \makebox[0.32cm][c]{A2R}  & \makebox[0.32cm][c]{C2A}  & \makebox[0.32cm][c]{C2P}  & \makebox[0.32cm][c]{C2R}  & \makebox[0.32cm][c]{P2A}  & \makebox[0.32cm][c]{P2C}  & \makebox[0.32cm][c]{P2R}  & \makebox[0.32cm][c]{R2A}  & \makebox[0.32cm][c]{R2C}  & \makebox[0.32cm][c]{R2P}  & \makebox[0.32cm][c]{Avg} & \\
        \midrule
        UAN & \makebox[0.32cm][c]{38.9} & \makebox[0.32cm][c]{46.8} & \makebox[0.32cm][c]{68.0} & \makebox[0.32cm][c]{68.8} & \makebox[0.32cm][c]{54.9} & \makebox[0.32cm][c]{53.0} & \makebox[0.32cm][c]{55.1} & \makebox[0.32cm][c]{40.3} & \makebox[0.32cm][c]{41.5} & \makebox[0.32cm][c]{46.1} & \makebox[0.32cm][c]{53.2} & \makebox[0.32cm][c]{48.0} & \makebox[0.32cm][c]{53.7} & \makebox[0.32cm][c]{40.6} & \makebox[0.32cm][c]{39.8} & \makebox[0.32cm][c]{52.5} & \makebox[0.32cm][c]{53.6} & \makebox[0.32cm][c]{43.7} & \makebox[0.32cm][c]{56.9} & \makebox[0.32cm][c]{47.5} & \makebox[0.32cm][c]{51.9} \\
        CMU & \makebox[0.32cm][c]{52.6} & \makebox[0.32cm][c]{55.7} & \makebox[0.32cm][c]{76.5} & \makebox[0.32cm][c]{75.9} & \makebox[0.32cm][c]{65.8} & \makebox[0.32cm][c]{64.7} & \makebox[0.32cm][c]{65.2} & \makebox[0.32cm][c]{45.1} & \makebox[0.32cm][c]{48.3} & \makebox[0.32cm][c]{51.7} & \makebox[0.32cm][c]{58.9} & \makebox[0.32cm][c]{55.4} & \makebox[0.32cm][c]{61.2} & \makebox[0.32cm][c]{46.5} & \makebox[0.32cm][c]{43.8} & \makebox[0.32cm][c]{58.0} & \makebox[0.32cm][c]{58.6} & \makebox[0.32cm][c]{50.1} & \makebox[0.32cm][c]{61.8} & \makebox[0.32cm][c]{53.3} & \makebox[0.32cm][c]{54.2} \\
        DANCE & \makebox[0.32cm][c]{84.9} & \makebox[0.32cm][c]{78.8} & \makebox[0.32cm][c]{79.1} & \makebox[0.32cm][c]{78.8} & \makebox[0.32cm][c]{68.3} & \makebox[0.32cm][c]{88.9} & \makebox[0.32cm][c]{79.8} & \makebox[0.32cm][c]{61.9} & \makebox[0.32cm][c]{61.3} & \makebox[0.32cm][c]{63.7} & \makebox[0.32cm][c]{64.2} & \makebox[0.32cm][c]{58.6} & \makebox[0.32cm][c]{62.6} & \makebox[0.32cm][c]{67.4} & \makebox[0.32cm][c]{61.0} & \makebox[0.32cm][c]{65.5} & \makebox[0.32cm][c]{65.9} & \makebox[0.32cm][c]{61.3} & \makebox[0.32cm][c]{64.2} & \makebox[0.32cm][c]{63.1} & \makebox[0.32cm][c]{67.5} \\
        DCC & \makebox[0.32cm][c]{58.3} & \makebox[0.32cm][c]{54.8} & \makebox[0.32cm][c]{67.2} & \makebox[0.32cm][c]{89.4} & \makebox[0.32cm][c]{85.3} & \makebox[0.32cm][c]{80.9} & \makebox[0.32cm][c]{72.7} & \makebox[0.32cm][c]{56.1} & \makebox[0.32cm][c]{67.5} & \makebox[0.32cm][c]{66.7} & \makebox[0.32cm][c]{49.6} & \makebox[0.32cm][c]{66.5} & \makebox[0.32cm][c]{64.0} & \makebox[0.32cm][c]{55.8} & \makebox[0.32cm][c]{53.0} & \makebox[0.32cm][c]{70.5} & \makebox[0.32cm][c]{61.6} & \makebox[0.32cm][c]{57.2} & \makebox[0.32cm][c]{71.9} & \makebox[0.32cm][c]{61.7} & \makebox[0.32cm][c]{59.6} \\
        GATE & \makebox[0.32cm][c]{88.4} & \makebox[0.32cm][c]{86.5} & \makebox[0.32cm][c]{84.2} & \makebox[0.32cm][c]{95.0} & \makebox[0.32cm][c]{86.1} & \makebox[0.32cm][c]{96.7} & \makebox[0.32cm][c]{89.5} & \makebox[0.32cm][c]{63.8} & \makebox[0.32cm][c]{70.5} & \makebox[0.32cm][c]{75.8} & \makebox[0.32cm][c]{66.4} & \makebox[0.32cm][c]{67.9} & \makebox[0.32cm][c]{71.7} & \makebox[0.32cm][c]{67.3} & \makebox[0.32cm][c]{61.5} & \makebox[0.32cm][c]{76.0} & \makebox[0.32cm][c]{70.4} & \makebox[0.32cm][c]{61.8} & \makebox[0.32cm][c]{75.1} & \makebox[0.32cm][c]{69.0} & \makebox[0.32cm][c]{70.8} \\
        TNT & \makebox[0.32cm][c]{85.8} & \makebox[0.32cm][c]{82.3} & \makebox[0.32cm][c]{80.7} & \makebox[0.32cm][c]{91.2} & \makebox[0.32cm][c]{81.5} & \makebox[0.32cm][c]{96.2} & \makebox[0.32cm][c]{86.3} & \makebox[0.32cm][c]{63.4} & \makebox[0.32cm][c]{67.9} & \makebox[0.32cm][c]{74.9} & \makebox[0.32cm][c]{65.7} & \makebox[0.32cm][c]{67.1} & \makebox[0.32cm][c]{68.3} & \makebox[0.32cm][c]{64.5} & \makebox[0.32cm][c]{58.1} & \makebox[0.32cm][c]{73.2} & \makebox[0.32cm][c]{67.8} & \makebox[0.32cm][c]{61.9} & \makebox[0.32cm][c]{74.5} & \makebox[0.32cm][c]{67.3} & \makebox[0.32cm][c]{\textbf{71.6}} \\
        NCAL & \makebox[0.32cm][c]{84.0} & \makebox[0.32cm][c]{93.4} & \makebox[0.32cm][c]{93.4} & \makebox[0.32cm][c]{85.4} & \makebox[0.32cm][c]{89.0} & \makebox[0.32cm][c]{87.2} & \makebox[0.32cm][c]{88.7} & \makebox[0.32cm][c]{64.2} & \makebox[0.32cm][c]{74.1} & \makebox[0.32cm][c]{80.5} & \makebox[0.32cm][c]{68.1} & \makebox[0.32cm][c]{72.5} & \makebox[0.32cm][c]{77.0} & \makebox[0.32cm][c]{66.9} & \makebox[0.32cm][c]{58.1} & \makebox[0.32cm][c]{79.1} & \makebox[0.32cm][c]{74.6} & \makebox[0.32cm][c]{63.5} & \makebox[0.32cm][c]{79.6} & \makebox[0.32cm][c]{\textbf{71.5}} & \makebox[0.32cm][c]{69.1} \\
        \midrule
        Baseline & \makebox[0.32cm][c]{88.2} & \makebox[0.32cm][c]{88.7} & \makebox[0.32cm][c]{86.9} & \makebox[0.32cm][c]{97.8} & \makebox[0.32cm][c]{89.4} & \makebox[0.32cm][c]{98.8} & \makebox[0.32cm][c]{91.6} & \makebox[0.32cm][c]{58.7} & \makebox[0.32cm][c]{66.5} & \makebox[0.32cm][c]{70.4} & \makebox[0.32cm][c]{61.5} & \makebox[0.32cm][c]{65.4} & \makebox[0.32cm][c]{68.4} & \makebox[0.32cm][c]{60.4} & \makebox[0.32cm][c]{53.5} & \makebox[0.32cm][c]{70.0} & \makebox[0.32cm][c]{68.2} & \makebox[0.32cm][c]{59.1} & \makebox[0.32cm][c]{67.0} & \makebox[0.32cm][c]{64.1} & \makebox[0.32cm][c]{61.6} \\
        Ours & \makebox[0.32cm][c]{93.0} & \makebox[0.32cm][c]{91.9} & \makebox[0.32cm][c]{86.9} & \makebox[0.32cm][c]{98.1} & \makebox[0.32cm][c]{87.5} & \makebox[0.32cm][c]{99.5} & \makebox[0.32cm][c]{\textbf{92.8}} & \makebox[0.32cm][c]{61.3} & \makebox[0.32cm][c]{69.9} & \makebox[0.32cm][c]{74.4} & \makebox[0.32cm][c]{63.1} & \makebox[0.32cm][c]{68.2} & \makebox[0.32cm][c]{70.4} & \makebox[0.32cm][c]{62.0} & \makebox[0.32cm][c]{59.9} & \makebox[0.32cm][c]{72.4} & \makebox[0.32cm][c]{69.1} & \makebox[0.32cm][c]{62.6} & \makebox[0.32cm][c]{71.1} & \makebox[0.32cm][c]{67.0} & \makebox[0.32cm][c]{63.9} \\
        \bottomrule
    \end{tabular}
    \caption{H-score comparison in the ODA setting. Some results are referred to previous work~\cite{chen2022geometric}.}
    \label{tab:oda}
\end{table*}
\begin{table*}[t]
    \centering
    \small
    \begin{tabular}{c *{20}{m{0.32cm}} c}
        \toprule
        \multirow{2}[2]{*}{\textbf{Method}} & \multicolumn{7}{c}{\textbf{Office(10/21/0)}} & \multicolumn{13}{c}{\textbf{OfficeHome(25/40/0)}} & \multirow{2}[2]{*}{\textbf{\makecell{VisDA\\(6/6/0)}}} \\
        \cmidrule(lr){2-8}\cmidrule(lr){9-21}
        & \makebox[0.32cm][c]{A2D}  & \makebox[0.32cm][c]{A2W}  & \makebox[0.32cm][c]{D2A}  & \makebox[0.32cm][c]{D2W}  & \makebox[0.32cm][c]{W2A}  & \makebox[0.32cm][c]{W2D}  & \makebox[0.32cm][c]{Avg} 
        & \makebox[0.32cm][c]{A2C}  & \makebox[0.32cm][c]{A2P}  & \makebox[0.32cm][c]{A2R}  & \makebox[0.32cm][c]{C2A}  & \makebox[0.32cm][c]{C2P}  & \makebox[0.32cm][c]{C2R}  & \makebox[0.32cm][c]{P2A}  & \makebox[0.32cm][c]{P2C}  & \makebox[0.32cm][c]{P2R}  & \makebox[0.32cm][c]{R2A}  & \makebox[0.32cm][c]{R2C}  & \makebox[0.32cm][c]{R2P}  & \makebox[0.32cm][c]{Avg} & \\
        \midrule
        UAN & \makebox[0.32cm][c]{79.7} & \makebox[0.32cm][c]{76.8} & \makebox[0.32cm][c]{82.7} & \makebox[0.32cm][c]{93.4} & \makebox[0.32cm][c]{83.7} & \makebox[0.32cm][c]{98.3} & \makebox[0.32cm][c]{85.8} & \makebox[0.32cm][c]{24.5} & \makebox[0.32cm][c]{35.0} & \makebox[0.32cm][c]{41.5} & \makebox[0.32cm][c]{34.7} & \makebox[0.32cm][c]{32.3} & \makebox[0.32cm][c]{32.7} & \makebox[0.32cm][c]{32.7} & \makebox[0.32cm][c]{21.1} & \makebox[0.32cm][c]{43.0} & \makebox[0.32cm][c]{39.7} & \makebox[0.32cm][c]{26.6} & \makebox[0.32cm][c]{46.0} & \makebox[0.32cm][c]{34.2} & \makebox[0.32cm][c]{39.7} \\
        CMU & \makebox[0.32cm][c]{84.1} & \makebox[0.32cm][c]{84.2} & \makebox[0.32cm][c]{69.2} & \makebox[0.32cm][c]{97.2} & \makebox[0.32cm][c]{66.8} & \makebox[0.32cm][c]{98.8} & \makebox[0.32cm][c]{83.4} & \makebox[0.32cm][c]{50.9} & \makebox[0.32cm][c]{74.2} & \makebox[0.32cm][c]{78.4} & \makebox[0.32cm][c]{62.2} & \makebox[0.32cm][c]{64.1} & \makebox[0.32cm][c]{72.5} & \makebox[0.32cm][c]{63.5} & \makebox[0.32cm][c]{47.9} & \makebox[0.32cm][c]{78.3} & \makebox[0.32cm][c]{72.4} & \makebox[0.32cm][c]{54.7} & \makebox[0.32cm][c]{78.9} & \makebox[0.32cm][c]{66.5} & \makebox[0.32cm][c]{65.5} \\
        DANCE & \makebox[0.32cm][c]{77.1} & \makebox[0.32cm][c]{71.2} & \makebox[0.32cm][c]{83.7} & \makebox[0.32cm][c]{94.6} & \makebox[0.32cm][c]{92.6} & \makebox[0.32cm][c]{96.8} & \makebox[0.32cm][c]{86.0} & \makebox[0.32cm][c]{53.6} & \makebox[0.32cm][c]{73.2} & \makebox[0.32cm][c]{84.9} & \makebox[0.32cm][c]{70.8} & \makebox[0.32cm][c]{67.3} & \makebox[0.32cm][c]{82.6} & \makebox[0.32cm][c]{70.0} & \makebox[0.32cm][c]{50.9} & \makebox[0.32cm][c]{84.8} & \makebox[0.32cm][c]{77.0} & \makebox[0.32cm][c]{55.9} & \makebox[0.32cm][c]{81.8} & \makebox[0.32cm][c]{71.1} & \makebox[0.32cm][c]{73.7} \\
        DCC & \makebox[0.32cm][c]{87.3} & \makebox[0.32cm][c]{81.3} & \makebox[0.32cm][c]{95.4} & \makebox[0.32cm][c]{100.0} & \makebox[0.32cm][c]{95.5} & \makebox[0.32cm][c]{100.0} & \makebox[0.32cm][c]{93.3} & \makebox[0.32cm][c]{54.2} & \makebox[0.32cm][c]{47.5} & \makebox[0.32cm][c]{57.5} & \makebox[0.32cm][c]{83.8} & \makebox[0.32cm][c]{71.6} & \makebox[0.32cm][c]{86.2} & \makebox[0.32cm][c]{63.7} & \makebox[0.32cm][c]{65.0} & \makebox[0.32cm][c]{75.2} & \makebox[0.32cm][c]{85.5} & \makebox[0.32cm][c]{78.2} & \makebox[0.32cm][c]{82.6} & \makebox[0.32cm][c]{70.9} & \makebox[0.32cm][c]{72.4} \\
        GATE & \makebox[0.32cm][c]{89.5} & \makebox[0.32cm][c]{86.2} & \makebox[0.32cm][c]{93.5} & \makebox[0.32cm][c]{100.0} & \makebox[0.32cm][c]{94.4} & \makebox[0.32cm][c]{98.6} & \makebox[0.32cm][c]{93.7} & \makebox[0.32cm][c]{55.8} & \makebox[0.32cm][c]{75.9} & \makebox[0.32cm][c]{85.3} & \makebox[0.32cm][c]{73.6} & \makebox[0.32cm][c]{70.2} & \makebox[0.32cm][c]{83.0} & \makebox[0.32cm][c]{72.1} & \makebox[0.32cm][c]{59.5} & \makebox[0.32cm][c]{84.7} & \makebox[0.32cm][c]{79.6} & \makebox[0.32cm][c]{63.9} & \makebox[0.32cm][c]{83.8} & \makebox[0.32cm][c]{74.0} & \makebox[0.32cm][c]{75.6} \\
        MATHS & \makebox[0.32cm][c]{-} & \makebox[0.32cm][c]{-} & \makebox[0.32cm][c]{-} & \makebox[0.32cm][c]{-} & \makebox[0.32cm][c]{-} & \makebox[0.32cm][c]{-} & \makebox[0.32cm][c]{-} & \makebox[0.32cm][c]{56.3} & \makebox[0.32cm][c]{74.8} & \makebox[0.32cm][c]{85.6} & \makebox[0.32cm][c]{71.2} & \makebox[0.32cm][c]{69.4} & \makebox[0.32cm][c]{83.5} & \makebox[0.32cm][c]{70.6} & \makebox[0.32cm][c]{52.7} & \makebox[0.32cm][c]{83.6} & \makebox[0.32cm][c]{76.5} & \makebox[0.32cm][c]{57.3} & \makebox[0.32cm][c]{82.9} & \makebox[0.32cm][c]{72.0} & \makebox[0.32cm][c]{74.8} \\
        TNT & \makebox[0.32cm][c]{88.2} & \makebox[0.32cm][c]{83.4} & \makebox[0.32cm][c]{92.7} & \makebox[0.32cm][c]{98.5} & \makebox[0.32cm][c]{93.9} & \makebox[0.32cm][c]{98.6} & \makebox[0.32cm][c]{92.6} & \makebox[0.32cm][c]{55.1} & \makebox[0.32cm][c]{75.3} & \makebox[0.32cm][c]{84.6} & \makebox[0.32cm][c]{72.9} & \makebox[0.32cm][c]{70.0} & \makebox[0.32cm][c]{82.5} & \makebox[0.32cm][c]{71.4} & \makebox[0.32cm][c]{58.7} & \makebox[0.32cm][c]{83.3} & \makebox[0.32cm][c]{79.1} & \makebox[0.32cm][c]{62.4} & \makebox[0.32cm][c]{83.2} & \makebox[0.32cm][c]{73.2} & \makebox[0.32cm][c]{75.2} \\
        NCAL & \makebox[0.32cm][c]{90.8} & \makebox[0.32cm][c]{86.5} & \makebox[0.32cm][c]{91.9} & \makebox[0.32cm][c]{99.5} & \makebox[0.32cm][c]{92.5} & \makebox[0.32cm][c]{100.0} & \makebox[0.32cm][c]{93.6} & \makebox[0.32cm][c]{-} & \makebox[0.32cm][c]{-} & \makebox[0.32cm][c]{-} & \makebox[0.32cm][c]{-} & \makebox[0.32cm][c]{-} & \makebox[0.32cm][c]{-} & \makebox[0.32cm][c]{-} & \makebox[0.32cm][c]{-} & \makebox[0.32cm][c]{-} & \makebox[0.32cm][c]{-} & \makebox[0.32cm][c]{-} & \makebox[0.32cm][c]{-} & \makebox[0.32cm][c]{-} & \makebox[0.32cm][c]{57.5} \\
        \midrule
        Baseline & \makebox[0.32cm][c]{85.4} & \makebox[0.32cm][c]{81.4} & \makebox[0.32cm][c]{87.3} & \makebox[0.32cm][c]{98.6} & \makebox[0.32cm][c]{87.7} & \makebox[0.32cm][c]{99.4} & \makebox[0.32cm][c]{90.0} & \makebox[0.32cm][c]{50.4} & \makebox[0.32cm][c]{70.1} & \makebox[0.32cm][c]{79.6} & \makebox[0.32cm][c]{61.9} & \makebox[0.32cm][c]{62.9} & \makebox[0.32cm][c]{70.0} & \makebox[0.32cm][c]{61.2} & \makebox[0.32cm][c]{45.9} & \makebox[0.32cm][c]{77.2} & \makebox[0.32cm][c]{71.6} & \makebox[0.32cm][c]{52.7} & \makebox[0.32cm][c]{77.3} & \makebox[0.32cm][c]{65.1} & \makebox[0.32cm][c]{60.8} \\
        Ours & \makebox[0.32cm][c]{92.4} & \makebox[0.32cm][c]{84.6} & \makebox[0.32cm][c]{94.7} & \makebox[0.32cm][c]{99.4} & \makebox[0.32cm][c]{94.9} & \makebox[0.32cm][c]{100.0} & \makebox[0.32cm][c]{\textbf{94.4}} & \makebox[0.32cm][c]{65.0} & \makebox[0.32cm][c]{79.7} & \makebox[0.32cm][c]{87.9} & \makebox[0.32cm][c]{77.2} & \makebox[0.32cm][c]{75.4} & \makebox[0.32cm][c]{86.0} & \makebox[0.32cm][c]{80.3} & \makebox[0.32cm][c]{63.9} & \makebox[0.32cm][c]{87.3} & \makebox[0.32cm][c]{81.4} & \makebox[0.32cm][c]{68.5} & \makebox[0.32cm][c]{84.6} & \makebox[0.32cm][c]{\textbf{78.1}} & \makebox[0.32cm][c]{\textbf{80.4}} \\
        \bottomrule
    \end{tabular}
    \caption{Accuracy comparison in the PDA setting. Some results are referred to previous work~\cite{chen2022geometric}.}
    \label{tab:pda}
\end{table*}

\begin{table*}[t]
    \centering
    \small
    \begin{tabular}{c *{20}{m{0.32cm}} c}
        \toprule
        \multirow{2}[2]{*}{\textbf{Method}} & \multicolumn{7}{c}{\textbf{Office(31/0/0)}} & \multicolumn{13}{c}{\textbf{OfficeHome(65/0/0)}} & \multirow{2}[2]{*}{\textbf{\makecell{VisDA\\(12/0/0)}}} \\
        \cmidrule(lr){2-8}\cmidrule(lr){9-21}
        & \makebox[0.32cm][c]{A2D}  & \makebox[0.32cm][c]{A2W}  & \makebox[0.32cm][c]{D2A}  & \makebox[0.32cm][c]{D2W}  & \makebox[0.32cm][c]{W2A}  & \makebox[0.32cm][c]{W2D}  & \makebox[0.32cm][c]{Avg} 
        & \makebox[0.32cm][c]{A2C}  & \makebox[0.32cm][c]{A2P}  & \makebox[0.32cm][c]{A2R}  & \makebox[0.32cm][c]{C2A}  & \makebox[0.32cm][c]{C2P}  & \makebox[0.32cm][c]{C2R}  & \makebox[0.32cm][c]{P2A}  & \makebox[0.32cm][c]{P2C}  & \makebox[0.32cm][c]{P2R}  & \makebox[0.32cm][c]{R2A}  & \makebox[0.32cm][c]{R2C}  & \makebox[0.32cm][c]{R2P}  & \makebox[0.32cm][c]{Avg} & \\
        \midrule
        UAN & \makebox[0.32cm][c]{97.0} & \makebox[0.32cm][c]{86.5} & \makebox[0.32cm][c]{69.6} & \makebox[0.32cm][c]{100.0} & \makebox[0.32cm][c]{68.7} & \makebox[0.32cm][c]{84.5} & \makebox[0.32cm][c]{84.4} & \makebox[0.32cm][c]{45.0} & \makebox[0.32cm][c]{63.6} & \makebox[0.32cm][c]{71.2} & \makebox[0.32cm][c]{51.4} & \makebox[0.32cm][c]{58.2} & \makebox[0.32cm][c]{63.2} & \makebox[0.32cm][c]{52.6} & \makebox[0.32cm][c]{40.9} & \makebox[0.32cm][c]{71.0} & \makebox[0.32cm][c]{63.3} & \makebox[0.32cm][c]{48.2} & \makebox[0.32cm][c]{75.4} & \makebox[0.32cm][c]{58.7} & \makebox[0.32cm][c]{66.4} \\
        CMU & \makebox[0.32cm][c]{78.3} & \makebox[0.32cm][c]{79.6} & \makebox[0.32cm][c]{62.3} & \makebox[0.32cm][c]{98.1} & \makebox[0.32cm][c]{63.4} & \makebox[0.32cm][c]{97.6} & \makebox[0.32cm][c]{79.9} & \makebox[0.32cm][c]{42.8} & \makebox[0.32cm][c]{65.6} & \makebox[0.32cm][c]{74.3} & \makebox[0.32cm][c]{58.1} & \makebox[0.32cm][c]{63.1} & \makebox[0.32cm][c]{67.4} & \makebox[0.32cm][c]{54.2} & \makebox[0.32cm][c]{41.2} & \makebox[0.32cm][c]{73.8} & \makebox[0.32cm][c]{66.9} & \makebox[0.32cm][c]{48.0} & \makebox[0.32cm][c]{78.7} & \makebox[0.32cm][c]{61.2} & \makebox[0.32cm][c]{56.9} \\
        DANCE & \makebox[0.32cm][c]{89.4} & \makebox[0.32cm][c]{88.6} & \makebox[0.32cm][c]{69.5} & \makebox[0.32cm][c]{97.5} & \makebox[0.32cm][c]{68.2} & \makebox[0.32cm][c]{100.0} & \makebox[0.32cm][c]{85.5} & \makebox[0.32cm][c]{54.3} & \makebox[0.32cm][c]{75.9} & \makebox[0.32cm][c]{78.4} & \makebox[0.32cm][c]{64.8} & \makebox[0.32cm][c]{72.1} & \makebox[0.32cm][c]{73.4} & \makebox[0.32cm][c]{63.2} & \makebox[0.32cm][c]{53.0} & \makebox[0.32cm][c]{79.4} & \makebox[0.32cm][c]{73.0} & \makebox[0.32cm][c]{58.2} & \makebox[0.32cm][c]{82.9} & \makebox[0.32cm][c]{69.1} & \makebox[0.32cm][c]{70.2} \\
        DCC & \makebox[0.32cm][c]{87.2} & \makebox[0.32cm][c]{89.1} & \makebox[0.32cm][c]{74.4} & \makebox[0.32cm][c]{96.8} & \makebox[0.32cm][c]{76.8} & \makebox[0.32cm][c]{100.0} & \makebox[0.32cm][c]{87.4} & \makebox[0.32cm][c]{35.4} & \makebox[0.32cm][c]{61.4} & \makebox[0.32cm][c]{75.2} & \makebox[0.32cm][c]{45.7} & \makebox[0.32cm][c]{59.1} & \makebox[0.32cm][c]{62.7} & \makebox[0.32cm][c]{43.9} & \makebox[0.32cm][c]{30.9} & \makebox[0.32cm][c]{70.2} & \makebox[0.32cm][c]{57.8} & \makebox[0.32cm][c]{41.0} & \makebox[0.32cm][c]{77.9} & \makebox[0.32cm][c]{55.1} & \makebox[0.32cm][c]{69.3} \\
        GATE & \makebox[0.32cm][c]{91.3} & \makebox[0.32cm][c]{90.5} & \makebox[0.32cm][c]{73.4} & \makebox[0.32cm][c]{98.7} & \makebox[0.32cm][c]{75.9} & \makebox[0.32cm][c]{100.0} & \makebox[0.32cm][c]{\textbf{88.3}} & \makebox[0.32cm][c]{54.6} & \makebox[0.32cm][c]{76.9} & \makebox[0.32cm][c]{79.8} & \makebox[0.32cm][c]{66.1} & \makebox[0.32cm][c]{73.5} & \makebox[0.32cm][c]{74.2} & \makebox[0.32cm][c]{65.3} & \makebox[0.32cm][c]{54.8} & \makebox[0.32cm][c]{80.6} & \makebox[0.32cm][c]{73.9} & \makebox[0.32cm][c]{59.5} & \makebox[0.32cm][c]{83.7} & \makebox[0.32cm][c]{\textbf{70.2}} & \makebox[0.32cm][c]{\textbf{74.8}} \\
        MATHS & \makebox[0.32cm][c]{90.7} & \makebox[0.32cm][c]{90.4} & \makebox[0.32cm][c]{71.6} & \makebox[0.32cm][c]{97.8} & \makebox[0.32cm][c]{70.5} & \makebox[0.32cm][c]{100.0} & \makebox[0.32cm][c]{86.8} & \makebox[0.32cm][c]{54.7} & \makebox[0.32cm][c]{76.3} & \makebox[0.32cm][c]{78.0} & \makebox[0.32cm][c]{65.4} & \makebox[0.32cm][c]{73.5} & \makebox[0.32cm][c]{74.6} & \makebox[0.32cm][c]{64.8} & \makebox[0.32cm][c]{55.7} & \makebox[0.32cm][c]{78.8} & \makebox[0.32cm][c]{73.8} & \makebox[0.32cm][c]{59.7} & \makebox[0.32cm][c]{83.4} & \makebox[0.32cm][c]{69.9} & \makebox[0.32cm][c]{72.9} \\
        NCAL & \makebox[0.32cm][c]{92.1} & \makebox[0.32cm][c]{87.8} & \makebox[0.32cm][c]{62.9} & \makebox[0.32cm][c]{98.4} & \makebox[0.32cm][c]{65.1} & \makebox[0.32cm][c]{100.0} & \makebox[0.32cm][c]{84.4} & \makebox[0.32cm][c]{-} & \makebox[0.32cm][c]{-} & \makebox[0.32cm][c]{-} & \makebox[0.32cm][c]{-} & \makebox[0.32cm][c]{-} & \makebox[0.32cm][c]{-} & \makebox[0.32cm][c]{-} & \makebox[0.32cm][c]{-} & \makebox[0.32cm][c]{-} & \makebox[0.32cm][c]{-} & \makebox[0.32cm][c]{-} & \makebox[0.32cm][c]{-} & \makebox[0.32cm][c]{-} & \makebox[0.32cm][c]{64.7} \\
        \midrule
        Baseline & \makebox[0.32cm][c]{83.5} & \makebox[0.32cm][c]{81.5} & \makebox[0.32cm][c]{66.5} & \makebox[0.32cm][c]{98.5} & \makebox[0.32cm][c]{66.1} & \makebox[0.32cm][c]{100.0} & \makebox[0.32cm][c]{82.7} & \makebox[0.32cm][c]{49.2} & \makebox[0.32cm][c]{68.6} & \makebox[0.32cm][c]{76.3} & \makebox[0.32cm][c]{58.8} & \makebox[0.32cm][c]{66.4} & \makebox[0.32cm][c]{69.6} & \makebox[0.32cm][c]{56.2} & \makebox[0.32cm][c]{45.1} & \makebox[0.32cm][c]{76.3} & \makebox[0.32cm][c]{69.4} & \makebox[0.32cm][c]{51.7} & \makebox[0.32cm][c]{79.7} & \makebox[0.32cm][c]{63.9} & \makebox[0.32cm][c]{58.5} \\
        Ours & \makebox[0.32cm][c]{91.2} & \makebox[0.32cm][c]{91.0} & \makebox[0.32cm][c]{69.3} & \makebox[0.32cm][c]{99.0} & \makebox[0.32cm][c]{70.6} & \makebox[0.32cm][c]{100.0} & \makebox[0.32cm][c]{86.9} & \makebox[0.32cm][c]{55.1} & \makebox[0.32cm][c]{75.9} & \makebox[0.32cm][c]{77.3} & \makebox[0.32cm][c]{64.8} & \makebox[0.32cm][c]{72.8} & \makebox[0.32cm][c]{72.5} & \makebox[0.32cm][c]{64.0} & \makebox[0.32cm][c]{54.3} & \makebox[0.32cm][c]{78.0} & \makebox[0.32cm][c]{72.5} & \makebox[0.32cm][c]{59.2} & \makebox[0.32cm][c]{83.6} & \makebox[0.32cm][c]{69.2} & \makebox[0.32cm][c]{73.0} \\
        \bottomrule
    \end{tabular}
    \caption{Accuracy comparison in the CDA setting. Some results are referred to previous work~\cite{chen2022geometric}.}
    \label{tab:cda}
\end{table*}

\end{document}